\documentclass{article}

\PassOptionsToPackage{numbers, compress}{natbib}

\usepackage[final]{neurips_2024}

\usepackage[utf8]{inputenc} 
\usepackage[T1]{fontenc}    
\usepackage{hyperref}       
\usepackage{url}            
\usepackage{booktabs}       
\usepackage{amsfonts}       
\usepackage{nicefrac}       
\usepackage{microtype}      
\usepackage{xcolor}         

\usepackage{amsmath}
\usepackage{amssymb}
\usepackage{mathtools}
\usepackage{amsthm}
\usepackage{mathabx}
\usepackage{bbm}
\usepackage{mathrsfs}

\usepackage{abbreviation_definitions}

\usepackage{enumerate}
\usepackage{enumitem}

\usepackage{caption}
\usepackage{graphicx}
\usepackage{subcaption}
\usepackage{multirow}

\usepackage{graphicx}
\usepackage{mwe}

\usepackage{tikz}
\usetikzlibrary{shapes,arrows,positioning,fit,decorations.pathreplacing,calc}

\usepackage{amsmath}
\usepackage{amssymb}
\usepackage{mathtools}
\usepackage{amsthm}
\usepackage{blkarray}
\usepackage{bm}
\usepackage{todonotes}

\usepackage{algorithm}
\usepackage{algpseudocode}

\usepackage[toc,page,header]{appendix}
\usepackage{minitoc}

\title{Nearly Minimax Optimal Regret for Multinomial Logistic Bandit}

\author{%
  Joongkyu Lee \\
  Seoul National University\\
  Seoul, South Korea\\
  \texttt{jklee0717@snu.ac.kr} \\
  \And
  Min-hwan Oh \\
  Seoul National University\\
Seoul, South Korea\\
  \texttt{minoh@snu.ac.kr} \\
}
\doparttoc 
\faketableofcontents 

\begin{document}

\maketitle

\begin{abstract}
In this paper, we study the contextual multinomial logistic (MNL) bandit problem in which a learning agent sequentially selects an assortment based on contextual information, and user feedback follows an MNL choice model.
There has been a significant discrepancy between lower and upper regret bounds, particularly regarding the maximum assortment size $K$.
Additionally, the variation in reward structures between these bounds complicates the quest for optimality.
Under uniform rewards, where all items have the same expected reward, we establish a regret lower bound of $\Omega(d\sqrt{\smash[b]{T/K}})$ and propose a constant-time algorithm, \AlgName{}, that achieves a matching upper bound of $\BigOTilde(d\sqrt{\smash[b]{T/K}})$. 
We also provide instance-dependent minimax regret bounds under uniform rewards.
Under non-uniform rewards, we prove a lower bound of $\Omega(d\sqrt{T})$ and an upper bound of $\BigOTilde(d\sqrt{T})$, also achievable by \AlgName{}. 
Our empirical studies support these theoretical findings. 
To the best of our knowledge, this is the first work in the contextual MNL bandit literature to prove minimax optimality --- for either uniform or non-uniform reward setting --- and to propose a computationally efficient algorithm that achieves this optimality up to logarithmic factors.
\end{abstract}

\section{Introduction}
\label{sec:Introduction}
The multinomial logistic (MNL) bandit framework~\citep{rusmevichientong2010dynamic, saure2013optimal, agrawal2017thompson, agrawal2019mnl, oh2019thompson, oh2021multinomial, perivier2022dynamic, agrawal2023tractable, zhang2024online}  describes sequential assortment selection problems in which an agent offer a sequence of assortments of at most $K$ items from a set of $N$ possible items and receives feedback \textit{only} for the chosen decisions.
The choice probability of each outcome is characterized by an MNL model~\citep{mcfadden1977modelling}.
This framework allows modeling of various real-world situations such as recommender systems and online retails, where selections of assortments are evaluated based on the 
user-choice feedback
among offered multiple options.

In this paper, we study the \textit{contextual} MNL bandit problem~\citep{agrawal2019mnl,agrawal2017thompson,
ou2018multinomial, chen2020dynamic, oh2019thompson, oh2021multinomial, perivier2022dynamic, agrawal2023tractable}, where the features of 
items and possibly contextual information about a user at each round are available.
Despite many recent advances, 
\citep{chen2020dynamic, oh2019thompson, oh2021multinomial, perivier2022dynamic, agrawal2023tractable},
however,
no previous studies have proven the minimax optimality of contextual MNL bandits.
\citet{chen2020dynamic} proposed a regret lower bound of $\Omega(d\sqrt{T}/K)$, where $d$ is the feature dimension, $T$ is the total number of rounds, and $K$ is the maximum size of assortments,
assuming the uniform rewards, 
i.e., rewards are all same for each of the total $N$ items.
Furthermore,~\citet{chen2018note} established a regret lower bound of $\Omega(\sqrt{NT})$  in the non-contextual setting (hence, dependence on $N$ appears instead of $d$), which is tighter in terms of $K$.
It is important to note the difference in the assumptions for the \textit{attraction parameter for the outside option} $v_0$.
\citet{chen2018note} assumed for the attraction parameter for the outside option to be $v_0 = K$,
whereas~\citet{chen2020dynamic} assumed
$v_0 = 1$. 
Therefore, it remains an \textit{open question whether
and how 
the value of $v_0$ affects 
both lower and upper bounds of 
regret}.

Regarding regret upper bounds,~\citet{chen2020dynamic} proposed an exponential runtime algorithm that achieves a regret of $\BigOTilde(d\sqrt{T}) $ in the setting with \textit{stochastic} contexts and the \textit{non-uniform} rewards. 
Under the same setting,~\citet{oh2021multinomial} and~\citet{oh2019thompson} introduced polynomial-time algorithms that attain regrets of $\BigOTilde(d\sqrt{T} / \kappa)$ and $\BigOTilde(d^{3/2}\sqrt{T} / \kappa)$ respectively, where $1/\kappa = \BigO(K^2)$ is a problem-dependent constant.
Recently,~\citet{perivier2022dynamic} improved the dependency on $\kappa$ in the \textit{adversarial} context setting, achieving a regret of  $\BigOTilde(d K \sqrt{\kappa^{\prime} T})$, where $\kappa^{\prime} = \BigO(1/K)$.
However, their approach focuses solely on the setting with \textit{uniform} rewards, which is a special case of non-uniform rewards, and currently, there is no tractable method to implement the algorithm.

As summarized in Table~\ref{tab:regrets}, there has been a gap between the upper and lower bounds in the existing works of contextual MNL bandits.
No previous studies have confirmed whether lower or upper bounds are tight,
obscuring what the optimal regret should be.
This ambiguity is further exacerbated because many studies introduce their methods under varying conditions such as different reward structures and values of $v_0$, without explicitly explaining how these factors impact regret. 
Additionally, there is currently no computationally efficient algorithm whose regret does not scale with $1/\kappa = \BigO(K^2)$ or directly with $K$.
Intuitively, increasing $K$ provides more information 
at least in the uniform reward setting, 
potentially leading to a more statistically efficient learning process.
However, no previous results have reflected such intuition.
Hence, the following research questions arise:
\begin{itemize}
    \item \textit{What is the \underline{optimal regret lower bound} in contextual MNL bandits?}
    \item \textit{Can we design a \underline{computationally efficient}, nearly \underline{minimax optimal} algorithm under the \underline{adversarial context setting}?} 
\end{itemize}
\begin{table}[t]
\caption{Comparisons of lower and upper regret bounds in related works on MNL bandits with $T$ rounds, $N$ items, the maximum size of assortments $K$, $d$-dimensional feature vectors, and problem-dependent constants $1/\kappa = \BigO(K^2)$ and $\kappa' = \BigO(1/K)$.
$\BigOTilde$ represents  big-$\BigO$ notation up to logarithmic factors.
For the computational cost (abbreviated as ``Comput.''), we consider only the dependence on the number of rounds $t$.
``Intractable'' means a non-polynomial runtime.
The notation ``$-$'' denotes \textit{not applicable}.
The starred ($^*$) papers only consider the non-contextual setting.
}
\resizebox{\textwidth}{!}{
\begin{tabular}{cllcccc}
\toprule
\multicolumn{1}{l}{}         &                            & Regret & Contexts & Rewards & $v_0$ &  Comput. per Round \\
\midrule
\multirow{5}{*}{\begin{tabular}[c]{@{}c@{}}Lower \\ Bound\end{tabular}} 
                             & \citet{chen2020dynamic}                            &   $\Omega(d\sqrt{T}/K)$      & $-$ &   Uniform      &  $\Theta(1)$   &    $-$ \\
                             & \citet{agrawal2019mnl}$^*$                               &   $\Omega(\sqrt{NT/K})$& $-$ &   Uniform     &  $\Theta(K)$   &    $-$ \\
                             & \citet{chen2018note}$^*$                               &   $\Omega(\sqrt{NT})$&     $-$ &   Uniform     &  $\Theta(K)$   &    $-$ \\
                             &\textbf{This work }(Theorem~\ref{thm:lower_bound}) &   $\Omega(\frac{\sqrt{v_0 K}}{ v_0 + K  } d\sqrt{T})$  &   $-$  &   Uniform      &  Any value  &    $-$ \\
                             &\textbf{This work} (Theorem~\ref{theorem:lower_bound_non_uniform})  &   $\Omega(d\sqrt{T})$   &  $-$  &   Non-uniform     &  $\Theta(1)$   &    $-$ \\
\midrule
\multirow{6}{*}{\begin{tabular}[c]{@{}c@{}}Upper \\ Bound\end{tabular}}
                             & \citet{chen2020dynamic}                            & $\BigOTilde(d\sqrt{T})$    &  Stochastic  &  Non-uniform  &  $\Theta(1)$    &   Intractable        \\
                             & \citet{oh2021multinomial}                          & $\BigOTilde(d \sqrt{T}/\kappa)$ & Stochastic &  Non-uniform  &  $\Theta(1)$    &   $\BigO(t)$ \\
                             & \citet{oh2019thompson}                          & $\BigOTilde(d^{3/2} \sqrt{T}/\kappa)$ & Adversarial &  Non-uniform  &  $\Theta(1)$    &   $\BigO(t)$ \\
                             & \citet{perivier2022dynamic}                           & $\BigOTilde(dK \sqrt{\kappa^{\prime} T})$    &  Adversarial &      Uniform  &  $\Theta(1)$    &    Intractable        \\
                             & \textbf{This work} (Theorem~\ref{thm:upper_bound})                 &  $\BigOTilde\left(  \frac{\sqrt{v_0 K}}{ v_0 + K  } d\sqrt{T} \right)$   & Adversarial   &      Uniform  &  Any value    &   $\BigO(1)$        \\
                             & \textbf{This work} (Theorem~\ref{theorem:upper_bound_non_uniform}) & $\BigOTilde(d \sqrt{T})$  &  Adversarial   &  Non-uniform  &  $\Theta(1)$  &    $\BigO(1)$        \\
\bottomrule
\end{tabular}
}
\label{tab:regrets}
\end{table}

In this paper, we affirmatively answer the questions by first tackling the contextual MNL bandit problem separately 
based on
the structure of rewards—uniform and non-uniform—and the value of the outside option $v_0$.
In the setting of uniform rewards, we establish 
the tightest regret lower bound, 
explicitly demonstrating the dependence of regret on $v_0$.
Specifically, we prove a regret lower bound of $\Omega(d\sqrt{\smash[b]{T/K}})$ when $v_0 = \Theta(1)$, a common assumption in contextual settings~\citep{agrawal2013thompson,  cheung2017thompson, ou2018multinomial, agrawal2019mnl,  oh2019thompson, oh2021multinomial, amani2021ucb, perivier2022dynamic, agrawal2023tractable, zhang2024online, lee2024improved} (see Appendix~\ref{app_subsec:outside} for more details), 
and a lower bound of $\Omega(d\sqrt{T})$  when $v_0 = \Theta(K)$.
Furthermore, in the adversarial context setting, we introduce a computationally efficient and provably optimal (up to logarithmic factors) algorithm, \AlgName{}. 
We prove that our proposed algorithm achieves a regret of $\BigOTilde(d\sqrt{T/K})$ when $v_0 = \Theta(1)$ and $\BigOTilde(d\sqrt{T})$ when $v_0 = \Theta(K)$, each of which matches the respective lower bounds that we establish up to logarithmic factors.
Furthermore, in the non-uniform reward setting, we provide the optimal lower bound of $\Omega(d\sqrt{T})$ assuming $v_0 = \Theta(1)$.
In the same setting, our proposed algorithm also attains a matching upper
bound of $\BigOTilde(d\sqrt{T})$ up to logarithmic factors.
Our main contributions are summarized as follows:
\begin{itemize}
    \item Under uniform rewards, we establish a regret lower bound of $\Omega(\sqrt{v_0 K}/(v_0 + K )d\sqrt{T})$ (Theorem~\ref{thm:lower_bound}), which is the tightest known lower bound in contextual MNL bandits. 
    We propose, for the first time, a computationally efficient and provably optimal algorithm, \AlgName{}, achieving a matching upper bound of $\BigO(\sqrt{v_0 K}/(v_0 + K )d\sqrt{T})$ (Theorem~\ref{thm:upper_bound}) up to logarithmic factors, while requiring only a constant computation cost per round.
    The results indicate that the regret improves as the assortment size $K$ increases, unless $v_0 = \Theta(K)$.
    To the best of our knowledge, this is the first study to demonstrate the dependence of regret on the attraction parameter for the outside option $v_0$ and to highlight the advantages of a larger assortment size $K$ which aligns with intuition.
    That is, this is the first work to show that a regret upper bound (in either contextual or non-contextual setting) decreases as $K$ increases.
    Additionally, we provide instance-dependent minimax regret bounds (Proposition~\ref{prop:lower_instance} and~\ref{prop:upper_instance}), up to logarithmic factors.
    
    \item Under non-uniform rewards, with setting $v_0= \Theta(1)$ following the convention in contextual MNL bandits~\citep{agrawal2013thompson,  cheung2017thompson, ou2018multinomial, agrawal2019mnl,  oh2019thompson, oh2021multinomial, amani2021ucb, perivier2022dynamic, agrawal2023tractable, zhang2024online, lee2024improved}, we establish a regret lower bound of $\Omega(d\sqrt{T})$ (Theorem~\ref{theorem:lower_bound_non_uniform}).
    To the best of our knowledge, this is the first and tightest lower bound established under non-uniform rewards.
    Moreover, \AlgName{} also achieves a matching upper bound (up to logarithmic factors) of $\BigOTilde(d\sqrt{T})$ (Theorem~\ref{theorem:upper_bound_non_uniform}) in this setting.

    \item We also conduct numerical experiments and show that our algorithm consistently outperforms the existing MNL bandit algorithms while maintaining a constant computation cost per round. 
    Furthermore, the empirical results corroborate our theoretical findings regarding the dependence of regret on the reward structure, $v_0$ and $K$.
\end{itemize}

Overall, our paper addresses the long-standing open problem of closing the gap between upper and lower bounds for contextual MNL bandits. 
Our proposed algorithm is the first to achieve both provably optimality (up to logarithmic factors) and practicality with improved computation.

\section{Related Work}
\label{sec:Related}

\textbf{Lower bounds of MNL bandits.}
In contextual MNL bandits, to the best of our knowledge, only~\citet{chen2020dynamic} proved a lower bound of $\Omega(d\sqrt{T} /K)$ with the attraction parameter for the outside option set at $v_0 = 1$.
However, in the non-contextual setting, there exist improved lower bounds in terms of $K$.
~\citet{agrawal2019mnl} demonstrated a lower bound of  $\Omega(\sqrt{NT/K})$,  and~\citet{chen2018note} established a lower bound of $\Omega(\sqrt{NT})$.
By setting $d=N$, one can derive equivalent lower bounds for the contextual setting, specifically $\Omega(\sqrt{dT/K})$ and $\Omega(\sqrt{dT})$, respectively.
However,~\citet{agrawal2019mnl} and~\citet{chen2018note} assumed $v_0 = K$ when establishing their lower bounds, which differs from the setting used by~\citet{chen2020dynamic}, where $v_0 = 1$.
Moreover, to the best of our knowledge, all existing works~\citet{chen2020dynamic, agrawal2019mnl, chen2018note} have established the lower bounds under uniform rewards.
Consequently, it remains unclear what the optimal regret is, depending on the value of $v_0$ and the reward structure.

\textbf{Upper bounds of contextual MNL bandits.}
\citet{ou2018multinomial} formulated a linear utility model and achieved $\BigOTilde(dK\sqrt{T})$ regret; however, they assumed that utilities are fixed over time.
\citet{chen2020dynamic} considered contextual MNL bandits with changing and stochastic contexts, establishing a regret of $\BigOTilde(d\sqrt{T}+ d^2K^2 )$.
However, they encountered computational issues due to the need to enumerate all possible ($N$ choose $K$) assortments.
To address this,~\citet{oh2021multinomial} proposed a polynomial-time assortment optimization algorithm, which maintains the confidence bounds in the parameter space and then calculates the upper confidence bounds of attraction parameter for each item, achieving a regret of $\BigOTilde(d\sqrt{T} / \kappa)$, where $1/\kappa = \BigO(K^2)$ is a problem-dependent constant.
\citet{perivier2022dynamic} considered the adversarial context and uniform reward setting and improved the dependency on $\kappa$ to $\BigOTilde(d K\sqrt{\kappa^{\prime} T} + d^2 K^4/\kappa)$, where $\kappa^{\prime} = \BigO(1/K)$.
However, their algorithm is intractable.
\citet{agrawal2023tractable} considered a uniform rewards setting (with $v_0=1$) and achieved a regret of $\BigOTilde (d \sqrt{T} )$.
However, due to significant technical errors in their paper (refer Appendix~\ref{app_sec:tech_error_agrawl}), we do not include a comparison with their results in this work.

Recently,~\citet{zhang2024online} utilized an online parameter update to construct a constant time algorithm.
However, they consider a \textit{multiple-parameter} choice model in which the learner estimates $K$ parameters and shares the contextual information $x_t$ across the items in the assortment. 
This model differs from ours; we use a \textit{single-parameter} choice model with varying the context for each item in the assortment.
Additionally, they make a stronger assumption regarding the reward than we do (see Assumption~\ref{assum:bounded_assumption}).
Moreover, while they fix the assortment size at $K$, we allow it to be smaller than or equal to $K$.
On the other hand,~\citet{zhang2024contextual} considered a general function approximation, achieving a regret bound of $\BigOTilde(K^{2.5} \sqrt{dNT})$.
However, this bound scales with $K$ and $N$, and the proposed algorithm is not  tractable.
To the best of our knowledge, all existing methods fail to show that the regret upper bound can improve as the assortment size $K$ increases.


\section{Problem Setting}
\label{sec:problem_setting}
\textbf{Notations.}
For a positive integer, $n$, we denote $[n] := \{1, 2, \ldots, n \}$. 
For a real-valued matrix $A$, we denote $\|A\|_2 := \sup_{x:\|x\|_2=1}\|Ax\|_2$ as the maximum singular value of $A$. 
For two symmetric matrices, $V$ and $W$ of the same dimensions, $V \succeq W$ means that $V-W$ is positive semi-definite.
Finally, we define $\mathcal{S}$ to be the set of candidate assortment with size constraint at most $K$, i.e., $ \mathcal{S} = \{S \subseteq [N]: |S| \leq K \}$.
While, for simplicity, we consider both $\mathcal{S}$ and the set of items $[N]$ to be stationary in this paper, it is important to note that both $\mathcal{S}$ and $[N]$ can vary over time.

\textbf{Contextual MNL bandits.}
We consider a sequential assortment selection problem which is defined as follows.
At each round $t$, the agent observes feature vectors $x_{ti} \in \RR^d$ for every item $i \in [N]$.
Based on this contextual information, the agent presents an assortment $S_t = \{ i_1, \dots, i_{l} \} \in \mathcal{S}$, where $l \leq K$, and then observes the user purchase decision $c_t \in S_t\cup \{0\}$, where $\{0\}$ represents the ``outside option'' which indicates that the user did not select any of the items in $S_t$.
The distribution of these selections follows a multinomial logistic (MNL) choice model~\citep{mcfadden1977modelling}, where the probability of choosing any item $i_k \in S_t$ (or the outside option) is defined as:
\begin{align}
    \label{eq:mnl_model}
    p_t(i_k | S_t, \wb^\star) 
    := \frac{\exp(x_{ti_k}^\top  \wb^\star )}{ v_0 \!+\!\sum_{j \in S_t }\exp( x_{tj}^\top \wb^\star )},
    \quad
    p_t(0 | S_t, \wb^\star) 
    := \frac{v_0}{ v_0 \!+ \!\sum_{j \in S_t }\exp( x_{tj}^\top \wb^\star )},
\end{align}
where $v_0$ is a \textit{known} attraction parameter for the outside option and $\wb^\star \in \RR^d$ is an \textit{unknown} parameter.
\begin{remark}
    In the existing literature on MNL bandits, it is commonly assumed that $v_0 = 1$~\citep{oh2019thompson, oh2021multinomial, perivier2022dynamic, agrawal2023tractable, zhang2024online}.
    On the other hand,~\citet{chen2018note, agrawal2019mnl} assume that $v_0 = K$~\footnote{
    \citet{chen2018note} indeed set $v_0=1$ and $v_1, \dots, v_N = \Theta(1/K)$.
    However, this is equivalent to the setting with $v_0= K$ and $v_1, \dots, v_N = \Theta(1)$.
    }  to induce a tighter lower bound in terms of $K$.
    Later, we will explore how these differing assumptions create fundamentally different problems, leading to different regret lower bounds (Subsection~\ref{subsec:main_lower}).
\end{remark}
The choice response for each item $i \in S_t \cup \{ 0\}$ is defined as $y_{ti} := \mathbbm{1}(c_t = i) \in \{0,1\}$.
Hence, the choice feedback variable $\yb_t := (y_{t0}, y_{ti_1}, \dots y_{ti_{l}}) $ is sampled from the following multinomial (MNL) distribution:
$ \yb_{t} \sim \operatorname{MNL} \{ 1, ( p_t(0 | S_t, \wb^\star), \dots, p_t(i_l | S_t, \wb^\star)  )\}$,
where the parameter 1 indicates that $\yb_t$ is a single-trial sample, i.e., $y_{t0} + \sum_{k=1}^l y_{ti_k} = 1$. 
For each $i \in S_t \cup \{ 0\}$, we define the noise $\epsilon_{ti}:= y_{ti} - p_{t}(i | S_t, \wb^\star)$.
Since each $\epsilon_{ti}$ is a bounded random variable in $[0,1]$, $\epsilon_{ti}$ is $1/4$-sub-Gaussian.
At every round $t$, the reward $r_{ti}$ for each item $i$ is also given.
Then, we define the expected revenue of the assortment $S$ as 
\begin{align*}
    R_{t}(S, \wb^\star):=  \sum_{i \in S} p_t(i | S, \wb^\star) r_{ti} 
    = \frac{ \sum_{i \in S }\exp(x_{ti}^\top  \wb^\star )r_{ti} }{ v_0 \!+\!\sum_{j \in S }\exp( x_{tj}^\top \wb^\star )}
\end{align*}
and define $S_{t}^\star$ as the offline optimal assortment at time $t$ when $\wb^\star$ is known a prior, i.e.,  $S_{t}^\star = \argmax_{S \in \mathcal{S}}  R_{t}(S, \wb^\star)$.
Our objective is to minimize the cumulative regret over the $T$ periods:
\begin{align*}
    \Regret (\wb^\star )
    = \sum_{t=1}^T   R_{t}(S_{t}^\star, \wb^\star) 
    - R_{t}(S_{t}, \wb^\star). 
\end{align*}
When $K=1$, $r_{t1}=1$, and $v_0=1$, the MNL bandit recovers the binary logistic bandit with $R_t(S=\{x\}, \wb^\star) = \sigma\left( x^\top \wb^\star \right) = 1/ (1 + \exp (-x^\top \wb^\star) )$, where $\sigma(\cdot)$ is the sigmoid function.

Consistent with previous works on MNL bandits~\citep{oh2021multinomial, perivier2022dynamic, agrawal2023tractable, zhang2024online}, we make the following assumptions:
\begin{assumption}[Bounded assumption] 
\label{assum:bounded_assumption}
We assume that $\| \wb^\star \|_2  \leq 1$, and for all $t \geq 1$, $i \in [N]$, $\| x_{ti} \|_2 \leq 1$ and $ r_{ti}  \in [0,1]$.
\end{assumption}
\begin{assumption}[Problem-dependent constant] 
\label{assum:kappa}
There exists $\kappa >0$ such that for every item $i \in S$, any $S \in \mathcal{S}$, and all round $t$, $\min_{\wb \in \mathcal{W}} p_t(i | S, \wb) p_t(0 | S, \wb) \geq \kappa$, where $\mathcal{W} = \{ \wb \in \RR^d \mid \| \wb \|_2 \leq 1 \}$.
\end{assumption}
In Assumption~\ref{assum:bounded_assumption}, we assume that the reward for each item $i$ is bounded by a constant, allowing the norm of the reward vector to depend on $K$, e.g., $\| \bm{\rho}_{t} \|_2 \leq \sqrt{K}$.
In contrast,~\citet{zhang2024online} assume that the norm of the reward vector $\bm{\rho}_{t} = [r_{t1}, \dots r_{t|S_t|}]^\top \in \RR^{|S_t|}$ is bounded by a constant, independent of $K$, e.g., $\| \bm{\rho}_{t} \|_2  \leq 1$.
Thus, our assumption regarding rewards is weaker than theirs.

Assumption~\ref{assum:kappa} is common in contextual MNL bandits~\citep{chen2020dynamic, oh2021multinomial, perivier2022dynamic, zhang2024online}.
Note that $1/\kappa$ depends on the maximum size of the assortment $K$, i.e., $1/\kappa = \BigO(K^{2} )$.
One of the primary goals of this paper is to show that as the assortment size $K$ increases, we can achieve an improved (or at least not worsened) regret bound.
To this end, we design a dynamic assortment policy that enjoys improved dependence on $\kappa$.
Note that our algorithm does not need to know $\kappa$ a priori, whereas \citet{oh2019thompson, oh2021multinomial} do.

\section{Existing Gap between Upper and Lower Bounds in MNL Bandits}
\label{sec:contribution}
The primary objective of this paper is to establish minimax regrets in contextual MNL bandits. 
To explore the optimality of regret, we analyze how it depends on the attraction parameter for the outside option $v_0$, 
the maximum assortment size $K$, and the structure of rewards.

\textbf{Dependence on \texorpdfstring{$v_0$}{}.}
Currently, the established lower bounds are $\Omega(d\sqrt{T}/K)$ by~\citet{chen2020dynamic}, $\Omega(\sqrt{\smash[b]{dT/K}})$ by the contextual version of~\citet{agrawal2019mnl}, and $\Omega(\sqrt{dT})$, which is the tightest in terms of $K$, by the contextual version of~\citet{chen2018note}.
These results can be misleading, as many subsequent studies~\citep{oh2021multinomial, miao2021dynamic, chen2021optimal, zhang2024contextual} have claimed that a $K$-independent regret is achievable, without clearly addressing the influence of the value of $v_0$.
In fact, the improved regret bounds (in terms of $K$) obtained by~\citet{agrawal2019mnl} and~\citet{chen2018note} were possible when $v_0=K$.
However, in the contextual setting, it is more common to set $v_0=\Theta(1)$.
This is because, given the context for the outside option $x_{t0}$, it is straightforward to construct an equivalent choice model where $v_0=\Theta(1)$ (refer Appendix~\ref{app_subsec:outside}).
In this paper, we rigorously show the regret dependency on the value of $v_0$.
In Theorem~\ref{thm:lower_bound}, we establish a regret lower bound of $\Omega\left( \sqrt{v_0 K}/( v_0 + K) d\sqrt{T} \right)$, which implies that the value of $v_0$, indeed, affects the regret.
Then, in Theorem~\ref{thm:upper_bound}, we show that our proposed computationally efficient algorithm, \AlgName{} achieves a regret of $\BigOTilde\left( \sqrt{v_0 K}/( v_0 + K) d\sqrt{T} \right)$, which is minimax optimal up to logarithmic factors in terms of all $d, T, K$ and even $v_0$.

\textbf{Dependence on \texorpdfstring{$K$}{} \& Uniform/Non-uniform rewards.}
To the best of our knowledge, the regret bound in all existing works in contextual MNL bandits does not decrease as the assortment size $K$ increases~\citep{chen2020dynamic, oh2019thompson, oh2021multinomial, perivier2022dynamic}.
However, intuitively, as the assortment size increases, we can gain more information because we receive more feedback.
Therefore, it makes sense that regret could be reduced as $K$ increases, at least in the uniform reward setting.
Under uniform rewards, the expected revenue (to be specified later) increases as more items are added in the assortment.
Consequently, both the optimistically chosen assortment and the optimal assortment always have a size of $K$.
Thus, the agent obtain information about exactly $K$ items in each round.
This phenomenon is also demonstrated empirically in Figure~\ref{fig:regret_main}.
In the uniform reward setting, as $K$ increases, the cumulative regrets of not only our proposed algorithm but also other baseline algorithms decrease. 
This indicates that the existing regret bounds are not tight enough in terms of $K$.
Conversely, in the non-uniform reward setting, the sizes of both the optimistically chosen assortment and the optimal assortment can be less than $K$, so performance improvement is not guaranteed.
In this paper, we show that the regret dependence on $K$ varies by case: uniform and non-uniform rewards.
When $v_0 = \Theta(1)$, we obtain a regret lower bound of $\Omega(d \sqrt{\smash[b]{T/K}})$ (Theorem~\ref{thm:lower_bound}) and a regret upper bound of $\BigOTilde(d \sqrt{\smash[b]{T/K}})$  (Theorem~\ref{thm:upper_bound}) under uniform rewards.
Additionally, we achieve a regret lower bound of (Theorem~\ref{theorem:lower_bound_non_uniform}) and a regret upper bound of $\BigOTilde(d \sqrt{T})$ (Theorem~\ref{theorem:upper_bound_non_uniform}) under non-uniform rewards.

\section{Algorithms and Main Results}
\label{sec:main}
In this section, we begin by proving the tightest regret lower bound under uniform rewards (Subsection~\ref{subsec:main_lower}), explicitly showing the dependence on the attraction parameter for the outside option $v_0$.
We then introduce \AlgName{}, an algorithm that achieves minimax optimality, up to logarithmic factors, under \textit{uniform rewards} (Subsection~\ref{subsec:main_upper}).
Notably, \AlgName{} is designed for efficiency, requiring only constant computation and storage costs.
Finally, we establish the tightest regret lower bound and a matching minimax optimal regret upper bound under \textit{non-uniform rewards} (Subsection~\ref{subsec:main_non_uniform}).

\subsection{Regret Lower Bound under Uniform Rewards}
\label{subsec:main_lower}
In this subsection, we present a lower bound for the worst-case expected regret in the uniform reward setting ($r_{ti}=1$).
This covers all applications where the objective is to maximize the appropriate ``click-through rate'' by offering the assortment.
\begin{theorem} [Regret lower bound, Uniform rewards] \label{thm:lower_bound}
    Let $d$ be divisible by $4$ and let Assumption~\ref{assum:bounded_assumption} hold true.
    Suppose $T \geq C \cdot d^4 (v_0 + K) / K$ for some constant $C>0$.
    Then, in the uniform reward setting, for any policy $\pi$, there exists a worst-case problem instance such that the worst-case expected regret of $\pi$ is lower bounded as follows:
    \begin{align*}
        \sup_{\wb} \EE^\pi_{\wb} \left[ \Regret(\wb) \right]
        = \Omega\left( \frac{\sqrt{v_0 K}}{ v_0 + K  } \cdot d\sqrt{T} \right).
    \end{align*}
\end{theorem}
\textbf{Discussion of Theorem~\ref{thm:lower_bound}.}
If $v_0=\Theta(1)$, Theorem~\ref{thm:lower_bound} demonstrates a regret lower bound of $\Omega(d \sqrt{\smash[b]{T/K}})$.
This indicates that, under uniform rewards, increasing the assortment size $K$ leads to an improvement in regret. 
Compared to the lower bound $\Omega(d \sqrt{T}/K)$ proposed by~\citet{chen2020dynamic}, our lower bound is improved by a factor of $\sqrt{K}$.
This improvement is mainly due to the establishment of a tighter upper bound for the KL divergence (Lemma~\ref{lemma:lower_KL}). 
Notably,~\citet{chen2020dynamic} also considered uniform rewards with $v_0=\Theta(1)$.
On the other hand,~\citet{chen2018note} and~\citet{agrawal2019mnl} established regret lower bounds of $\Omega(\sqrt{NT})$ and $\Omega(\sqrt{NT/K})$,  respectively, in non-contextual MNL bandits with uniform rewards, by setting $v_0 = K$ to achieve these regrets.
Theorem~\ref{thm:lower_bound} shows that if $v_0 = \Theta(K)$, we can obtain a regret lower bound of $\Omega(d \sqrt{T})$, which is consistent with the $K$-independent regret in~\citet{chen2018note}.
To the best of our knowledge, this result is the first to explicitly show the dependency of regret on the attraction parameter for the outside option $v_0$.
The proof is deferred to Appendix~\ref{app_sec:proof_lower_bound}.

\subsection{Minimax Optimal Regret Upper Bound under Uniform Rewards}
\label{subsec:main_upper}
In this subsection, we propose a new algorithm~\AlgName{}, which enjoys minimax optimal regret up to logarithmic factors in the case of uniform rewards.
Note that, since the revenue is an increasing function when rewards are uniform, maximizing the expected revenue $ R_{t}(S, \wb)$ over all $S \in \mathcal{S}$ always yields exactly $K$ items, i.e., $|S_t| = |S_t^\star| = K$.

Our first step involves constructing the confidence set for the online parameter.

\textbf{Online parameter estimation.}
Instead of performing MLE as in previous works~\citep{chen2020dynamic,oh2021multinomial, perivier2022dynamic}, inspired by~\citet{zhang2024online}, we use the online mirror descent algorithm to estimate parameter.
We first define the multinomial logistic loss function at round $t$ as:
\begin{align}
    \label{eq:loss}
    \ell_t(\wb) := - \sum_{i \in S_t} y_{ti} \log p_t(i | S_t, \wb).
\end{align}
In Proposition~\ref{prop:self_concordant}, we will show that the loss function has the constant parameter self-concordant-like property.
We estimate the true parameter $\wb^\star$ as follows:
\begin{equation} \label{eq:online_update}
    \wb_{t+1} 
    = \argmin_{\wb \in \mathcal{W}} \langle \nabla \ell_{t}(\wb_t), \wb \rangle
    + \frac{1}{2 \eta} \| \wb - \wb_{t} \|_{\tilde{H}_{t}}^2 \, , \quad \forall t \geq 1,
\end{equation}
where $\eta > 0$ is the step-size parameter to be specified later, and
$\tilde{H}_{t} := H_t + \eta \Gcal_t(\wb_t)$, where 
\begin{align*}
    \Gcal_t(\wb) := \sum_{i \in S_t} p_t(i | S_t, \wb) x_{ti} x_{ti}^\top 
    -  \sum_{i \in S_t}  \sum_{j \in S_t} p_t(i | S_t, \wb) p_t(j | S_t, \wb) x_{ti} x_{tj}^\top,
\end{align*}
and $H_t := \lambda \Ib_d + \sum_{s=1}^{t-1} \Gcal_s(\wb_{s+1})$.
Note that $\Gcal_t(\wb) = \nabla^2 \ell_{t}(\wb)$.
This online estimator is efficient in terms of both computation and storage.
By a standard online mirror descent formulation~\citep{orabona2019modern}, the optimization problem in Equation~\eqref{eq:online_update} can be solved using a single projected gradient step through the following equivalent formula:
\begin{align}
    \wb^{\prime}_{t+1} = \wb_t - \eta \tilde{H}_t^{-1} \nabla \ell_t(\wb_t),
    \quad
    \text{and}
    \quad
    \wb_{t+1} = \argmin_{\wb \in \mathcal{W}} \| \wb - \wb^{\prime}_{t+1} \|_{\tilde{H}_t},
\end{align}
which enjoys a computational cost of only $\mathcal{O}(Kd^3)$, completely independent of $t$~\citep{mhammedi2019lipschitz, zhang2024online}.
Regarding storage costs, the estimator does not need to store all historical data because both $\tilde{H}_t$ and $H_t$ can be updated incrementally, requiring only $\mathcal{O}(d^2)$ storage.

Furthermore, the estimator allows for a $\kappa$-independent confidence set, leading to an improved regret.
\begin{lemma} [Online parameter confidence set] \label{lemma:online_confidence_set}
    Let $\delta \in (0, 1]$. 
    Under Assumption~\ref{assum:bounded_assumption}, with $\eta = \frac{1}{2}\log (K+1) + 2$ and  $\lambda = 84 \sqrt{2}d \eta$, we define the following confidence set
    \begin{align*}
        \Ccal_t(\delta) 
        := \{ \wb \in \mathcal{W} \mid \| \wb_t - \wb \|_{H_t} \leq \beta_{t}(\delta) \},
    \end{align*}
    where $\beta_{t}(\delta) =\BigO \left( \sqrt{d} \log t \log K \right)$.
    Then, we have $\textup{Pr}[\forall t \geq 1, \wb^\star \in \Ccal_t(\delta)] \geq 1- \delta$.
\end{lemma}
Armed with the online estimator, we construct the computationally efficient optimistic revenue.
%
\begin{algorithm}[t!]
   \caption{\AlgName{} }
   \label{alg:minimax}
    \begin{algorithmic}[1]
        \State \textbf{Inputs:} regularization parameter $\lambda$, probability $\delta$, confidence radius $\beta_t(\delta)$, step size $\eta$.
        \State \textbf{Initialize:} $H_1 = \lambda \mathbf{I}_d$ and $\wb_1$ as any point in $\mathcal{W}$.
        \For{round $t=1,2, \cdots, T$}
            \State Compute $\alpha_{ti} = x_{ti}^\top \wb_t + \beta_{t}(\delta) \|x_{ti}\|_{H_t^{-1}}$ for all $i \in [N]$.
            \State Offer $S_t = \argmax_{S \in \mathcal{S}} \tilde{R}_t (S)$ and observe $\yb_t$.
            \State Update $\tilde{H}_{t} = H_t + \eta \Gcal_t(\wb_t)$,
            and update the estimator $\wb_{t+1}$ by Equation~\eqref{eq:online_update}.
            \State Update $H_{t+1} = H_t + \Gcal_t(\wb_{t+1})$.
        \EndFor
    \end{algorithmic}
\end{algorithm}
%

\textbf{Computationally efficient optimistic expected revenue.} 
To balance the exploration and exploitation trade-off, we use the upper confidence bounds (UCB) technique, which have been widely studied in many bandit problems, including $K$-arm bandits~\citep{auer2002finite, lattimore2020bandit} and linear bandits~\citep{abbasi2011improved, chu2011contextual}.

At each time $t$, given the confidence set in Lemma~\ref{lemma:online_confidence_set}, we first calculate the optimistic utility $\alpha_{ti}$ as:
\begin{align}
    \alpha_{ti} := x_{ti}^\top \wb_t + \beta_t(\delta) \| x_{ti} \|_{H_t^{-1}},
    \quad \, \forall i \in [N]
    .
    \label{eq:optimistic_utility}
\end{align}
The optimistic utility $\alpha_{ti}$ is composed of two parts: the mean utility estimate $x_{ti}^\top \wb_t$ and the standard deviation $\beta_t(\delta) \| x_{ti} \|_{H_t^{-1}}$.
In the proof of the regret upper bound, we show that $\alpha_{ti}$ serves as an upper bound for $x_{ti}^\top \wb^\star$, assuming that the true parameter $\wb^\star$ falls within the confidence set $\Ccal_t(\delta)$.
Based on $\alpha_{ti}$, we construct the optimistic expected revenue for the assortment $S$, defined as follows:
\begin{align}
    \tilde{R}_{t}(S) 
    := \frac{\sum_{i \in S} \exp( \alpha_{ti} ) r_{ti} }{v_0 + \sum_{j \in S} \exp(\alpha_{tj})}, \label{eq:opt_revenue}
\end{align}
where $r_{ti} = 1$.
Then, we offer the set $S_t$ that maximizes the optimistic expected revenue,  $S_t = \argmax_{S \in \mathcal{S}} \tilde{R}_{t}(S) $.
Given our assumption that all rewards are of unit value, the optimization problem is equivalent to selecting the $K$ items with the highest optimistic utility $\alpha_{ti}$.
Consequently, solving the optimization problem incurs a constant computational cost of $\mathcal{O}(N)$.

\begin{remark} [Comparison to~\citet{zhang2024online}]
    In \citet{zhang2024online}, the MNL choice model is outlined with a shared context  $x_t$ and distinct parameters $\wb_{1}^\star, \dots, \wb_K^\star$ for each choice.
    Conversely, our model employs a single parameter $\wb^\star$ across all choices and has varying contexts for each item in the assortment $S$, $x_{t1}, \dots x_{ti_{|S|}}$. 
    Due to this discrepancy in the choice model, directly applying Proposition 1 from~\citet{zhang2024online}, which constructs the optimistic revenue by adding bonus terms to the estimated revenue, incurs an exponential computational cost in our problem setting.
    This complexity arises because the optimistic revenue must be calculated for every possible assortment $S \in \mathcal{S}$; therefore, it is necessary to enumerate all potential assortments ($N$ choose $K$) to identify the one that maximizes the optimistic revenue
    As a result, extending the approach in~\citet{zhang2024online} to our setting is non-trivial, requiring a different analysis.
\end{remark}

We now present the regret upper bound of \AlgName{} in the uniform reward setting.

\begin{theorem} [Regret upper bound of~\AlgName{}, Uniform rewards] \label{thm:upper_bound}
 Let $\delta \in (0, 1]$ and Assumptions~\ref{assum:bounded_assumption} and~\ref{assum:kappa} hold.
    In the uniform reward setting, by setting $\eta = \frac{1}{2}\log (K+1) + 2$ and  $\lambda = 84 \sqrt{2}d \eta$, with probability at least $1-\delta$, the cumulative regret of \textup{\AlgName{}} is upper-bounded by
    \begin{align*}
        \Regret (\wb^\star )
        = \BigOTilde  \left(  \frac{\sqrt{ v_0K}}{ v_0 + K  } \cdot d\sqrt{T} 
        +  \frac{1}{\kappa} d^2 
        \right).
    \end{align*}
\end{theorem}
\textbf{Discussion of Theorem~\ref{thm:upper_bound}.}
If $T \geq \BigO (d^2 (v_0 + K)^2/(\kappa^2 v_0 K))$,
Theorem~\ref{thm:upper_bound} shows that our algorithm~\AlgName{} achieves  minimax optimal regret (up to logarithmic factor) in terms of all $d$, $T$, $K$, and even $v_0$.
To the best of our knowledge, ignoring logarithmic factors, our proposed algorithm is the first computationally efficient, minimax optimal algorithm in (adversarial) contextual MNL bandits.
When $v_0 = \Theta(1)$, which is the convention in existing MNL bandit literature~\citep{oh2019thompson, oh2021multinomial, perivier2022dynamic, agrawal2023tractable, zhang2024online},~\AlgName{} obtains $\BigOTilde(d \sqrt{T/K})$ regret.
This represents an improvement over the previous upper bound of~\citet{perivier2022dynamic}~\footnote{~\citet{perivier2022dynamic} also consider the uniform rewards ($r_{ti}=1$) with $v_0=1$.}, which is $\BigOTilde(d K \sqrt{\kappa^{\prime} T } + d^2K^4/\kappa )$, where $\kappa^{\prime}= \BigO(1/K)$, by a factor of $K$.
This improvement can be attributed to two key factors: an improved, constant, self-concordant-like property of the loss function (Proposition~\ref{prop:self_concordant}) and a $K$-free elliptical potential lemma (Lemma~\ref{lemma:elliptical}).
Furthermore, by employing an improved bound for the second derivative of the revenue (Lemma~\ref{lemma:revenue_second_pd}), we achieve an enhancement in the regret for the second term,  $d^2/\kappa $, by a factor of $K^4$, in comparison to~\citet{perivier2022dynamic}.
Unless $v_0 = \Theta(K)$, Theorem~\ref{thm:upper_bound} indicates that the regret decreases as the assortment size $K$ increases.
To the best of our knowledge, this is the first algorithm in MNL bandits to show that increasing $K$ results in a reduction in regret.
Moreover, when reduced to the logistic bandit, i.e., $K=1$, $r_{t1}=1$, and $v_0=1$, our algorithm can also achieve a regret of $\BigOTilde(d\sqrt{\kappa T})$ by Corollary 1 in~\citet{zhang2024online}, which is consistent with the results in~\citet{abeille2021instance, faury2022jointly}.
The proof is deferred to Appendix~\ref{app_sec:proof_of_thm_upper}.

\begin{remark} [Efficiency of~\AlgName{}] \label{remark:upper_computation_cost}
The proposed algorithm is computationally efficient in both parameter updates and assortment selections. 
Since we employ online parameter estimation, akin  to~\citet{zhang2024online}, our algorithm demonstrates computational efficiency in parameter estimation, incurring only incurring $\mathcal{O}(Kd^3)$ computation cost and $\mathcal{O}(d^2)$ storage cost, which are completely independent of $t$.
Furthermore, a naive approach to selecting the optimistic assortment requires enumerating all possible ($N$ choose $K$) assortments, resulting in  exponential computational cost~\citep{chen2020dynamic}.
However, by constructing the optimistic expected revenue according to Equation~\eqref{eq:opt_revenue} (inspired by~\citet{oh2021multinomial}), our algorithm needs only $\BigO(N)$ computational cost.
\end{remark}

\subsection{Regret Upper \& Lower Bounds under Non-Uniform Rewards}
\label{subsec:main_non_uniform}
In this subsection, we propose regret upper and lower bounds in the non-uniform reward setting.
In this scenario, the sizes of both the chosen assortment $S_t$, and the optimal assortment, $S_t^\star$ are not fixed at $K$.
Therefore, we cannot guarantee an improvement in regret even as $K$ increases.

We first prove the regret lower bound in the non-uniform reward setting.
\begin{theorem} [Regret lower bound, Non-uniform rewards]
    \label{theorem:lower_bound_non_uniform}
    Suppose Assumption~\ref{assum:bounded_assumption} holds and
    $T \geq C \cdot d^4 (v_0 + 1) / 1$ for some constant $C>0$.
    Let $d$ be divisible by $4$ and $v_0=\Theta(1)$.
    Then, under the non-uniform reward setting, for any policy $\pi$, there exists an instance such that the worst-case expected regret of $\pi$ is lower bounded as follows:
    \begin{align*}
        \sup_{\wb} \EE^\pi_{\wb} \left[ \Regret(\wb) \right] 
        = \Omega \left(  d\sqrt{T} \right).
    \end{align*}
\end{theorem}
\textbf{Discussion of Theorem~\ref{theorem:lower_bound_non_uniform}.}
In contrast to Theorem~\ref{thm:lower_bound}, which considers uniform rewards, the regret lower bound is independent of the assortment size $K$. 
Note that Theorem~\ref{theorem:lower_bound_non_uniform} does not claim that non-uniform rewards inherently make the problem more difficult. 
Rather, it implies that there exists an instance with \textit{adversarial} non-uniform rewards, where regret does not improve even with an increase in $K$.
Moreover, the assumption that $v_0=\Theta(1)$ is common in the existing literature on contextual MNL bandits~\citep{oh2019thompson, oh2021multinomial, perivier2022dynamic, agrawal2023tractable, zhang2024online} (refer Appendix~\ref{app_subsec:outside}).
To the best of our knowledge, this is the first established lower bound for non-uniform rewards in MNL bandits, even in the non-contextual setting.
The proof is deferred to Appendix~\ref{app_sec:proof_for_lower_non_uniform}.

We also prove a matching upper bound up to logarithmic factors.
The algorithm~\AlgName{} is also applicable in the case of non-uniform rewards. 
However, because the optimistic expected revenue $\tilde{R}_t(S)$ is no longer an increasing function of $\alpha_{ti}$, optimizing for $S_t = \argmax_{S \in \mathcal{S}} \tilde{R}_t (S)$ no longer equates to simply selecting the top $K$ items with the highest optimistic utility.
Instead, we employ assortment optimization methods introduced in~\citet{rusmevichientong2010dynamic, davis2014assortment}, which are efficient polynomial-time (independent of $t$)~\footnote{An interior point method would generally solve the problem with a computational complexity of $\BigO(N^{3.5})$.} algorithms available for solving this optimization problem.
Therefore, our algorithm is also computationally efficient under non-uniform rewards.

\begin{theorem} [Regret upper bound of~\AlgName{}, Non-uniform rewards]
    \label{theorem:upper_bound_non_uniform}
    Under the same assumptions and parameter settings as Theorem~\ref{thm:upper_bound}, if the rewards are non-uniform and $v_0=\Theta(1)$, then with a probability at least $1-\delta$, the cumulative regret of \textup{\AlgName{}} is upper-bounded by
    \begin{align*}
        \Regret (\wb^\star )
        &= \BigOTilde \left( d\sqrt{T} + \frac{1}{\kappa}d^2 \right).
    \end{align*}
\end{theorem}
\textbf{Discussion of Theorem~\ref{theorem:upper_bound_non_uniform}.}
If $T \geq \BigO (d^2 / \kappa^2)$, our algorithm achieves a regret of $\BigOTilde(d \sqrt{T})$ when the reward for each item is non-uniform, demonstrating that~\AlgName{} is minimax optimal up to a logarithmic factor. 
Recall that we relax the bounded assumption on the reward compared to~\citet{zhang2024online} (refer Assumption~\ref{assum:bounded_assumption}); thus, we allow the sum of the squared rewards in the assortment to scale with $K$.
Consequently, we need a novel approach to achieve the regret that does not scale with $K$.
To this end, we \textit{centralize} the features, 
i.e., $\tilde{x}_{ti} = x_{ti} - \EE_{ j \sim p_t(\cdot | S_t, \wb_{t+1}) }[x_{tj}]$,
and propose a novel elliptical potential lemma for them, as detailed in Lemma~\ref{lemma:elliptical_x_tilde}.
Note that our algorithm is capable of achieving $1/\kappa$-free regret (in the leading term) under both uniform and non-uniform rewards. 
In contrast, the algorithm in~\citet{perivier2022dynamic} is limited to achieving this only in the uniform reward setting.
Furthermore, compared to the regret bound in~\citet{chen2020dynamic}, which is $\BigOTilde(d\sqrt{T})$, our regret bounds has the same order of regret with theirs.
However, their algorithm is computationally intractable as it requires enumerating all possible assortments, whereas our algorithm incurs only a constant computational cost per round.
The proof is deferred to Appendix~\ref{app_sec:upper_bound_non_uniform}.

\section{Instance-Dependent Bounds}
\label{sec:problem_dependent_bound}
In this section, we show that instance-dependent upper and lower bounds are also achievable under uniform rewards.
We define the degree of non-linearity for the optimal assortment $S^\star_t$ at round $t$ under the true parameter $\wb^\star$ as 
$\kappa^\star_t := \sum_{i \in S^\star_t} p_t(i | S^\star_t, \wb^\star) p_t(0 | S^\star_t, \wb^\star)$.
We first establish the instance-dependent lower bound under uniform rewards.
\begin{proposition} [Instance-dependent regret lower bound, Uniform rewards]
    \label{prop:lower_instance}
    Under the same conditions as Theorem~\ref{thm:lower_bound},
    for any policy $\pi$ and for $T \geq  d^2/ \kappa$, there exists a worst-case problem instance such that the worst-case expected regret of $\pi$ is lower bounded as follows:
    \begin{align*}
        \sup_{\wb} \EE^\pi_{\wb} \left[ \Regret(\wb) \right] 
        = \Omega \left(  d\sqrt{ \sum_{t=1}^T \kappa^\star_t  }  \right).
    \end{align*}
\end{proposition}
The proof is deferred to Appendix~\ref{app_sec:proof_lower_instance}.
We also provide the matching upper bound.
\begin{proposition} [Instance-dependent regret upper bound of~\AlgName{}, Uniform rewards]
    \label{prop:upper_instance}
    Under the same assumptions, parameter settings, and reward structure as Theorem~\ref{thm:upper_bound}, with a probability  at least $1-\delta$, the cumulative regret of \textup{\AlgName{}} is upper-bounded by
    \begin{align*}
         \Regret (\wb^\star )
        &= \BigOTilde \left( d\sqrt{  \sum_{t=1}^T \kappa^\star_t  } + \frac{1}{\kappa}d^2 \right).
    \end{align*}
\end{proposition}
The proof is deferred to Appendix~\ref{app_sec:proof_upper_instance}.
For sufficiently large $T$, the regret upper bound (Proposition~\ref{prop:upper_instance}) matches the regret lower bound (Proposition~\ref{prop:lower_instance}), up to logarithmic factor. 
To the best of our knowledge, these are the first minimax instance-dependent regret bounds under uniform rewards.
Note that, in the worst case, $ \kappa^\star_t = \BigOTilde (\sqrt{v_0 K} / (v_0 + K)  )$, which indicates that these results provide a strict improvement over the worst-case bounds given in Theorems~\ref{thm:lower_bound} and~\ref{thm:upper_bound}.

Some readers may expect instance-dependent regret bounds for non-uniform rewards as well. 
Unfortunately, we were unable to establish these.
Recall that $\kappa^\star_t$ represents the degree of non-linearity for the optimal assortment $S^\star_t$.
However, in the proofs for regret bounds, we encounter terms associated with the chosen assortment  $S_t$, such as $\sum_{i \in S^\star_t} p_t(i | S_t, \wb^\star) p_t(0 | S_t, \wb^\star)$.
To address this, we use the mean value theorem-based analysis (Lemma~\ref{lemma:kappa_star_bound}) to replace this quantity with $\kappa^\star_t$.
Under non-uniform rewards, however, the mean value theorem does not apply because the sizes and rewards of $S^\star_t$ and $S_t$ may differ.
Addressing this problem would be an interesting direction for future research.

\section{Numerical Experiments}
\label{sec:experiments}
\begin{figure*}[t]
    \centering
    \begin{subfigure}[b]{0.245\textwidth}
        \includegraphics[width=\textwidth, trim=0mm 0mm 11mm 5mm, clip]{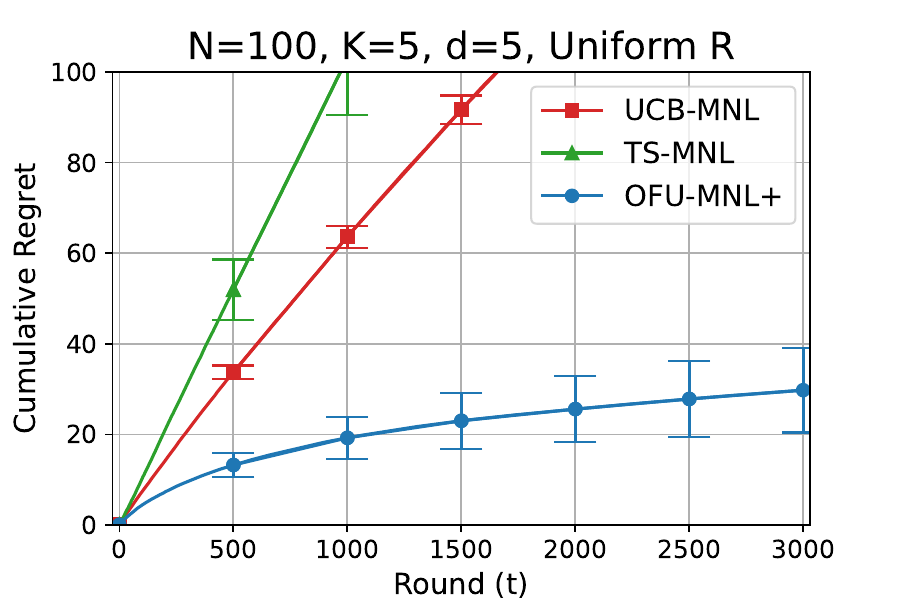}
        \label{fig:N=100_K=5_d=5_dist=0}
    \end{subfigure}
    \begin{subfigure}[b]{0.245\textwidth}
        \includegraphics[width=\textwidth, trim=0mm 0mm 11mm 5mm, clip]{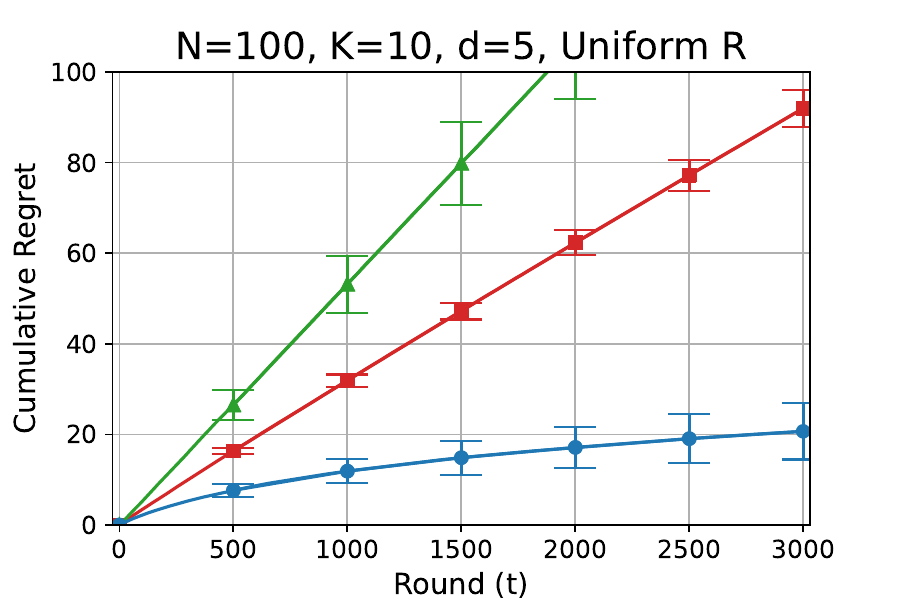}
        \label{fig:N=100_K=10_d=5_dist=0}
    \end{subfigure}
    \begin{subfigure}[b]{0.245\textwidth}
        \includegraphics[width=\textwidth, trim=0mm 0mm 11mm 5mm, clip]{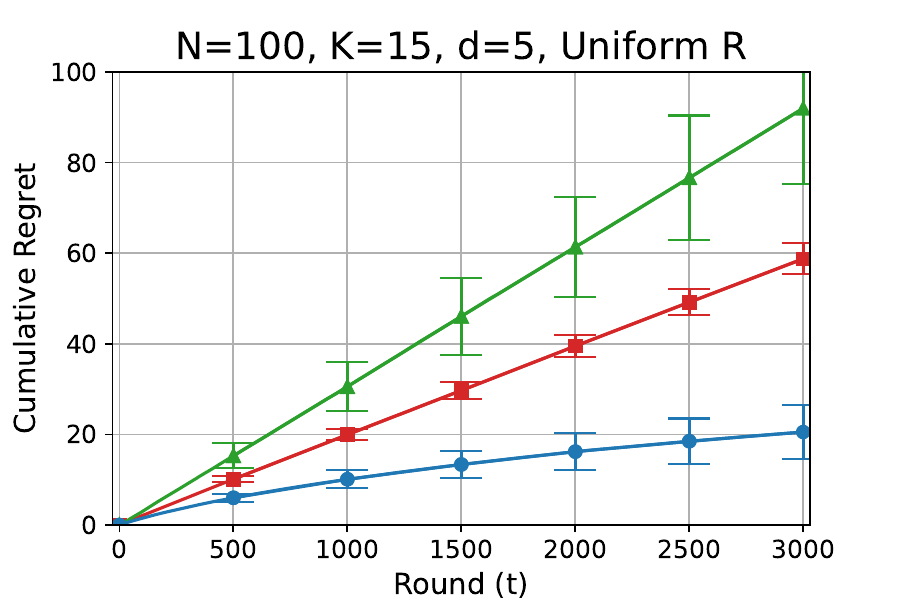}
        \label{fig:N=100_K=15_d=5_dist=0}
    \end{subfigure}
    \begin{subfigure}[b]{0.245\textwidth}
        \includegraphics[width=\textwidth, trim=0mm 0mm 11mm 5mm, clip]{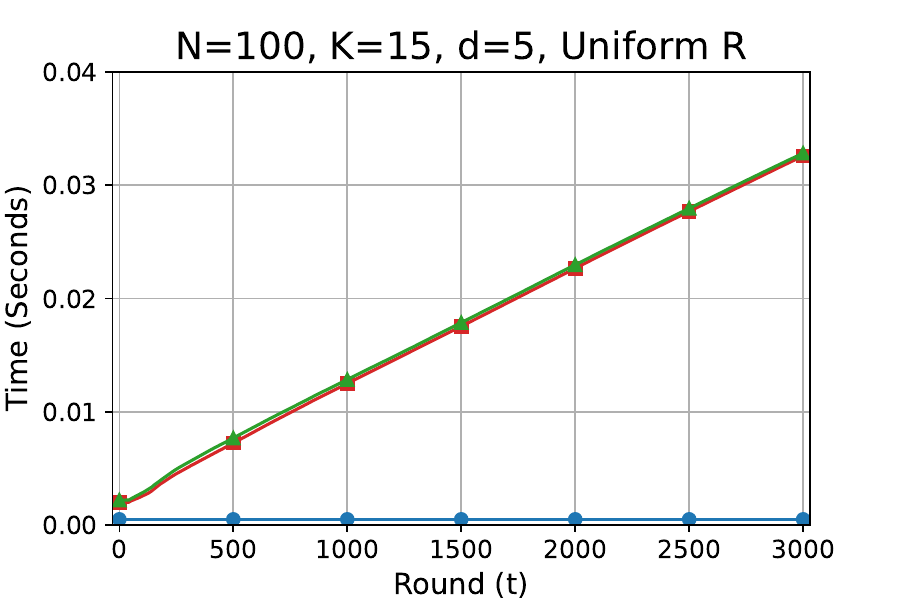}
        \label{fig:N=100_K=15_d=5_dist=0_r_main}
    \end{subfigure}

    \begin{subfigure}[b]{0.245\textwidth}
        \includegraphics[width=\textwidth, trim=0mm 0mm 11mm 5mm, clip]{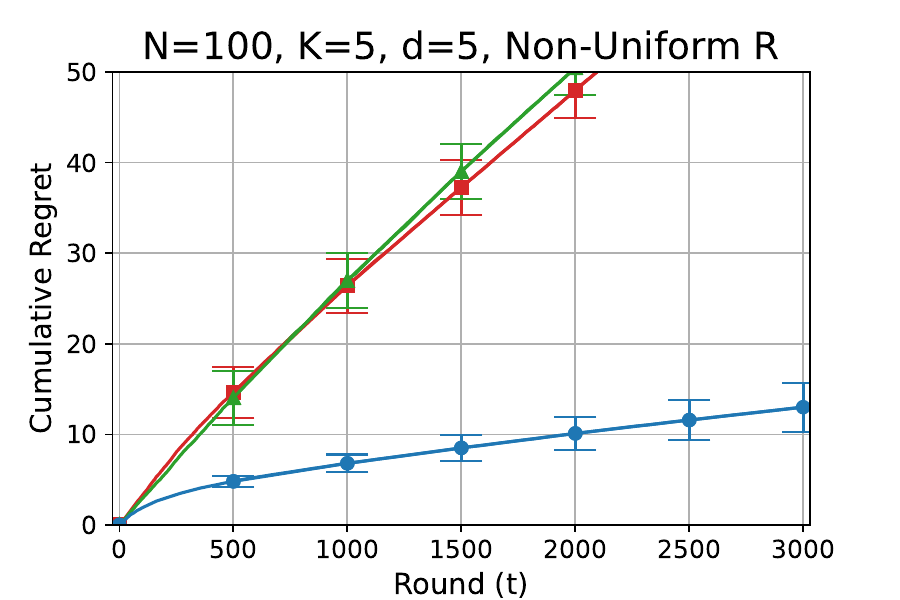}
        \label{fig:N=100_K=5_d=5_dist=0_nu}
    \end{subfigure}
    \begin{subfigure}[b]{0.245\textwidth}
        \includegraphics[width=\textwidth, trim=0mm 0mm 11mm 5mm, clip]{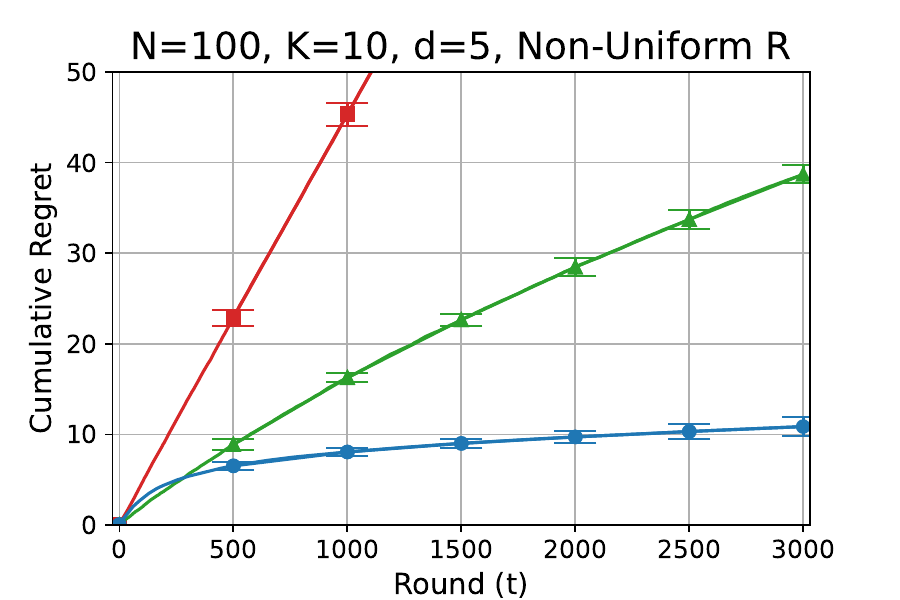}
        \label{fig:N=100_K=10_d=5_dist=0_nu}
    \end{subfigure}
    \begin{subfigure}[b]{0.245\textwidth}
        \includegraphics[width=\textwidth, trim=0mm 0mm 11mm 5mm, clip]{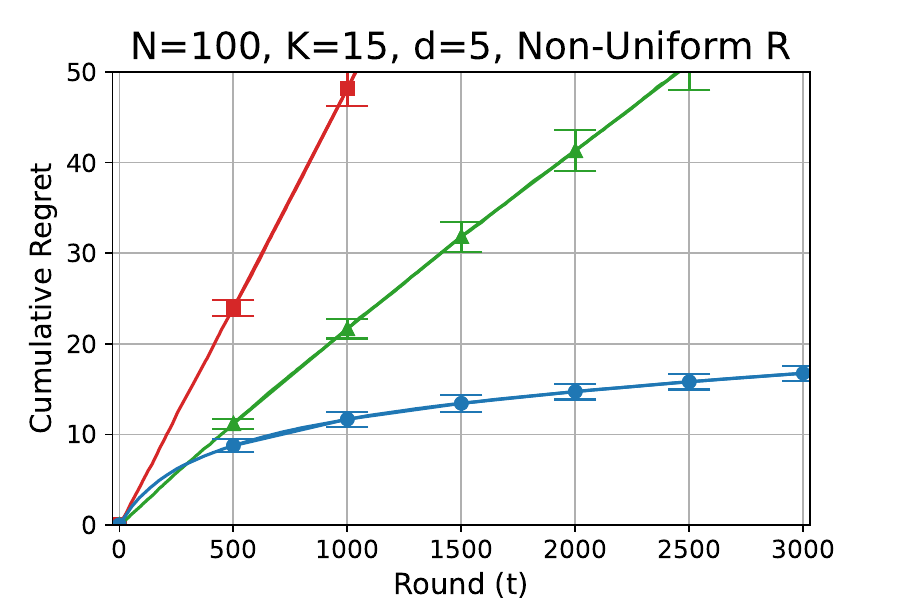}
        \label{fig:N=100_K=15_d=5_dist=0_nu}
    \end{subfigure}
    \begin{subfigure}[b]{0.245\textwidth}
        \includegraphics[width=\textwidth, trim=0mm 0mm 11mm 5mm, clip]{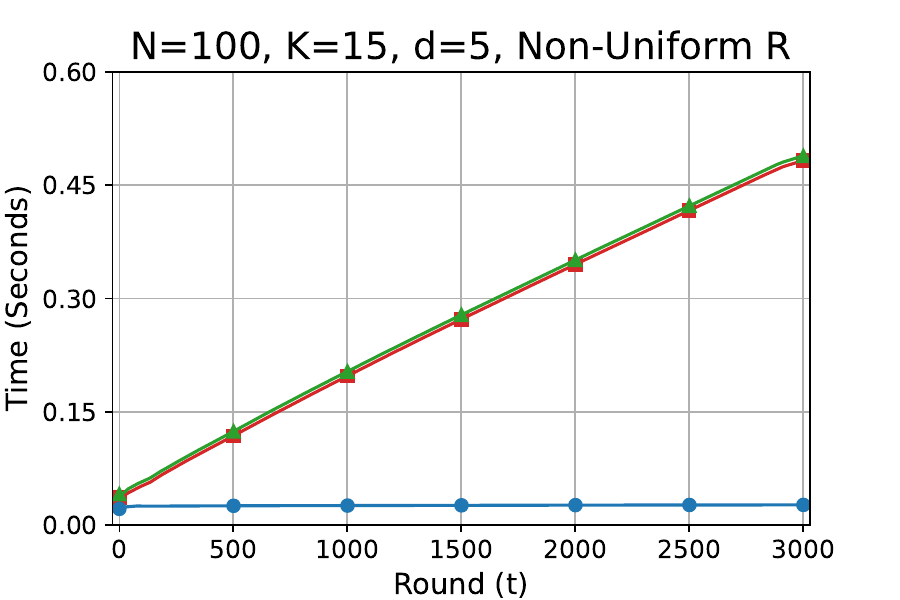}
        \label{fig:N=100_K=15_d=5_dist=0_r_nu_main}
    \end{subfigure}
    \caption{Cumulative regret (left three, $K=5,10,15$) and runtime per round (rightmost one, $K=15$) under uniform rewards (first row) and non-uniform rewards (second row) with $v_0 = 1$.}
    \label{fig:regret_main}
\end{figure*}
In this section, we empirically evaluate the performance of our algorithm~\AlgName{}.
We measure cumulative regret over $T=3000$ rounds.
For each experimental setup, we run the algorithms across $20$ independent instances and report the average performance.
In each instance, the underlying parameter $\wb^\star$ is randomly sampled from a $d$-dimensional uniform distribution, where each element of $\wb^\star$ lies within the range $[-1/\sqrt{d},1/\sqrt{d}]$ and is not known to the algorithms.
Additionally, the context features $x_{ti}$ are drawn from a $d$-dimensional multivariate Gaussian distribution, with each element of $x_{ti}$ clipped to the range $[-1/\sqrt{d},1/\sqrt{d}]$.
This setup ensures compliance with Assumption~\ref{assum:bounded_assumption}.
In the uniform reward setting (first row of Figure~\ref{fig:regret_main}), the combinatorial optimization step to choose the assortment reduces to sorting items by their utility estimate. 
In the non-uniform reward setting (second row of Figure~\ref{fig:regret_main}), the rewards are sampled from a uniform distribution in each round, i.e., $r_{ti} \sim \operatorname{Unif}(0,1)$.
Refer Appendix~\ref{app_sec:experimat_details} for more details.

We compare the performance of~\AlgName{} with those of the practical and state-of-the-art algorithms: the Upper Confidence Bound-based algorithm, \texttt{UCB-MNL}~\citep{oh2019thompson}, and the Thompson Sampling-based algorithm, \texttt{TS-MNL}~\citep{oh2019thompson}.
Figure~\ref{fig:regret_main} demonstrates that our algorithm significantly outperforms other baseline algorithms. 
In the uniform reward setting, as $K$ increases, the cumulative regrets of all algorithms tend to decrease. 
In contrast, this trend is not observed in the non-uniform reward setting. 
Furthermore, the results also show that our algorithm maintains a constant computation cost per round, while the other algorithms exhibit a linear dependence on $t$.
In Appendix~\ref{app_sec:experimat_details}, we present the additional runtime curves (Figure~\ref{fig:runtime}) as well as the regret curves of the other configuration where $v_0 = \Theta(K)$ (Figure~\ref{fig:varying_v0}).
All of these empirical results align with our theoretical results.

\section{Conclusion}
\label{sec:conclusion}
In this paper, we propose minimax optimal lower and upper bounds for both uniform and non-uniform reward settings.
We propose a computationally efficient algorithm,~\AlgName{}, that achieves a regret of $\BigOTilde(d\sqrt{\smash[b]{T/K}})$ under uniform rewards and $\BigOTilde(d \sqrt{T})$ under non-uniform rewards.
We also prove matching lower bounds of $\Omega(d\sqrt{\smash[b]{T/K}})$ and $\Omega(d \sqrt{T})$ for each setting, respectively.
Moreover, our empirical results support our theoretical findings, demonstrating that~\AlgName{} is not only provably but also experimentally efficient.

\clearpage
\section*{Acknowledgements}
This work was supported by the National Research Foundation of Korea(NRF) grant funded by the Korea government(MSIT) (No. 2022R1C1C1006859, 2022R1A4A1030579, and RS-2023-00222663) and by AI-Bio Research Grant through Seoul National University.

\bibliography{main_bib}
\bibliographystyle{plainnat}

\newpage
\appendix
\onecolumn
\counterwithin{table}{section}
\counterwithin{lemma}{section}
\counterwithin{corollary}{section}
\counterwithin{theorem}{section}
\counterwithin{algorithm}{section}
\counterwithin{assumption}{section}
\counterwithin{figure}{section}
\counterwithin{equation}{section}
\counterwithin{condition}{section}
\counterwithin{remark}{section}
\counterwithin{definition}{section}
\counterwithin{proposition}{section}

\addcontentsline{toc}{section}{Appendix} 
\part{Appendix} 
\parttoc 
\section{Further Related Work}
\label{app_sec:further_related}
In this section, we discuss additional related works that complement Section~\ref{sec:Related}.
For simplicity, we consider only the dependence on the number of rounds $t$ for a computation cost in big-$\BigO$ notation.

\textbf{Logistic Bandits.}
The logistic bandit model~\citep{filippi2010parametric, faury2020improved, abeille2021instance, faury2022jointly} focuses on environments with \textit{binary} rewards and explores the impact of non-linearity on the exploration-exploitation trade-off for parametrized bandits. 
The main research interest has been the algorithms' dependence on the degree of non-linearity $\kappa$, which can grow exponentially in terms of the diameter of the decision domain $\mathcal{W}$.
\citet{zhang2016online} introduced the first efficient algorithm for binary logistic bandits with a $\BigO(1)$ computation cost, achieving a regret of $\BigOTilde(d\sqrt{T}/\kappa)$.
\citet{faury2020improved} enhanced the regret to $\BigOTilde(d\sqrt{\smash[b]{T/\kappa}})$ with a $\BigO(t)$ computation cost.
However, their regret bounds still suffered from a harmful dependence on $1/\kappa$. 
\citet{abeille2021instance} addressed this by achieving the tightest regret upper bound of $\BigOTilde(d\sqrt{\kappa T})$ with a $\BigO( t)$ computation cost, while \citet{faury2022jointly} achieved the same regret with an improved computation cost of  $\BigO(\log t)$.
More recently,~\citet{zhang2024online} proposed a jointly efficient algorithm that achieves the optimal regret with a constant $\BigO(1)$ computation cost.
Note that the logistic bandit is a special case of the multinomial logistic (MNL) bandit. 
When the maximum assortment size is one ($K=1$), rewards are uniform ($r_{t1}=1$), and the attraction parameter for the outside option is one $v_0=1$, the MNL bandit reduces to the logistic bandit.
In this logistic bandit setting, our proposed algorithm, \AlgName{}, can achieve a regret upper bound of $\BigOTilde(d\sqrt{\kappa T})$ with a constant $\BigO(1)$ computation cost, consistent with the result in~\citet{zhang2024online}.

\textbf{Multinomial Logistic (MNL) Bandits.}
There are two main approaches to multinomial logistic (MNL) bandits: the \textit{multiple-parameter} choice model and the \textit{single-parameter} choice model.
In the multiple-parameter choice model, the learner estimates parameters for each choice in the assortment ($\wb_1^\star, \dots, \wb_K^\star$) with a shared context $x_t$. 
In this setting,~\citet{amani2021ucb} proposed a feasible algorithm that achieves a regret upper bound of $\BigOTilde(dK\sqrt{\kappa T})$ with a $\BigO(t)$ computation cost.
They also proposed an intractable algorithm that achieves an improved regret of $\BigOTilde(dK^{3/2}\sqrt{ T})$.
\citet{zhang2024online} introduced a computationally and statistically efficient algorithm that obtains a regret of $\BigOTilde(dK\sqrt{ T})$.
Recently,~\citet{lee2024improved} further improved the regret by a factor of $\sqrt{K}$, achieving $\BigOTilde(d \sqrt{KT})$ regret.
In the multiple-parameter case, the regret's dependence on $K$ is unavoidable since the number of unknown parameters depends on $K$.

On the other hand, the single-parameter choice model, closely related to ours, shares the parameter $\wb^\star$ cross the choices, with varying contexts for each choice. 
The learner offers a set of items $S_t$, with $|S_t|\leq K$ at each round.
This setting involves a combinatorial optimization to choose the assortment $S_t$, making it more challenging to devise a tractable algorithm.
As extensively discussed in Section~\ref{sec:Related}, no previous studies have definitively confirmed whether the existing lower or upper bounds are tight. 
As shown in Table~\ref{tab:regrets}, many studies have presented their results in inconsistent settings with varying reward structures and values of $v_0$, adding to the ambiguity about the bounds’ optimality. 
In this paper, we address these issues by bridging the gap between the lower and upper bounds of regret through a careful categorization of the settings. 
We propose an algorithm that is both provably optimal, up to logarithmic factors, and computationally efficient, significantly enhancing the theoretical and practical understanding of MNL bandits.

\textbf{Generalized Linear Bandits.}
In generalized linear bandits~\citep{filippi2010parametric, jun2017scalable, li2017provably, abeille2017Linear, kveton2020randomized, kazerouni2021best, kim2023double, lee2024unified}, the expected rewards are modeled using a generalized linear model. 
These problems generalize logistic bandits by incorporating a general exponential family link function instead of the logistic link function. 
The algorithms proposed for generalized linear bandits also exhibit a dependence on the nonlinear term $\kappa$. 
However, our problem setting (single-parameter MNL bandits) considers a more complex state space where multiple arms are pulled simultaneously.

\textbf{Combinatorial Bandits.}
Another related stream of literature is combinatorial bandits~\citep{chen2013combinatorial, qin2014contextual, kveton2015tight, zong2016cascading, pmlr-v117-rejwan20a, liu2023contextual}, particularly top-$k$ combinatorial bandits~\citep{pmlr-v117-rejwan20a}. 
In top-$k$ combinatorial bandits, the decision set includes all subsets of size $k$ out of $n$ arms, and the reward for each action is the sum of the rewards of the $k$ selected arms. 
In this framework, the rewards are assumed to be independent of the entire set of arms played in round $t$. 
In contrast, in our setting, the reward for each individual arm depends on the whole set of arms played.

Recently,~\citet{choi2024cascading, cao2024tiered}  have considered the cascading assortment bandit problem, which encompasses the MNL bandit problem as a special case where the cascading length is $1$.
However, these works do not strictly encompass our results.
\citet{choi2024cascading} only  consider uniform rewards, and achieve a regret upper bound of $\BigOTilde(d\sqrt{T})$, 
which avoids dependence on both the cascading length and $\kappa$.
When the cascading length is $1$, our result (Theorem 2) improves upon theirs by a factor of $\sqrt{K}$.
Moreover, their computation cost per round is $\BigO(t)$ since they employ MLE to estimate the parameter.
\citet{cao2024tiered} consider non-uniform rewards, and achieve a regret upper bound of $\tilde{\mathcal{O}}(h^2d\sqrt{MT})$, where $M$ is the cascading length and $h \geq \frac{p(i 
| S, \mathbf{w})}{p(i | S, \mathbf{w'})}$ for all $\mathbf{w}, \mathbf{w}' \in \mathcal{W}$, $S\in \mathcal{S}$, and $i \in S \cup \{0\}$. 
However, their regret bound still suffers from a harmful dependence on $h^2$, which can be exponentially large.

\section{Notation} 
\label{app_sec:notation}
We denote $T$ as the total number of rounds and $t \in [T]$  as the current round.
We denote $N$ as the total number of items, $K$ as the maximum size of assortments, and $d$ as the dimension of feature vectors.

For notational simplicity, we define the loss function in two different forms throughout the proof:
\begin{align*}
    \ell_t(\wb) &= - \sum_{i \in S_t} y_{ti} \log p_t(i | S_t, \wb)
    =- \sum_{i \in S_t} y_{ti} \log \left( \frac{\exp(x_{ti}^\top  \wb )}{ v_0 \!+\!\sum_{j \in S_t }\exp( x_{tj}^\top \wb )}\right),
    \\
    \ell(\zb_t, \yb_t) &= - \sum_{i \in S_t} y_{ti} \log \left( \frac{\exp( z_{ti} )}{ v_0 \!+\!\sum_{j \in S_t }\exp( z_{tj} )}\right),
\end{align*}
where  $z_{ti} = x_{ti}^\top \wb$, $\zb_t = (z_{ti})_{i \in S_t} \in \RR^{|S_t|}$, and $\yb_t = (y_{ti})_{i \in S_t} \in \RR^{|S_t|}$.
Thus, $\ell_t(\wb) = \ell(\zb_t, \yb_t)$.

We offer a Table~\ref{table_symbols} for convenient reference.
\begin{table}[htp!]
\centering
    \caption{Symbols}
    \label{table_symbols}
    \begin{tabular}{ll}
         \toprule
         $x_{ti}$       &      feature vector for item $i$ given at round $t$\\[0.1cm]
         $r_{ti}$       &      reward for item $i$ given at round $t$\\[0.1cm]
         $S_t$       &      assortment chosen by an algorithm at round $t$ \\[0.1cm]   
         $0$       &      outside option \\[0.1cm]   
         $y_{ti}$       &      choice response for each item $i \in S_t \cup \{0\}$ at round $t$ \\[0.1cm] 
         $ R_{t}(S, \wb^\star)$       &  $:=  \sum_{i \in S} p_t(i | S, \wb^\star) r_{ti} $,    expected revenue of the assortment $S$ at round $t$\\[0.1cm]
         $\ell_t(\wb) $       &   $:= - \sum_{i \in S_t} y_{ti} \log \left( \frac{\exp(x_{ti}^\top  \wb )}{ v_0 \!+\!\sum_{j \in S_t }\exp( x_{tj}^\top \wb )}\right)$, loss function at round $t$ \\[0.1cm]
         $\ell(\zb_t, \yb_t)$       &   $:=  - \sum_{i \in S_t} y_{ti} \log \left( \frac{\exp( z_{ti} )}{ v_0 \!+\!\sum_{j \in S_t }\exp( z_{tj} )}\right)$,  loss function at round $t$, $z_{ti} = x_{ti}^\top \wb$\\[0.1cm]
         $\lambda$       &    regularization parameter \\[0.1cm] 
         $\Gcal_t(\wb)$       &    $:= \sum_{i \in S_t} p_t(i | S_t, \wb) x_{ti} x_{ti}^\top 
    -  \sum_{i \in S_t}  \sum_{j \in S_t} p_t(i | S_t, \wb) p_t(j | S_t, \wb) x_{ti} x_{tj}^\top$ \\[0.1cm] 
         $H_t$       &    $:= \lambda \Ib_d + \sum_{s=1}^{t-1} \Gcal_s(\wb_{s+1})$ \\[0.1cm] 
         $\tilde{H}_{t}$       &  $:= H_t + \eta \Gcal_t(\wb_t)$ \\[0.1cm] 
         $\alpha_{ti}$       &    $:= x_{ti}^\top \wb_t + \beta_t(\delta) \| x_{ti} \|_{H_t^{-1}}$,  optimistic utility for item $i$ at round $t$ \\[0.1cm] 
         $\beta_t(\delta)$       &    $:= \BigO \left( \sqrt{d} \log t \log K \right)$,  confidence radius at round $t$ \\[0.1cm] 
         $\tilde{R}_{t}(S)$       &  $:= \frac{\sum_{i \in S} \exp( \alpha_{ti} ) r_{ti} }{v_0 + \sum_{j \in S} \exp(\alpha_{tj})}$,    optimistic expected revenue for the assortment $S$ at round $t$ \\[0.1cm] 
         \bottomrule
    \end{tabular}
\end{table}

\section{Properties of MNL function} \label{app_sec:properties}
In this section, we present key properties of the MNL function and its associated loss, which are used throughout the paper.
\subsection{Attraction parameter for Outside Option: \texorpdfstring{$v_0=\Theta(1)$}{} is Common in Contextual MNL Bandits} \label{app_subsec:outside}
In this subsection, we explain why the assumption that $v_0=\Theta(1)$ is made without loss of generality.
Let the original feature vectors be $x'_{ti} \in \RR^d$ for every item $i \in [N]$.
Suppose that a context for the outside option $x_{t0}'$ is given and the probability of choosing any item $i \in S_t \cup \{0\}$ is defined as 
\begin{align*}
    p_t(i | S_t, \wb^\star) 
    = \frac{\exp((x'_{ti})^\top  \wb^\star )}{\sum_{j \in S_t \cup \{0\} }\exp( (x'_{tj})^\top \wb^\star )}.
\end{align*}
Then, by dividing the denominator and numerator by $\exp\left( (x'_{t0})^\top \wb^\star \right)$, and defining $x_{ti} := x'_{ti} - x'_{t0}$, we obtain the MNL probability in the form presented in Equation~\eqref{eq:mnl_model} with $v_0 = \exp(0) = 1$.
Note that this division does not change the probability.
Therefore, $v_0 = \Theta(1)$ is natural and common in contextual MNL bandit literature.

\subsection{Self-concordant-like Function} \label{app_subsec:self_concordant}
\begin{definition}[Self-concordant-like function,~\citealt{tran2015composite}]
\label{def:self-concordant-like}
A convex function $f \in \mathcal{C}^3(\RR^m)$ is $M$-self-concordant-like  function with constant $M$ if:
    \begin{align*}
        |\phi^{\prime\prime\prime} (s) | \leq M \|\bb \|_2 \phi^{\prime\prime}(s) .
    \end{align*}
    for $s \in \RR$ and $M > 0$, where $\phi(s) := f(\ab + s\bb)$ for any  $\ab, \bb \in \RR^m$.
\end{definition}
Then, the MNL loss defined in Equation~\eqref{eq:loss} is $3\sqrt{2}$-self-concordant-like function.
\begin{proposition} \label{prop:self_concordant}
    For any $t \in [T]$, the multinomial logistic loss $\ell_{t}(\wb)$, defined in Equation~\eqref{eq:loss}, is  $3\sqrt{2}$-self-concordant-like.
\end{proposition}
\begin{proof}   
    Consider the function $\phi(s) := \log \left( \sum_{i =0}^n e^{a_i s + b_i} \right)$, where $\ab = [a_0, \dots, a_{n}]^\top \in \RR^{n+1}$ and  $\bb = [b_0, \dots, b_{n}]^\top \in \RR^{n+1}$.
    Then, by simple calculus, we have
    \begin{align*}
        \phi^{\prime\prime} (s)
        = \frac{\sum_{i<j} (a_i - a_j)^2 e^{a_is + b_i} e^{a_js + b_j} }{\left(\sum_{i=0}^n e^{a_is + b_i} \right)^2}
        \geq 0,
    \end{align*}
    and
    \begin{align*}
        \phi^{\prime\prime\prime} (s)
        = \frac{\sum_{i<j} (a_i - a_j)^2 e^{a_is + b_i} e^{a_js + b_j} 
        \left[ \sum_{k=0}^n (a_i + a_j -2a_k)e^{a_ks + b_k} \right]}{\left(\sum_{i=0}^n e^{a_is + b_i} \right)^3}.
        \numberthis \label{eq:self_concordant_third_derivative}
    \end{align*}
    Note that for all $i,j,k = 0, \dots, n$, 
    \begin{align}
        |a_i + a_j -2a_k| \leq \sqrt{6}\sqrt{a_i^2 + a_j^2 + a_k^2} \leq 3\sqrt{2} \max_{i=0, \dots, n} | a_i|. 
        \label{eq:self_concordant_improved_norm}
    \end{align}
    Therefore, we have
    \begin{align*}
        \left| \sum_{k=0}^n (a_i + a_j -2a_k)e^{a_ks + b_k} \right|
        \leq  \sum_{k=0}^n \left|a_i + a_j -2a_k\right|e^{a_ks + b_k} 
        \leq 3\sqrt{2} \max_{i=0, \dots, n} | a_i | \sum_{k=0}^n e^{a_ks + b_k}.
        \numberthis \label{eq:self_concordant_improved_norm2}
    \end{align*}
    Plugging in Equation~\eqref{eq:self_concordant_improved_norm2} into Equation~\eqref{eq:self_concordant_third_derivative}, we obtain
    \begin{align}
        \phi^{\prime\prime\prime}(s)
        \leq 3\sqrt{2} \max_{i=0, \dots, n} |a_i | \phi^{\prime\prime}(s).
        \label{eq:self_concordant_like_improved}
    \end{align}
    Now, we are ready to prove the proposition.
    For any $t \in [T]$, let $n=|S_t|$ and $c_1 = x_{ti_1}, c_2 = x_{ti_2}, \dots, c_n = x_{ti_n}$.
    Define a function  $f \in \mathcal{C}^3: \RR^d \rightarrow\ \RR$ as $f(\thetab) := \log \left( v_0  + \sum_{i=1}^n e^{c_i^\top \thetab } \right)$.
    Let $\deltab  \in \RR^d$ and let $f(\thetab + s \deltab) = \log \left( v_0  + \sum_{i=1}^n e^{c_i^\top \thetab + s  c_i^\top \deltab}  \right)  = \log \left( 
    \sum_{i =0}^n  e^{a_i s +  b_i}  \right) = \phi(s)$, where $a_i = c_i^\top \deltab$, $b_i = c_i^\top \thetab$ for $i = 1, \dots, n$, and $a_i =0$ and $b_i = \log v_0$ for $i=0$.
    Then, by Equation~\eqref{eq:self_concordant_like_improved}, we get
    \begin{align*}
         |\phi^{\prime\prime\prime}(s)| 
        &\leq 3\sqrt{2} \max_{i=0, \dots, n} |a_i | \phi^{\prime \prime}(s)
        = 3\sqrt{2} \max_{i=1, \dots, n} | c_i^\top \deltab | \phi^{\prime \prime}(s)
        \\
        &\leq 3\sqrt{2} \max_{i=1, \dots, n} \| c_i \|_2 \|\deltab \|_2 \phi^{\prime \prime}(s)
        \leq 3\sqrt{2} \|\deltab \|_2 \phi^{\prime \prime}(s)
        ,
    \end{align*}
    where the last inequality holds due to Assumption~~\ref{assum:bounded_assumption} that $\|c_i\|_2 = \| x_{tj_i} \|_2 \leq 1$.
    Then, by Definition~\ref{def:self-concordant-like}, $f$ is $3\sqrt{2}$-self-concordant-like.
    Since $\ell_t$ is the sum of $f$ and a linear operator, which has third derivatives equal to zero, it follows that $\ell_t$ is also $3\sqrt{2}$-self-concordant-like function.
\end{proof}
\begin{remark} \label{remark:self_concordant}
    Contrary to the findings of~\citet{perivier2022dynamic}, which suggest that the MNL loss function $\sqrt{6K}$-self-concordant-like, our loss function is $3\sqrt{2}$-self-concordant-like.
    This yields an improved regret bound on the order of  $\mathcal{O}(\sqrt{K})$.
    The improvement arises due to a $K$-independent self-concordant-like property of $\ell_t$, as shown in Proposition~\ref{prop:self_concordant}.
    In~\citet{perivier2022dynamic}, Lemma 4 from ~\citet{tran2015composite} is used, which describes a $\sqrt{6} \|\ab\|_2$ self-concordant-like property.
    However, in the analysis of~\ref{eq:self_concordant_improved_norm}, we show that their analysis is not tight because they bound the term $\sqrt{\smash[b]{a_i^2 + a_j^2 + a_k^2}}$ by $\|\ab\|_2 = \sqrt{\sum_{i=0}^n a_i^2}$, thus making its upper bound dependent on $K$, i.e., $n= |S_t| \leq K$.
    In contrast, we bound the same term by a constant, $\max_{i=1,\dots,n} \| a_i \|_2$, which allows our loss function to exhibit a constant $3\sqrt{2}$-self-concordant-like property.
    This key difference accounts for the $\sqrt{K}$-improved regret.
\end{remark}
\begin{lemma} [Theorem 3 of~\citealt{tran2015composite}]
\label{lemma:tran_thm3}
    A convex function $\ell \in \mathcal{C}^3: \RR^d \rightarrow \RR$ is $M$-self-concordant-like if and only if for any $\vb, \ub_1, \ub_2, \ub_3 \in \RR^d$, we have
    \begin{align*}
        | \langle D^3 \ell(\vb)[\ub_1]\ub_2, \ub_3  \rangle |
        \leq M \|\ub_1 \|_2 \| \ub_2 \|_{ \nabla^2 \ell(\vb) } \| \ub_3 \|_{ \nabla^2 \ell(\vb) }.
    \end{align*}
\end{lemma}

\section{Proof of Theorem~\ref{thm:lower_bound}} 
\label{app_sec:proof_lower_bound}
In this section, we provide the proof of Theorem~\ref{thm:lower_bound}.
The proof structure is similar to the one presented in~\citet{chen2020dynamic}.
However, unlike their approach, we explicitly derive a bound that includes $v_0$. 
Furthermore, by establishing a tighter upper bound for the KL divergence (Lemma~\ref{lemma:lower_KL}),
we derive a bound that is tighter than the one provided by~\citet{chen2020dynamic}.

\subsection{Adversarial Construction and Bayes Risk}
\label{app_subsec:lower_adversarial_setting}
Let $\epsilon \in ( 0, 1/d\sqrt{d} )$ be a small positive parameter to be specified later.
For every subset $V \subseteq [d]$, we define the corresponding parameter $\wb_{V} \in \RR^d$ as $[\wb_{V}]_{j} = \epsilon$ for all $j \in V$, and $[\wb_{V}]_j = 0$ for all $j \notin V$.
Then, we consider the following parameter set
\begin{align*}
    \wb \in \mathcal{W} := \{ \wb_{V} : V \in \mathcal{V}_{d/4} \}
    := \{ \wb_{V}: V \subseteq [d], |V| = d/4 \},
\end{align*}
where $\mathcal{V}_k$ denotes the class of all subsets of $[d]$ whose size is $k$.
Moreover, note that $d/4$ is a positive integer, as $d$ is divisible by $4$ by construction.

The context vectors $x_{ti}$ are constructed to be invariant across rounds $t$. 
For each $t$ and $U \in \mathcal{V}_{d/4}$, $K$ identical context vectors~\footnote{Recall that $K$ is the maximum allowed assortment capacity.} $x_{U}$ are constructed as follows:
\begin{align*}
    [x_{U}]_{j} = 1/\sqrt{d} \quad \text{for} \,\, j \in U;
    \quad
    [x_{U}]_{j} = 0 \quad \text{for} \,\, j \notin U.
\end{align*}
For all $V, U \in \mathcal{V}_{d/4}$, it can be verified that $\wb_V$ and $x_{U}$  satisfy the requirements of a bounded assumption (Assumption~\ref{assum:bounded_assumption}) as follows:
\begin{align*}
    \| \wb_V \|_{2} \leq \sqrt{d \epsilon^2} \leq 1,
    \quad
    \| x_{U} \|_2 \leq \sqrt{d \cdot 1/d } = 1.
\end{align*}
Therefore, the worst-case expected regret of any policy $\pi$ can be lower bounded by the worst-case expected regret of parameters belonging to $\mathcal{W}$, which can be further lower bounded by the ``average'' regret over a uniform prior over $\mathcal{W}$ as follows:
\begin{align*}
    \sup_{\wb} \EE_{\wb}^{\pi} \left[ \Regret(\wb) \right]
    &= \sup_{\wb} \EE_{\wb}^{\pi} \sum_{t=1}^T  R(S^\star, \wb) 
    - R(S_{t}, \wb) 
    \\
    &\geq \max_{\wb_V } \EE_{\wb_V}^{\pi} \sum_{t=1}^T  R(S^\star, \wb_V) - R(S_{t}, \wb_V) 
    \\
    &\geq \frac{1}{|\mathcal{V}_{d/4}|} \sum_{V \in \mathcal{V}_{d/4}} \EE_{\wb_V}^{\pi} \sum_{t=1}^T R(S^\star, \wb_V) - R(S_{t}, \wb_V) 
    \\
    &= \frac{1}{|\mathcal{V}_{d/4}|} \sum_{V \in \mathcal{V}_{d/4}} \EE_{\wb_V}^{\pi} \sum_{t=1}^T \left[ \sum_{i \in S^\star} p(i | S^\star, \wb_V) - \sum_{i \in S_t} p(i | S_t, \wb_V)  \right]
    .
    \numberthis \label{eq:lower_bayes_risk}
\end{align*}
This reduces the task of lower bounding the worst-case regret of any policy to the task of lower bounding the \textit{Bayes risk} of the constructed parameter set.

\subsection{Main Proof of Theorem~\ref{thm:lower_bound}}
\label{app_subsec:lower_bound_main}
\begin{proof}[Proof of Theorem~\ref{thm:lower_bound}]
For any sequence of assortments $\{S_t \}_{t=1}^T$ produced by policy $\pi$, we denote an alternative sequence $\{ \tilde{S}_t \}_{t=1}^T$ that provably enjoys less regret under parameterization $\wb_{V}$.

Let $x_{U_1}, \dots, x_{U_L}$ be the distinct feature vectors contained in assortments $S_t$ (if $S_t = \emptyset$, then one may choose an arbitrary feature $x_U$) with $U_1, \dots, U_L \in \mathcal{V}_{d/4}$.
Let $U^\star$ be the subset among $U_1, \dots, U_L$ that maximizes $x_U^\top \wb_V$, i.e., $U^\star \in \argmax_{U \in \{U_1, \dots, U_L \}} x_U^\top \wb_V$, where $\wb_V$ is the underlying parameter.
Then, we define $\tilde{S}_t$ as the assortment consisting of all $K$ items corresponding to feature $x_{U^\star}$, i.e., $\tilde{S}_t = \{ \underbrace{x_{U^\star}, \dots, x_{U^\star}}_{K} \}$.

Since the expected revenue is an increasing function, we have the following observation:
\begin{proposition} [Proposition 1 in~\citealt{chen2020dynamic}] \label{prop:lower_S_tilde}
    \begin{align*}
        \sum_{i \in S_t} p(i | S_t, \wb_V)
        \leq \sum_{i \in \tilde{S}_t} p(i | \tilde{S}_t, \wb_V).
    \end{align*}
\end{proposition}
Proposition~\ref{prop:lower_S_tilde} implies that $\sum_{i \in S^\star} p(i | S^\star, \wb_V) - \sum_{i \in S_t} p(i | S_t, \wb_V)  \geq \sum_{i \in S^\star} p(i | S^\star, \wb_V) - \sum_{i \in \tilde{S}_t} p(i | \tilde{S}_t, \wb_V)$.
Hence, it is sufficient to bound $\sum_{i \in S^\star} p(i | S^\star, \wb_V) - \sum_{i \in \tilde{S}_t} p(i | \tilde{S}_t, \wb_V) $ instead of $\sum_{i \in S^\star} p(i | S^\star, \wb_V) - \sum_{i \in S_t} p(i | S_t, \wb_V)$.

To simplify notation, we denote $\tilde{U}_t$ as the unique $U^\star \in \mathcal{V}_{d/4}$ in $\tilde{S}_t$.
We also use $\EE_{V}$ and $\mathbb{P}_{V}$ to denote the expected value and probability, respectively, as governed by the law parameterized by  $\wb_V$ and under policy $\pi$.
Then, we can establish a lower bound for $\sum_{i \in S^\star} p(i | S^\star, \wb_V) - \sum_{i \in \tilde{S}_t} p(i | \tilde{S}_t, \wb_V) $ as follows:
\begin{lemma} \label{lemma:lower_revenue_gap}
    Suppose $\epsilon \in (0, 1/d\sqrt{d})$ and define $\delta := d/4 - | \tilde{U}_t \cap V |$.
    Then, we have
    \begin{align*}
        \sum_{i \in S^\star} p(i | S^\star, \wb_V) - \sum_{i \in \tilde{S}_t} p(i | \tilde{S}_t, \wb_V) 
        \geq \frac{v_0 K}{(v_0 + K e)^2} \cdot \frac{\delta \epsilon}{2 \sqrt{d}}
        .
    \end{align*}
\end{lemma}
For any $j \in V$, define random variables $\tilde{M}_j := \sum_{t=1}^T \mathbf{1} \{ j \in \tilde{U}_t \}$.
Then, by Lemma~\ref{lemma:lower_revenue_gap}, for all $V \in \mathcal{V}_{d/4}$, we have
\begin{align*}
    \EE_{V} 
    \sum_{t=1}^T
    \left[
    \sum_{i \in S^\star} p(i | S^\star, \wb_V) - \sum_{i \in \tilde{S}_t} p(i | \tilde{S}_t, \wb_V)
    \right]
    \geq \frac{v_0 K}{(v_0 + K e)^2} \cdot \frac{\epsilon}{2 \sqrt{d}} \left( \frac{dT}{4} - \sum_{j \in V} \EE_{V}[\tilde{M}_j] \right).
    \numberthis \label{eq:lower_expected_revenue_gap}
\end{align*}
Furthermore, we define $\mathcal{V}_{d/4}^{(j)} := \{ V \in \mathcal{V}_{d/4} : j \in V \}$ and $\mathcal{V}_{d/4 -1}:= \{ V \subseteq [d] : |V| = d/4 -1 \}$.
By taking the average of both sides of Equation~\eqref{eq:lower_expected_revenue_gap} with respect to all $V \in \mathcal{V}_{d/4}$, we obtain
\begin{align*}
    &\frac{1}{\left|\mathcal{V}_{d/4} \right|} \sum_{V \in \mathcal{V}_{d/4}} 
    \EE_{V} 
    \sum_{t=1}^T
    \left[
    \sum_{i \in S^\star} p(i | S^\star, \wb_V) - \sum_{i \in \tilde{S}_t} p(i | \tilde{S}_t, \wb_V)
    \right]
    \\
    &\geq \frac{v_0 K}{(v_0 + K e)^2} \cdot \frac{\epsilon}{2 \sqrt{d}} \cdot 
    \frac{1}{\left|\mathcal{V}_{d/4} \right|} \sum_{V \in \mathcal{V}_{d/4}} \left( \frac{dT}{4} - \sum_{j \in V} \EE_{V}[\tilde{M}_j] \right)
    \\
    &= \frac{v_0 K}{(v_0 + K e)^2} \cdot \frac{\epsilon}{2 \sqrt{d}}  
    \left( \frac{dT}{4} - \frac{1}{\left|\mathcal{V}_{d/4} \right|} \sum_{j=1}^d \sum_{V \in \mathcal{V}_{d/4}^{(j)}} \EE_{V}[ \tilde{M}_j] \right) 
    \\
    &= \frac{v_0 K}{(v_0 + K e)^2} \cdot \frac{\epsilon}{2 \sqrt{d}}  
    \left(\frac{dT}{4} - \frac{1}{\left|\mathcal{V}_{d/4} \right|}  \sum_{V \in \mathcal{V}_{d/4 - 1}} \sum_{j \notin V} \EE_{V \cup \{j\}}[ \tilde{M}_j]   \right)
    \\
    &\geq \frac{v_0 K}{(v_0 + K e)^2} \cdot \frac{\epsilon}{2 \sqrt{d}}  
    \left(\frac{dT}{4} - \frac{ \left| \mathcal{V}_{d/4 -1} \right| }{\left|\mathcal{V}_{d/4} \right|}    
    \max_{V \in \mathcal{V}_{d/4 -1}}  \sum_{j \notin V} \EE_{V \cup \{j\}}[ \tilde{M}_j] \right)
    \\
    &= \frac{v_0 K}{(v_0 + K e)^2} \cdot \frac{\epsilon}{2 \sqrt{d}}  
    \left(\frac{dT}{4} - \frac{ \left| \mathcal{V}_{d/4 -1} \right| }{\left|\mathcal{V}_{d/4} \right|}    
    \max_{V \in \mathcal{V}_{d/4 -1}}
    \sum_{j \notin V}
    \EE_{V} [ \tilde{M}_j] 
    +\EE_{V \cup \{j\}}[ \tilde{M}_j]
    - \EE_{V} [ \tilde{M}_j]
    \right).
\end{align*}
For any fixed $V$, we get $\sum_{j \notin V} \EE_{V} [ \tilde{M}_j]  \leq \sum_{j=1}^d \EE_{V} [ \tilde{M}_j ] \leq dT/4$.
Also, we have $\frac{ \left| \mathcal{V}_{d/4 -1} \right| }{\left|\mathcal{V}_{d/4} \right|} = \binom{d}{d/4 - 1} / \binom{d}{d/4} = \frac{d/4}{3d/4 +1} \leq \frac{1}{3}$.
Consequently, we derive that
\begin{align*}
    \frac{1}{\left|\mathcal{V}_{d/4} \right|} &\sum_{V \in \mathcal{V}_{d/4}} \EE_{V}
    \sum_{t=1}^T
    \left[\sum_{i \in S^\star} p(i | S^\star, \wb_V) - \sum_{i \in \tilde{S}_t} p(i | \tilde{S}_t, \wb_V)\right]
    \\
    &\geq \frac{v_0 K}{(v_0 + K e)^2} \cdot \frac{\epsilon}{2 \sqrt{d}} 
    \left( \frac{dT}{6} - \max_{V \in \mathcal{V}_{d/4 -1}} \sum_{j \notin V} 
    \left| \EE_{V \cup \{j\}}[ \tilde{M}_j]
    - \EE_{V} [ \tilde{M}_j] \right| \right).
    \numberthis \label{eq:lower_expected_revenue_gap_before_pinsker}
\end{align*}
Now we bound the term $\left| \EE_{V \cup \{j\}}[ \tilde{M}_j] - \EE_{V} [ \tilde{M}_j] \right|$ in Equation~\eqref{eq:lower_expected_revenue_gap_before_pinsker} for any $V \in \mathcal{V}_{d/4 -1}$.
For simplicity, let $P = \mathbb{P}_V$ and $Q = \mathbb{P}_{V \cup \{j\}}$ denote the laws under $\wb_V$ and $\wb_{V \cup j}$, respectively.
Then, we have
\begin{align*}
    \left| \EE_{P}[ \tilde{M}_j] - \EE_{Q} [ \tilde{M}_j\right] |
    &\leq \sum_{t=0}^T t \cdot \left| P[\tilde{M}_j =t ] - Q [\tilde{M}_j = t] \right|
    \\
    &\leq T \cdot \sum_{t=0}^T \left| P[\tilde{M}_j =t ] - Q [\tilde{M}_j = t] \right|
    \\
    &\leq T \cdot \| P-Q \|_{\operatorname{TV}} 
    \leq T \cdot \sqrt{\frac{1}{2} \operatorname{KL}(P \| Q)},
    \numberthis \label{eq:lower_pinskers}
\end{align*}
where $\| P-Q \|_{\operatorname{TV}} = \sup_{A} |P(A) - Q(A) |$ | is the total variation distance between $P$ and $Q$, $\operatorname{KL}(P \| Q) = \int (\log \dd P / \dd Q)\dd P$ is s the Kullback-Leibler (KL) divergence between $P$ and $Q$, and the last inequality holds by Pinsker's inequality.
Now, we bound the KL divergence term using the following Lemma.
\begin{lemma} \label{lemma:lower_KL}
    For any $V \in \mathcal{V}_{d/4 - 1}$ and $j \in [d]$, there exists a positive constant $C_{\operatorname{KL}} >0$ such that
    \begin{align*}
        \operatorname{KL}(P_{V} \| Q_{V \cup \{j\}}) 
        \leq  C_{\operatorname{KL}}  \cdot  \frac{v_0 K}{(v_0 + K)^2}  \cdot \frac{ \EE_{V}[ \tilde{M}_j]\epsilon^2}{d}.
    \end{align*}
\end{lemma}
Therefore, combining Equation~\eqref{eq:lower_expected_revenue_gap_before_pinsker}, Equation~\eqref{eq:lower_pinskers}, and Lemma~\ref{lemma:lower_KL}, we have
\begin{align*}
     \frac{1}{\left|\mathcal{V}_{d/4} \right|} &\sum_{V \in \mathcal{V}_{d/4}} 
     \EE_{V} 
    \sum_{t=1}^T
    \left[
    \sum_{i \in S^\star} p(i | S^\star, \wb_V) - \sum_{i \in \tilde{S}_t} p(i | \tilde{S}_t, \wb_V)
    \right]
     \\
     &\geq \frac{v_0 K}{(v_0 + K e)^2} \cdot \frac{\epsilon}{2 \sqrt{d}} 
    \left( \frac{dT}{6} - T \sum_{j=1}^d \sqrt{ C_{\operatorname{KL}}  \cdot  \frac{v_0 K}{(v_0 + K)^2}  \cdot \frac{ \EE_{V}[ \tilde{M}_j]\epsilon^2}{d} } \right)
    \\
    &\geq \frac{v_0 K}{(v_0 + K e)^2} \cdot \frac{\epsilon}{2 \sqrt{d}} 
    \left( \frac{dT}{6} - T \sqrt{d} \cdot \sqrt{ \sum_{j=1}^d C_{\operatorname{KL}}  \cdot  \frac{v_0 K}{(v_0 + K)^2}  \cdot \frac{ \EE_{V}[ \tilde{M}_j]\epsilon^2}{d} } \right)
    \\
    &\geq \frac{v_0 K}{(v_0 + K e)^2} \cdot \frac{\epsilon}{2 \sqrt{d}} 
    \left( \frac{dT}{6} - T \sqrt{d} \cdot \sqrt{ \frac{C_{\operatorname{KL}} }{4}  \cdot  \frac{v_0 K}{(v_0 + K)^2}  \cdot T \epsilon^2  } \right)
    ,
\end{align*}
where the second inequality is due to the Cauchy-Schwartz inequality and the last inequality holds because $\sum_{j=1}^d \EE_{V}[ \tilde{M}_j] \leq dT/4$.
Let $C_{\operatorname{KL}}' = C_{\operatorname{KL}}/4$.

By setting $\epsilon = \sqrt{\frac{d}{144 C_{\operatorname{KL}}' T } \cdot \frac{(v_0 + K)^2}{v_0 K}}$, we have
\begin{align*}
    \sup_{\wb} \EE_{\wb}^{\pi} \left[ \Regret(\wb) \right]
     &\geq \frac{v_0 K}{(v_0 + K e)^2} \cdot \frac{\epsilon}{2 \sqrt{d}} 
    \left( \frac{dT}{6} -  \sqrt{C_{\operatorname{KL}}'\cdot  \frac{v_0 K}{(v_0 + K)^2}   d T \epsilon^2 } \right)
    \\
    &= \frac{v_0 K }{(v_0 + K e)^2} \cdot \sqrt{\frac{(v_0 + K)^2}{v_0 K} } \cdot \frac{1}{288 \sqrt{C_{\operatorname{KL}}'} } d\sqrt{T}
    \\
    &= \Omega \left( \frac{\sqrt{v_0 K}}{v_0 + K} \cdot d\sqrt{T} \right).
\end{align*}
This concludes the proof of Theorem~\ref{thm:lower_bound}.
\end{proof}

\subsection{Proofs of Lemmas for Theorem~\ref{thm:lower_bound}}
\subsubsection{Proof of Lemma~\ref{lemma:lower_revenue_gap}}
\begin{proof} [Proof of Lemma~\ref{lemma:lower_revenue_gap}]
    Let $x = x_{V}$ and $\hat{x} = x_{\tilde{U}_t}$ be the corresponding context vectors.
    Then, we have
    \begin{align*}
        \sum_{i \in S^\star} p(i | S^\star, \wb_V) - \sum_{i \in \tilde{S}_t} p(i | \tilde{S}_t, \wb_V) 
        &= \frac{K \exp \left( x^\top \wb_{V} \right)}{v_0 + K \exp \left( x^\top \wb_{V} \right)} 
        - \frac{K \exp \left( \hat{x}^\top \wb_{V} \right)}{v_0 + K \exp \left( \hat{x}^\top \wb_{V} \right)}
        \\
        &= \frac{v_0 K \left( \exp \left( x^\top \wb_{V} \right) - \exp \left( \hat{x}^\top \wb_{V} \right) \right)}{\left( v_0 + K \exp \left( x^\top \wb_{V} \right)\right) \left(v_0 + K \exp \left( \hat{x}^\top \wb_{V} \right) \right)}
        \\
        &\geq \frac{v_0 K \left( \exp \left( x^\top \wb_{V} \right) - \exp \left( \hat{x}^\top \wb_{V} \right) \right)}{(v_0 + K e)^2},
        \numberthis \label{eq:lemma:lower_revenue_gap_intermediate}
    \end{align*}
    where the inequality holds since $\max\left\{\exp \left( x^\top \wb_{V} \right), \exp \left( \hat{x}^\top \wb_{V} \right) \right\} \leq e$.
    To further bound the right-hand side of Equation~\eqref{eq:lemma:lower_revenue_gap_intermediate}, we use the fact that $1 + a \leq e^a \leq 1 + a + a^2/2$ for all $a \in [0,1]$, which can be easily shown by Taylor expansion.
    Thus, we get
    \begin{align*}
        \sum_{i \in S^\star} p(i | S^\star, \wb_V) - \sum_{i \in \tilde{S}_t} p(i | \tilde{S}_t, \wb_V) 
        &\geq \frac{v_0 K \left( (x - \hat{x})^\top \wb_V - (\hat{x}^\top \wb_V)^2/2 \right)}{(v_0 + K e)^2}
        \\
        &\geq \frac{v_0 K \left( \delta \epsilon/ \sqrt{d} - (\sqrt{d} \epsilon)^2/2 \right) }{(v_0 + K e)^2}
        \\
        &\geq \frac{v_0 K \delta \epsilon }{2 \sqrt{d}(v_0 + K e)^2},
    \end{align*}
    where the last inequality holds because $(\sqrt{d} \epsilon)^2 \leq \delta \epsilon/\sqrt{d}$ when $\epsilon \in (0, 1/d\sqrt{d})$.
    This concludes the proof.
\end{proof}

\subsubsection{Proof of Lemma~\ref{lemma:lower_KL}}
\begin{proof} [Proof of Lemma~\ref{lemma:lower_KL}]
    Fix a round $t$, an assortment $\tilde{S}_t$, and $\tilde{U}_t$.
    Let $U = \tilde{U}_t$.
    Define $m_j(\tilde{S}_t) := \sum_{x_U \in \tilde{S}_t}\mathbf{1} \{j \in U \} /K$.
    Let $\{p_i\}_{i \in \tilde{S}_t \cup \{0\}}$ and $\{q_i\}_{i \in \tilde{S}_t \cup \{0\}}$ be the probabilities of choosing item $i$ under parameterizations  $\wb_{V}$ and $\wb_{V \cup \{j\}}$, respectively.
    Then, we have
    \begin{align*}
        \operatorname{KL}\left( \mathbb{P}_{V} (\cdot | \tilde{S}_t ) \| \mathbb{P}_{V \cup \{j\}} (\cdot |  \tilde{S}_t) \right)
        = \sum_{i \in \tilde{S}_t \cup \{0\} } p_i \log \frac{p_i}{q_i}
        \leq \sum_{i \in \tilde{S}_t \cup \{0\} } p_i \frac{p_i - q_i}{q_i} 
        \leq \sum_{i \in \tilde{S}_t \cup \{0\} } \frac{(p_i - q_i)^2}{q_i},
    \end{align*}
    where the first inequality holds because $\log (1+x) \leq x$ for all $x > -1$.

    Let $\hat{x} = x_{U}$.
    Now, we separately upper bound $(p_i - q_i)^2/q_i$, by analyzing the following three cases:

    \textbf{Case 1. } The outside option, $i=0$.
    
    For $i=0$, $q_i \geq \frac{v_0}{v_0 + Ke}$.
    Thus, we have
    \begin{align*}
        |p_i - q_i|
        &= \left| \frac{v_0}{v_0 +  \sum_{i \in \tilde{S}_t}\exp\left( x_i^\top \wb_V \right) } 
        - \frac{v_0}{v_0 + \sum_{i \in \tilde{S}_t} \exp\left( x_i^\top \wb_{V \cup \{j\} } \right) } \right|
        \\
        &= \left| \frac{v_0}{v_0 +  K\exp\left( \hat{x}^\top \wb_V \right) } 
        - \frac{v_0}{v_0 + K \exp\left( \hat{x}^\top \wb_{V \cup \{j\} } \right) } \right|
        \\
        &\leq \frac{v_0 K }{(v_0 + K/e)^2}  \left| \exp\left( \hat{x}^\top \wb_V \right) -  \exp\left( \hat{x}^\top \wb_{V \cup \{j\} } \right)  \right| 
        \\
        &= \frac{v_0 K}{(v_0 + K/e)^2} \left| e^{\bar{c}_1} (\hat{x}^\top \wb_V -  \hat{x}^\top \wb_{V \cup \{j\} }) \right|
        \\
        &\leq \frac{v_0 K e}{(v_0 + K/e)^2}  \left| \hat{x}^\top \left(\wb_V -  \wb_{V \cup \{j\} } \right) \right|
        \leq \frac{v_0 K e}{(v_0 + K/e)^2} \cdot \frac{m_j(\tilde{S}_t) \epsilon}{\sqrt{d}}
        ,
    \end{align*}
    where the third equality holds by applying the mean value theorem for the exponential function,  with $\bar{c}_1 := (1-u)(\hat{x}^\top \wb_V) +u (\hat{x}^\top \wb_{V \cup \{j\} })$ for some $u \in (0,1)$.
    Then, there exist an absolute constant $C_0$ such that
    \begin{align*}
        \frac{(p_0 - q_0)^2}{q_0}
        &\leq \frac{v_0^2 K^2 e^2}{(v_0  + K/e)^4} \cdot \frac{ \left(m_j(\tilde{S}_t)\right)^2 \epsilon^2}{d}
        \cdot 
        \frac{v_0 + Ke}{v_0}
        \\
        &\leq C_0 \cdot \frac{v_0 K^2}{(v_0 + K)^3 } \cdot \frac{  m_j(\tilde{S}_t) \epsilon^2}{ d },
        \numberthis \label{eq:lower_pinsker_case1}
    \end{align*}
    where the last inequality holds since $m_j(\tilde{S}_t)\leq 1$.

    \textbf{Case 2. }$i \in \tilde{S}_t$ and $j \notin U$.

    Then, for any $i \in \tilde{S}_t$ corresponding to $x_i = \hat{x}$ and $j \notin U$, we have
    \begin{align*}
        |p_i - q_i|
        &= \left| \frac{\exp\left( \hat{x}^\top \wb_{V} \right) }{v_0 + K \exp\left( \hat{x}^\top \wb_V \right) } 
        - \frac{\exp\left( \hat{x}^\top \wb_{V \cup \{j \}} \right)}{v_0 + K \exp\left( \hat{x}^\top \wb_{V \cup \{j\} } \right) } \right|
        =0
        ,
    \end{align*}
    where the last equality holds because $\exp\left( \hat{x}^\top \wb_{V} \right) =  \exp\left( \hat{x}^\top \wb_{V \cup \{j \}} \right)$, given that $j \notin U$.
    Thus, we get
    \begin{align*}
        \sum_{i \in \tilde{S}_t, j \notin U} \frac{(p_i - q_i)^2}{q_i}  
        =0,
        \numberthis \label{eq:lower_pinsker_case2}
    \end{align*}

    \textbf{Case 3. }$i \in \tilde{S}_t$ and $j \in U$.

    Recall that for any $i \in \tilde{S}_t$, $q_i \geq \frac{e^{-1}}{v_0 + Ke}$.
    Then, for any $i \in \tilde{S}_t$ corresponding to $x_i = \hat{x}$ and $j \in U$, we have
    \begin{align*}
        |p_i - q_i|
        &= \left| \frac{\exp\left( \hat{x}^\top \wb_{V} \right) }{v_0 + K  \exp\left( \hat{x}^\top \wb_V \right) } 
        - \frac{\exp\left( \hat{x}^\top \wb_{V \cup \{j \}} \right)}{v_0 + K \exp\left( \hat{x}^\top \wb_{V \cup \{j\} } \right) } \right|
        \\
        &= \left| \frac{ \exp\left(\bar{c}_2 \right) }{v_0 + K \exp\left(\bar{c}_2 \right) }
        \cdot \hat{x}^\top \left(\wb_V -  \wb_{V \cup \{j\} } \right)
        - \frac{ K \exp\left(2\bar{c}_2 \right) }{ \left(v_0 + K \exp\left(\bar{c}_2 \right)  \right)^2}
        \cdot \hat{x}^\top \left(\wb_V -  \wb_{V \cup \{j\} } \right)
        \right|
        \\
        &=  \frac{ \exp\left(\bar{c}_2 \right) v_0 }{\left(v_0 + K \exp\left(\bar{c}_2 \right) \right)^2}
        \left| \hat{x}^\top \left(\wb_V -  \wb_{V \cup \{j\} } \right)
        \right|
        \\
        &\leq
        \frac{ v_0 e }{(v_0 + K/e)^2}
        \left| \hat{x}^\top \left(\wb_V -  \wb_{V \cup \{j\} } \right)
        \right|
        \leq \frac{ v_0 e }{(v_0 + K/e)^2} \cdot \frac{m_j(\tilde{S}_t) \epsilon}{\sqrt{d}}
        ,
    \end{align*}
     the second equality holds by applying the mean value theorem,  with $\bar{c}_2 := (1-u)(\hat{x}^\top \wb_V) +u (\hat{x}^\top \wb_{V \cup \{j\} })$ for some $u \in (0,1)$.
    Then, there exist an absolute constant $C_1$ such that
    \begin{align*}
        \sum_{i \in \tilde{S}_t, j \in U} \frac{(p_i - q_i)^2}{q_i}
        &\leq K m_j(\tilde{S}_t) \cdot
        \frac{ v_0^2 e^2 }{(v_0 + K/e)^4}  \cdot \frac{\left(m_j(\tilde{S}_t)\right)^2 \epsilon^2}{d}
        \cdot \frac{v_0 + Ke}{e^{-1}}
        \\
        &\leq C_1 \cdot \frac{v_0^2 K}{(v_0 + K)^3} \cdot \frac{  m_j(\tilde{S}_t) \epsilon^2}{ d }
        ,
        \numberthis \label{eq:lower_pinsker_case3}
    \end{align*}
    where the last inequality holds since $m_j(\tilde{S}_t)\leq 1$.

    Combining Equation~\eqref{eq:lower_pinsker_case1},~\eqref{eq:lower_pinsker_case2}, and~\eqref{eq:lower_pinsker_case3}, we derive that
    \begin{align*}
        \sum_{i \in \tilde{S}_t \cup \{0\} } \frac{(p_i - q_i)^2}{q_i} 
        &\leq
        \left( C_0 \cdot  \frac{v_0 K^2}{(v_0 + K)^3 } 
        + C_1 \cdot \frac{v_0^2 K}{(v_0 + K)^3} \right) 
        \cdot \frac{ m_j(\tilde{S}_t) \epsilon^2}{d}
        \\
        &\leq
        \max\{C_0, C_1\}  \cdot \frac{v_0K}{(v_0 + K)^2} \cdot \frac{ m_j(\tilde{S}_t) \epsilon^2}{d}
         \\
        &= C_{\operatorname{KL}} \cdot \frac{v_0K}{(v_0 + K)^2} \cdot \frac{ m_j(\tilde{S}_t) \epsilon^2}{d},
    \end{align*}
    where $C_{\operatorname{KL}}=\max\{C_0, C_1\}$.
    Since $\tilde{M}_j = \sum_{t=1}^T m_j(\tilde{S}_t)$ by definition, and subsequently summing over all $t=1$ to $T$, we have
    \begin{align*}
        \operatorname{KL}\left( P_V \| Q_{V \cup \{j\} } \right)
        &= \sum_{t=1}^T \EE_{V} \left[ \operatorname{KL}\left( \mathbb{P}_{V} (\cdot | \tilde{S}_t ) \| \mathbb{P}_{V \cup \{j\}} (\cdot |  \tilde{S}_t) \right) \right]
        \\
        &\leq  C_{\operatorname{KL}}  \cdot   \frac{v_0K}{(v_0 + K)^2}  \cdot \frac{ \EE_{V}[ \tilde{M}_j]\epsilon^2}{d},
    \end{align*}
    where the equality holds by the chain rule of relative entropy (cf. Exercise 14.11 of~\citet{lattimore2020bandit}).
    This concludes the proof.
\end{proof}

\section{Proof of Theorem~\ref{thm:upper_bound}} 
\label{app_sec:proof_of_thm_upper}
In this section, we present the proof of Theorem~\ref{thm:upper_bound}.
Note that when the rewards are uniform, the revenue increases as a function of the assortment size. 
Therefore,  maximizing the expected revenue $ R_{t}(S, \wb)$ across all possible assortments $S \in \mathcal{S}$  always contains exactly $K$ items.
In other words, the size of the chosen assortment $S_t$ and the size of the optimal assortment $S_t^\star$ both equal to $K$.
\subsection{Main Proof of Theorem~\ref{thm:upper_bound}}
\label{app_subsec:main_proof_thm_upper}

Before presenting the proof, we introduce useful lemmas, whose proof can be found in Appendix~\ref{app_subsec:useful_lemmas_thm1}.
Lemma~\ref{lemma:utility} shows the optimistic utility for the context vectors.

\begin{lemma} \label{lemma:utility}
    Let $\alpha_{ti} = x_{ti}^\top \wb_{t} + \beta_t (\delta) \|  x_{ti}\|_{H_t^{-1}}$.
    If $\wb^\star \in \mathcal{C}_t(\delta)$, then we have
    \begin{align*}
        0 \leq \alpha_{ti} - x_{ti}^\top \wb^\star \leq 2 \beta_t(\delta) \| x_{ti} \|_{H_t^{-1}}.
    \end{align*}
\end{lemma}
Lemma~\ref{lemma:revenue_second_pd} is a $K$-free elliptical potential lemma that improves upon the one presented in Lemma 10 of~\citet{perivier2022dynamic} in terms of $K$.
Lemma 10 of~\citet{perivier2022dynamic} states: 
$\sum_{s=1}^{t} \sum_{i \in S_s} p_s(i | S_s, \wb^\star) p_s(0 | S_s, \wb^\star) \| x_{si}\|_{H_s(\wb^\star)^{-1} }^2 \leq 2d K \log \left( \lambda_{t+1} + \frac{2tK}{d} \right)$ 
and 
$\sum_{s=1}^{t} \max_{i \in S_s} \| x_{si}\|_{H_s(\wb^\star)^{-1} }^2 \leq  2d\left( K + \frac{1}{\kappa} \right)  \log \left(\lambda_{t+1} + \frac{2tK}{d} \right)$,
where $H_t(\wb) = \sum_{s=1}^{t-1} \Gcal_s(\wb) + \lambda_t \Ib_d$.
\begin{lemma}   \label{lemma:elliptical}
Let $H_t = \lambda \Ib_d + \sum_{s=1}^{t-1} \Gcal_s(\wb_{s+1})$, where $\Gcal_s(\wb) = \sum_{i \in S_s} p_s(i | S_s, \wb) x_{si} x_{si}^\top -  \sum_{i \in S_s}  \sum_{j \in S_s} p_s(i | S_s, \wb) p_s(j | S_s, \wb) x_{si} x_{sj}^\top$.
Suppose $\lambda \geq 1$.
Then the following statements hold true:
\begin{enumerate}[label={(\arabic*)}]
    \item $\sum_{s=1}^{t} \sum_{i \in S_s} p_s(i | S_s, \wb_{s+1}) p_s(0 | S_s, \wb_{s+1}) \| x_{si}\|_{H_s^{-1}}^2
    \leq 2d \log \left( 1+ \frac{t}{d \lambda} \right)$,

    \item $\sum_{s=1}^{t} \max_{i \in S_s} \| x_{si}\|_{H_s^{-1}}^2
    \leq \frac{2}{\kappa} d \log \left( 1+ \frac{t}{d \lambda} \right)$.
\end{enumerate}
\end{lemma}
Moreover, we provide a tighter bound for the second derivative of the expected revenue than that presented in Lemma 12 of~\citet{perivier2022dynamic}.
Lemma 12 of~\citet{perivier2022dynamic} states: 
$\left| \frac{\partial^2 Q}{\partial i \partial j} \right| \leq 5$.

\begin{lemma}
\label{lemma:revenue_second_pd}
Define $Q:\RR^K \rightarrow \RR$, such that for any $\ub = (u_1, \dots, u_K) \in \RR^K$, $Q(\ub) = \sum_{i=1}^K \frac{\exp(u_i)}{v_0 + \sum_{k=1}^K \exp(u_k)}$.
Let $p_i(\ub) = \frac{\exp(u_i)}{v_0 + \sum_{k=1}^K \exp(u_k)}$.
Then, for all $i \in [K]$, we have
\begin{align*}
    \left| \frac{\partial^2 Q}{\partial i \partial j} \right|
    \leq
    \begin{cases}
        3 p_i(\ub) & \text{if} \,\,\, i=j,
        \\
        2p_i(\ub) p_j(\ub) & \text{if} \,\,\, i \neq j.
    \end{cases}
\end{align*}
\end{lemma}

Now, we are ready to prove Theorem~\ref{thm:upper_bound}.
\begin{proof}[Proof of Theorem~\ref{thm:upper_bound}]
    First, we bound the regret as follows:
    \begin{align*}
        \sum_{t=1}^T  &R_{t}(S_{t}^\star, \wb^\star) -  R_{t}(S_{t}, \wb^\star)  
        = \sum_{t=1}^T \left[ \sum_{i \in S_t^\star} p_t(i | S_t^\star, \wb^\star) - \sum_{i \in S_t} p_t (i | S_t, \wb^\star)   \right]
        \\
        &= \sum_{t=1}^T \left[ \frac{\sum_{i \in S_t^\star} \exp(x_{ti}^\top \wb^\star)}{v_0 + \sum_{j \in S_t^\star} \exp(x_{tj}^\top \wb^\star) }  
        -  \frac{\sum_{i \in S_t} \exp(x_{ti}^\top \wb^\star)}{v_0 + \sum_{j \in S_t} \exp(x_{tj}^\top \wb^\star)}
        \right]
        \\
        &\leq \sum_{t=1}^T \left[ \frac{\sum_{i \in S_t^\star} \exp(\alpha_{ti})}{v_0 + \sum_{j \in S_t^\star} \exp(\alpha_{tj}) }  
        -  \frac{\sum_{i \in S_t} \exp(x_{ti}^\top \wb^\star)}{v_0 + \sum_{j \in S_t} \exp(x_{tj}^\top \wb^\star)}
        \right]
        \\
        &\leq \sum_{t=1}^T \left[ \frac{\sum_{i \in S_t} \exp(\alpha_{ti})}{v_0 + \sum_{j \in S_t} \exp(\alpha_{tj}) }  
        -  \frac{\sum_{i \in S_t} \exp(x_{ti}^\top \wb^\star)}{v_0 + \sum_{j \in S_t} \exp(x_{tj}^\top \wb^\star)}
        \right]
        = \sum_{t=1}^T  \tilde{R}_t(S_{t}) -  R_{t}(S_{t}, \wb^\star) 
        ,
    \end{align*}
    where the first inequality holds by Lemma~\ref{lemma:utility}, and the last inequality holds by the assortment selection of Algorithm~\ref{alg:minimax}.

    Now, we define $Q:\RR^K \rightarrow \RR$, such that for all $\ub = (u_1, \dots, u_K)^\top \in \RR^K$, $Q(\ub) = \sum_{i=1}^K \frac{\exp(u_i)}{v_0 + \sum_{j=1}^K \exp(u_j)}$.
    Noting that $S_t$ always contains $K$ elements since the expected revenue is an increasing function in the uniform reward setting, we can write $S_t = \{i_1, \dots, i_K \}$.
    Moreover, for all $t \geq 1$, let $\ub_t = (u_{ti_1}, \dots u_{ti_K} )^\top = (\alpha_{ti_1} , \dots, \alpha_{ti_{K}} )^\top$ 
    and $\ub^\star_t = (u_{ti_1}^\star, \dots u_{ti_K}^\star )^\top = (x_{ti_1}^\top \wb^\star, \dots, x_{ti_{K}}^\top \wb^\star)^\top$.
    Then, by a second order Taylor expansion, we have
    \begin{align*}
        \sum_{t=1}^T  \tilde{R}_t(S_{t}) -  R_{t}(S_{t}, \wb^\star) 
        &=  \sum_{t=1}^T   Q(\ub_t) - Q(\ub^\star_t) 
        \\
        &=  \underbrace{  \sum_{t=1}^T \nabla Q(\ub_t^\star)^\top (\ub_t - \ub^\star_t) }_{\texttt{(A)}}
        + \underbrace{\frac{1}{2} \sum_{t=1}^T (\ub_t - \ub^\star_t)^\top \nabla^2 Q(\bar{\ub}_t) (\ub_t - \ub^\star_t)}_{\texttt{(B)}},
        \numberthis \label{eq:upper_uniform_regret_decompose}
    \end{align*}
    where $\bar{\ub}_t = (\bar{u}_{ti_1}, \dots, \bar{u}_{ti_K})^\top \in \RR^K$ is the convex combination of $\ub_t$ and $\ub^\star_t$.

    First, we bound the term \texttt{(A)}.
    \begin{align*}
        & \sum_{t=1}^T \nabla Q(\ub_t^\star)^\top (\ub_t - \ub^\star_t) 
        \\
        &=  \sum_{t=1}^T \sum_{i \in S_t} \frac{\exp(x_{ti}^\top \wb^\star )}{v_0 + \sum_{k\in S_t} \exp(x_{tk}^\top \wb^\star ) } (u_{ti} - u_{ti}^\star ) 
        -  \sum_{i \in S_t}  \sum_{j \in S_t} \frac{\exp(x_{ti}^\top \wb^\star ) \exp(x_{tj}^\top \wb^\star )}{(v_0 + \sum_{k\in S_t} \exp(x_{tk}^\top \wb^\star ))^2 } (u_{ti} - u_{ti}^\star )   
        \\
        &=  \sum_{t=1}^T \sum_{i \in S_t} p_t(i | S_t, \wb^\star)  (u_{ti} - u_{ti}^\star )
        - \sum_{i \in S_t}  \sum_{j \in S_t} p_t(i | S_t, \wb^\star) p_t(j | S_t, \wb^\star)  (u_{ti} - u_{ti}^\star )
        \\
        &=  \sum_{t=1}^T \sum_{i \in S_t} p_t(i | S_t, \wb^\star) \left( 1 - \sum_{j \in S_t} p_t(j | S_t, \wb^\star) \right)  (u_{ti} - u_{ti}^\star )   
        \\
        &=  \sum_{t=1}^T \sum_{t=1}^T \sum_{i \in S_t} p_t(i | S_t, \wb^\star) p_t(0 | S_t, \wb^\star)  (u_{ti} - u_{ti}^\star ) 
        \\
        &\leq   \sum_{t=1}^T \sum_{i \in S_t} p_t(i | S_t, \wb^\star) p_t(0 | S_t, \wb^\star)  2 \beta_t(\delta) \| x_{ti} \|_{H_t^{-1}} 
        \\
        &\leq 2  \beta_T(\delta)  \sum_{t=1}^T \sum_{i \in S_t} p_t(i | S_t, \wb^\star) p_t(0 | S_t, \wb^\star)  \| x_{ti} \|_{H_t^{-1}},
        \numberthis \label{eq:eff_term_A}
    \end{align*}
    where the first inequality holds by Lemma~\ref{lemma:utility}, and the last inequality holds because  $\beta_t(\delta)$ is increasing for $t \in [T]$.

    Now we bound the term \texttt{(B)}.
    Let $p_i(\bar{\ub}_t) = \frac{\exp(\bar{u}_{ti})}{v_0 + \sum_{k=1}^K \exp(\bar{u}_{tk})}$.
    Then, we have
    \begin{align*}
        &\frac{1}{2} \sum_{t=1}^T (\ub_t - \ub^\star_t)^\top \nabla^2 Q(\bar{\ub}_t) (\ub_t - \ub^\star_t)
        \\
        &= \frac{1}{2} \sum_{t=1}^T \sum_{i \in S_t} \sum_{j \in S_t}(u_{ti} - u_{ti}^\star ) \frac{\partial^2 Q}{\partial i \partial j} (u_{tj} - u_{tj}^\star )
        \\
        &= \frac{1}{2} \sum_{t=1}^T \sum_{i \in S_t} \sum_{j \in S_t, j \neq i} (u_{ti} - u_{ti}^\star ) \frac{\partial^2 Q}{\partial i \partial j} (u_{tj} - u_{tj}^\star )
        + \frac{1}{2} \sum_{t=1}^T \sum_{i \in S_t } (u_{ti} - u_{ti}^\star ) \frac{\partial^2 Q}{\partial i \partial i} (u_{ti} - u_{ti}^\star )
        \\
        &\leq \sum_{t=1}^T \sum_{i \in S_t} \sum_{j \in S_t, j \neq i} |u_{ti} - u_{ti}^\star| p_i(\bar{\ub}_t) p_j(\bar{\ub}_t) |u_{tj} - u_{tj}^\star|
        + \frac{3}{2} \sum_{t=1}^T \sum_{i \in S_t } (u_{ti} - u_{ti}^\star)^2 p_i(\bar{\ub}_t),
        \numberthis \label{eq:term_B}
    \end{align*}
    where the inequality is by Lemma~\ref{lemma:revenue_second_pd}.
    To bound the first term in Equation~\eqref{eq:term_B}, by applying the AM-GM inequality, we get
    \begin{align*}
        \sum_{t=1}^T & \sum_{i \in S_t} \sum_{j \in S_t, j \neq i} |u_{ti} - u_{ti}^\star| p_i(\bar{\ub}_t) p_j(\bar{\ub}_t) |u_{tj} - u_{tj}^\star|
        \\
        &\leq \sum_{t=1}^T\sum_{i \in S_t} \sum_{j \in S_t} |u_{ti} - u_{ti}^\star| p_i(\bar{\ub}_t) p_j(\bar{\ub}_t) |u_{tj} - u_{tj}^\star|
        \\
        &\leq \frac{1}{2} \sum_{t=1}^T\sum_{i \in S_t} \sum_{j \in S_t} (u_{ti} - u_{ti}^\star)^2 p_i(\bar{\ub}_t) p_j(\bar{\ub}_t) 
        + \frac{1}{2} \sum_{i \in S_t} \sum_{j \in S_t}  (u_{tj} - u_{tj}^\star)^2 p_i(\bar{\ub}_t) p_j(\bar{\ub}_t)
        \\
        &\leq  \sum_{t=1}^T\sum_{i \in S_t} (u_{ti} - u_{ti}^\star)^2 p_i(\bar{\ub}_t) 
        .
        \numberthis \label{eq:term_B_first}
    \end{align*}
    By plugging Equation~\eqref{eq:term_B_first} into Equation~\eqref{eq:term_B}, we have
    \begin{align*}
        \frac{1}{2} \sum_{t=1}^T (\ub_t - \ub^\star_t)^\top \nabla^2 Q(\bar{\ub}_t) (\ub_t - \ub^\star_t)
        &\leq \frac{5}{2} \sum_{t=1}^T \sum_{i \in S_t } (u_{ti} - u_{ti}^\star)^2 p_i(\bar{\ub}_t)
        \\
        &\leq 10 \sum_{t=1}^T\sum_{i \in S_t }  p_i(\bar{\ub}_t) \beta_t(\delta)^2  \| x_{ti} \|_{H_t^{-1}}^2 
        \\
        &\leq  10 \sum_{t=1}^T \max_{i \in S_t } \beta_t(\delta)^2  \| x_{ti} \|_{H_t^{-1}}^2 
        \\
        &\leq 10 \beta_T(\delta)^2  \sum_{t=1}^T \max_{i \in S_t }  \| x_{ti} \|_{H_t^{-1}}^2 
        ,
        \numberthis \label{eq:term_B_last}
    \end{align*}
    where the second inequality holds by Lemma~\ref{lemma:utility}.
    Combining the upper bound for the terms \texttt{(A)} and \texttt{(B)}, with probability at least $1-\delta$, we have  
    \begin{align*}
         \sum_{t=1}^T  \tilde{R}_t(S_{t}) -  R_{t}(S_{t}, \wb^\star) 
         &\leq 
         2  \beta_T(\delta)  \sum_{t=1}^T \sum_{i \in S_t} p_t(i | S_t, \wb^\star) p_t(0 | S_t, \wb^\star)  \| x_{ti} \|_{H_t^{-1}}
         \\
         &+ 10 \beta_T(\delta)^2  \sum_{t=1}^T \max_{i \in S_t }  \| x_{ti} \|_{H_t^{-1}}^2.
         \numberthis
         \label{eq:regret_decomposition}
    \end{align*}
    Now, we bound each term of Equation~\eqref{eq:regret_decomposition} respectively.
    For the first term, we decompose it as follows:
    \begin{align*}
        \sum_{t=1}^T \sum_{i \in S_t} &p_t(i | S_t, \wb^\star) p_t(0 | S_t, \wb^\star)  \|x_{ti} \|_{H_t^{-1}} 
        \\
        &= \sum_{t=1}^T  \sum_{i \in S_t} p_t(i | S_t, \wb_{t+1}) p_t(0 | S_t, \wb_{t+1})  \|x_{ti} \|_{H_t^{-1}}
        \\
        &+ \sum_{t=1}^T \sum_{i \in S_t} \left( p_t(i | S_t, \wb^\star) - p_t(i | S_t, \wb_{t+1}) \right) p_t(0 | S_t, \wb_{t+1})  \|x_{ti} \|_{H_t^{-1}}
        \\
        &+ \sum_{t=1}^T \sum_{i \in S_t} p_t(i | S_t, \wb^\star) \left( p_t(0 | S_t, \wb^\star) -  p_t(0 | S_t, \wb_{t+1}) \right)  \|x_{ti} \|_{H_t^{-1}}.
        \numberthis \label{eq:bouns_first_decomposition}
    \end{align*}
    To bound the first term on the right-hand side of Equation~\eqref{eq:bouns_first_decomposition}, we apply the Cauchy-Schwarz inequality.
    \begin{align*}
        &\sum_{t=1}^T \sum_{i \in S_t} p_t(i | S_t, \wb_{t+1}) p_t(0 | S_t, \wb_{t+1})  \|x_{ti} \|_{H_t^{-1}}
        \\
        &\leq \sqrt{ \sum_{t=1}^T \sum_{i \in S_t} p_t(i | S_t, \wb_{t+1}) p_t(0 | S_t, \wb_{t+1})} 
        \sqrt{\sum_{t=1}^T \sum_{i \in S_t} p_t(i | S_t, \wb_{t+1}) p_t(0 | S_t, \wb_{t+1})  \|x_{ti} \|_{H_t^{-1}}^2 }
        \\
        &\leq  \frac{\sqrt{v_0 K}}{(v_0 + K e^{-1})} \sqrt{T} 
        \sqrt{\sum_{t=1}^T \sum_{i \in S_t} p_t(i | S_t, \wb_{t+1}) p_t(0 | S_t, \wb_{t+1})  \|x_{ti} \|_{H_t^{-1}}^2 }
        \\
        &\leq  \frac{\sqrt{v_0 K}}{(v_0 + K e^{-1})} \sqrt{   T \cdot  2d \log \left( 1+ \frac{T}{d \lambda} \right)},
        \numberthis \label{eq:bouns_first_decomposition_1}
    \end{align*}
    where the last inequality holds by Lemma~\ref{lemma:elliptical}.

    Now, we bound the second term on the right-hand side of Equation~\eqref{eq:bouns_first_decomposition}.
    Let the \textit{virtual} context for the outside option be $x_{t0} = \mathbf{0}$.
    Then, by the mean value theorem, there exists $\xib_t = (1-c) \wb^\star + c \wb_{t+1}$ for some $c \in (0,1)$ such that 
    \begin{align*}
        & \sum_{i \in S_t} \left( p_t(i | S_t, \wb^\star) - p_t(i | S_t, \wb_{t+1}) \right) p_t(0 | S_t, \wb_{t+1})  \|x_{ti} \|_{H_t^{-1}}
        \\
        &= \sum_{i \in S_t}  \nabla p_t(i | S_t, \xib_t)^\top (\wb^\star - \wb_{t+1}) 
        p_t(0 | S_t, \wb_{t+1}) \| x_{ti} \|_{H_t^{-1}}
        \\
        &=  
        \sum_{i \in S_t}  \left( p_t(i | S_t, \xib_t) x_{ti} -  p_t(i | S_t, \xib_t) \sum_{j \in S_t} p_t(j | S_t, \xib_t) x_{tj}  \right)^\top 
        \!\!(\wb^\star - \wb_{t+1}) 
        p_t(0 | S_t, \wb_{t+1})\| x_{ti} \|_{H_t^{-1}}
        \\
        &\leq 
        \sum_{i \in S_t}   p_t(i | S_t, \xib_t) \left|x_{ti}^\top (\wb^\star - \wb_{t+1}) \right| p_t(0 | S_t, \wb_{t+1})\| x_{ti} \|_{H_t^{-1}}
        \\
        &+ \sum_{i \in S_t}
          p_t(i | S_t, \xib_t) \| x_{ti} \|_{H_t^{-1}}
          \sum_{j \in S_t} p_t(j | S_t, \xib_t) \left| x_{tj}^\top  
        (\wb^\star - \wb_{t+1}) \right|
        p_t(0 | S_t, \wb_{t+1})
        \\
        &\leq \sum_{i \in S_t}   p_t(i | S_t, \xib_t)  \| x_{ti} \|_{H_t^{-1}}^2 \|\wb^\star - \wb_{t+1}\|_{H_t}
        + \left( \sum_{i \in S_t}
          p_t(i | S_t, \xib_t) \| x_{ti} \|_{H_t^{-1}} \right)^2  \|\wb^\star - \wb_{t+1}\|_{H_t}.
        \\
    \end{align*}
    Then, since $x_{t0} = \mathbf{0}$, we can further bound the right-hand side as:
    \begin{align*}
        &\sum_{i \in S_t}   p_t(i | S_t, \xib_t)  \| x_{ti} \|_{H_t^{-1}}^2 \|\wb^\star - \wb_{t+1}\|_{H_t}
        + \left( \sum_{i \in S_t}
          p_t(i | S_t, \xib_t) \| x_{ti} \|_{H_t^{-1}} \right)^2  \|\wb^\star - \wb_{t+1}\|_{H_t}
        \\
        &= \sum_{i \in S_t }   p_t(i | S_t, \xib_t)  \| x_{ti} \|_{H_t^{-1}}^2 \|\wb^\star - \wb_{t+1}\|_{H_t}
        + \left( \sum_{i \in S_t  \cup \{0\} }
          p_t(i | S_t, \xib_t) \| x_{ti} \|_{H_t^{-1}} \right)^2  \|\wb^\star - \wb_{t+1}\|_{H_t}
        \\
        &\leq \sum_{i \in S_t}   p_t(i | S_t, \xib_t)  \| x_{ti} \|_{H_t^{-1}}^2 \|\wb^\star - \wb_{t+1}\|_{H_t}
        + \sum_{i \in S_t}
          p_t(i | S_t, \xib_t) \| x_{ti} \|_{H_t^{-1}}^2   \|\wb^\star - \wb_{t+1}\|_{H_t}
        \\
        &\leq 2\sum_{i \in S_t}   p_t(i | S_t, \xib_t)  \| x_{ti} \|_{H_t^{-1}}^2 \|\wb^\star - \wb_{t+1}\|_{H_t}
        \\
        &\leq 2 \beta_t(\delta) \sum_{i \in S_t}   p_t(i | S_t, \xib_t)  \| x_{ti} \|_{H_t^{-1}}^2
        \leq 2 \beta_t(\delta) \max_{i \in S_t} \| x_{ti} \|_{H_t^{-1}}^2
        ,
    \end{align*}
    where the first inequality holds due to Jensen's inequality and the second-to-last inequality holds by Lemma~\ref{lemma:online_confidence_set}.
    Hence, we get
    \begin{align*}
        \sum_{t=1}^T  \sum_{i \in S_t} \left( p_t(i | S_t, \wb^\star) - p_t(i | S_t, \wb_{t+1}) \right) p_t(0 | S_t, \wb_{t+1})  \|x_{ti} \|_{H_t^{-1}}
        &\leq 2 \beta_T(\delta)  \sum_{t=1}^T \max_{i \in S_t} \| x_{ti} \|_{H_t^{-1}}^2
        \\
        &\leq \frac{4d }{\kappa}  \beta_T(\delta) \log \left( 1+ \frac{T}{d \lambda} \right)
        \numberthis \label{eq:bouns_first_decomposition_2}
    \end{align*}
    where the last inequality holds by Lemma~\ref{lemma:elliptical}.
    
    Finally, we bound the third term on the right-hand side of Equation~\eqref{eq:bouns_first_decomposition}.
    By the mean value theorem, there exists $\xib'_t = (1-c') \wb^\star + c' \wb_{t+1}$ for some $c' \in (0,1)$ such that
    \begin{align*}
         &\sum_{i \in S_t} p_t(i | S_t, \wb^\star) \left( p_t(0 | S_t, \wb^\star) -  p_t(0 | S_t, \wb_{t+1}) \right)  \|x_{ti} \|_{H_t^{-1}}
        \\
         &= \sum_{i \in S_t}  p_t(i | S_t, \wb^\star) 
        \nabla p_t(0 | S_t, \xib'_t)^\top (\wb^\star - \wb_{t+1})  \| x_{ti} \|_{H_t^{-1}}
        \\
         &= - \sum_{i \in S_t} p_t(i | S_t, \wb^\star) 
         p_t(0 | S_t, \xib'_t) \sum_{j \in S_t}  p_t(j | S_t, \xib'_t) x_{tj}^\top  (\wb^\star - \wb_{t+1})  \| x_{ti} \|_{H_t^{-1}}
        \\
         &\leq \sum_{i \in S_t} p_t(i | S_t, \wb^\star) \| x_{ti} \|_{H_t^{-1}}
         p_t(0 | S_t, \xib'_t) \sum_{j \in S_t}  p_t(j | S_t, \xib'_t) \|x_{tj}\|_{H_t^{-1}}  \|\wb^\star - \wb_{t+1}\|_{H_t}
         \\
         &\leq \sum_{i \in S_t} p_t(i | S_t, \wb^\star) \| x_{ti} \|_{H_t^{-1}}
          \sum_{j \in S_t}  p_t(j | S_t, \xib'_t) \|x_{tj}\|_{H_t^{-1}}  \|\wb^\star - \wb_{t+1}\|_{H_t}
          \\
         &\leq  \beta_t(\delta) \sum_{i \in S_t} p_t(i | S_t, \wb^\star) \| x_{ti} \|_{H_t^{-1}}
          \sum_{j \in S_t}  p_t(j | S_t, \xib'_t) \|x_{tj}\|_{H_t^{-1}} 
          \\
          &\leq  \beta_t(\delta) \left( \max_{i \in S_t} \| x_{ti} \|_{H_t^{-1}}\right)^2 
          = \beta_t(\delta)  \max_{i \in S_t} \| x_{ti} \|_{H_t^{-1}}^2, 
    \end{align*}
    where the third inequality holds by Lemma~\ref{lemma:online_confidence_set}, and the last inequality holds since $(\max_i a_i)^2 = \max_i a_i^2$ for any $a_i \geq 0$.
    Therefore, we have
    \begin{align*}
        \sum_{t=1}^T \sum_{i \in S_t} p_t(i | S_t, \wb^\star) \left( p_t(0 | S_t, \wb^\star) -  p_t(0 | S_t, \wb_{t+1}) \right)  \|x_{ti} \|_{H_t^{-1}}
        &\leq \beta_T(\delta) \sum_{t=1}^T   \max_{i \in S_t} \| x_{ti} \|_{H_t^{-1}}^2
        \\
        &\leq \frac{2d}{\kappa} \beta_T(\delta)  \log \left( 1+ \frac{T}{d \lambda} \right),
        \numberthis \label{eq:bouns_first_decomposition_3}
    \end{align*}
     where the last inequality holds by Lemma~\ref{lemma:elliptical}.
     By plugging Equation~\eqref{eq:bouns_first_decomposition_1},~\eqref{eq:bouns_first_decomposition_2}, and~\eqref{eq:bouns_first_decomposition_3} into Equation~\eqref{eq:bouns_first_decomposition} and multiplying $2  \beta_T(\delta)$, we get
     \begin{align*}
          2  \beta_T(\delta) &\sum_{t=1}^T \sum_{i \in S_t} p_t(i | S_t, \wb^\star) p_t(0 | S_t, \wb^\star)  \|x_{ti} \|_{H_t^{-1}} 
          \\
          &\leq 2\sqrt{2}  \frac{\sqrt{v_0 K}}{(v_0 + K e^{-1})}  \beta_T(\delta) \sqrt{dT}  \sqrt{  \log \left( 1+ \frac{T}{d \lambda} \right) }
          + \frac{12 d}{\kappa}   \beta_T(\delta)^2    \log \left( 1+ \frac{T}{d \lambda} \right)
          .
          \numberthis \label{eq:regret_decomposition_1}
     \end{align*}
    Moreover, by applying Lemma~\ref{lemma:elliptical}, we can directly bound the second term of Equation~\eqref{eq:regret_decomposition}.
    \begin{align*}
        10 \beta_T(\delta)^2 \sum_{t=1}^T    \max_{i \in S_t }  \| x_{ti} \|_{H_t^{-1}}^2
        \leq  10 \beta_T(\delta)^2 \cdot \frac{2}{\kappa} d \log \left( 1+ \frac{T}{d \lambda} \right).
        \numberthis \label{eq:regret_decomposition_2}
    \end{align*}
    Finally, plugging Equation~\eqref{eq:regret_decomposition_1} and~\eqref{eq:regret_decomposition_2} into Equation~\eqref{eq:regret_decomposition}, we obtain
    \begin{align*}
        \Regret(\wb^\star)
        &\leq 2\sqrt{2}  \frac{\sqrt{v_0 K}}{(v_0 + K e^{-1})}  \beta_T(\delta) \sqrt{dT}  \sqrt{  \log \left( 1+ \frac{T}{d \lambda} \right) }
        + \frac{32d}{\kappa} \beta_T(\delta)^2   \log \left( 1+ \frac{T}{d \lambda} \right)
        \\
        &= \BigOTilde \left ( \frac{\sqrt{v_0 K}}{v_0 + K }  d\sqrt{T}  + \frac{1}{\kappa}d^2 \right),
    \end{align*}
    where $\beta_{T}(\delta) =\BigO \left( \sqrt{d} \log T \log K \right)$.
    This concludes the proof of Theorem~\ref{thm:upper_bound}.
\end{proof}

\begin{remark} \label{remark:B-bound_upper_uniform}
    If the boundedness assumption on the parameter is relaxed to
    $\| \wb \|_2 \leq B$,
    since
    $\beta_t(\delta)
    = \BigO \left( B \sqrt{d} \log t \log K  
        + B^{3/2} \sqrt{d \log K }
        \right)$ (refer Corollary~\ref{corollary:B-bound_confidence}),
    we have 
    $\Regret(\wb^\star)
    = \BigOTilde \left ( B^{3/2} e^B \frac{\sqrt{v_0 K}}{v_0 + K }  d\sqrt{T}  + \frac{1}{\kappa} B^3 d^2 \right).
    $
    It's important to note that one of our main goals is to explicitly demonstrate the regret depends on $K$ and $v_0$. 
    In deriving such a result, the dependence on  $e^B$ is unavoidable to our best knowledge. 
    Note that for non-uniform rewards, the regret bound does not depend on $e^B$ (refer Remark~\ref{remark:B-bound_upper_non-uniform}).
\end{remark}

\subsection{Proofs of Lemmas for Theorem~\ref{thm:upper_bound}} 
\label{app_subsec:useful_lemmas_thm1}

\subsubsection{Proof of Lemma~\ref{lemma:utility}}
\label{app_subsubsec:proof_of_lemma:utility}
\begin{proof}[Proof of Lemma~\ref{lemma:utility}]
Under the condition  $\wb^\star \in \mathcal{C}_t(\delta)$, we have
    \begin{align*}
        \left| x_{ti}^\top \wb_{t}  - x_{ti}^\top \wb^\star \right|
        \leq \| x_{ti} \|_{H_t^{-1}} \| \wb_{t}- \wb^\star \|_{H_t}
        \leq \beta_t(\delta) \| x_{ti} \|_{H_t^{-1}},
    \end{align*}
    where the first inequality is by the Hölder's inequality, and the last inequality holds by Lemma~\ref{lemma:online_confidence_set}.
    Hence, it follows that
    \begin{align*}
        \alpha_{ti} - x_{ti}^\top \wb^\star
        =  x_{ti}^\top \wb_{t} - x_{ti}^\top \wb^\star + \beta_t(\delta) \| x_{ti} \|_{H_t^{-1}} 
        \leq 2 \beta_t(\delta) \| x_{ti} \|_{H_t^{-1}}.
    \end{align*}
    Moreover, from $ x_{ti}^\top \wb_{t}  - x_{ti}^\top \wb^\star \geq -\beta_t(\delta) \| x_{ti} \|_{H_t^{-1}}$, we also have
    \begin{align*}
        \alpha_{ti} - x_{ti}^\top \wb^\star 
        =  x_{ti}^\top \wb_{t} - x_{ti}^\top \wb^\star + \beta_t(\delta) \| x_{ti} \|_{H_t^{-1}}
        \geq 0.
    \end{align*}
    This concludes the proof.
\end{proof}

\subsubsection{Proof of Lemma~\ref{lemma:elliptical}}
\label{app_subsubsec:proof_of_lemma:elliptical}
\begin{proof}[Proof of Lemma~\ref{lemma:elliptical}]
    Since $x x^\top + y y^\top \succeq xy^\top + yx^\top$ for any $x,y \in \RR^d$, it follows that
    \begin{align*}
        &\Gcal_s(\wb_{s+1}) 
        \\
        &= \sum_{i \in S_s} p_s(i | S_s, \wb_{s+1}) x_{si} x_{si}^\top -  \sum_{i \in S_s}  \sum_{j \in S_s} p_s(i | S_s, \wb_{s+1}) p_s(j | S_s, \wb_{s+1}) x_{si} x_{sj}^\top
        \\
        &= \sum_{i \in S_s} p_s(i | S_s, \wb_{s+1}) x_{si} x_{si}^\top 
        - \frac{1}{2} \sum_{i \in S_s}  \sum_{j \in S_s} p_s(i | S_s, \wb_{s+1}) p_s(j | S_s, \wb_{s+1}) (x_{si} x_{sj}^\top + x_{sj} x_{si}^\top )
        \\
        &\succeq \sum_{i \in S_s} p_s(i | S_s, \wb_{s+1}) x_{si} x_{si}^\top 
        - \frac{1}{2} \sum_{i \in S_s}  \sum_{j \in S_s} p_s(i | S_s, \wb_{s+1}) p_s(j | S_s, \wb_{s+1}) (x_{si} x_{si}^\top + x_{sj} x_{sj}^\top )
        \\
        &= \sum_{i \in S_s} p_s(i | S_s, \wb_{s+1}) x_{si} x_{si}^\top 
        -  \sum_{i \in S_s}  \sum_{j \in S_s} p_s(i | S_s, \wb_{s+1}) p_s(j | S_s, \wb_{s+1}) x_{si} x_{si}^\top.
    \end{align*}
    Hence, we have
    \begin{align*}
        \Gcal_s(\wb_{s+1}) 
        &\succeq \sum_{i \in S_s} p_s(i | S_s, \wb_{s+1}) \left( 1 -  \sum_{j \in S_s}p_s(j | S_s, \wb_{s+1}) \right) x_{si} x_{si}^\top 
        \\
        &=\sum_{i \in S_s} p_s(i | S_s, \wb_{s+1}) p_s(0 | S_s, \wb_{s+1})  x_{si} x_{si}^\top 
        \numberthis \label{eq:hessian_for_elliptical_lowerbound},
    \end{align*}
    which implies that
    \begin{align*}
        H_{t+1} \succeq H_t +  \sum_{i \in S_t} p_t(i | S_t, \wb_{t+1}) p_t(0 | S_t, \wb_{t+1})  x_{ti} x_{ti}^\top.
    \end{align*}
    Then, we get
    \begin{align*}
        \det \left( H_{t+1} \right)
        \geq \det \left( H_t \right) \left( 1 +  \sum_{i \in S_t}  p_t(i | S_t, \wb_{t+1}) p_0(i | S_t, \wb_{t+1}) \| x_{ti} \|_{H_{t}^{-1}}^2 \right).
    \end{align*}
    Since $\lambda \geq 1$, for all $t \geq 1$, we have $\sum_{i \in S_t}  p_t(i | S_t, \wb_{t+1}) p_0(i | S_t, \wb_{t+1}) \| x_{ti} \|_{H_{t}^{-1}}^2 \leq 1$.
    Then, using the fact that $z \leq 2 \log (1 + z)$ for any $z \in [0,1]$, we get
    \begin{align*}
        \sum_{s=1}^{t} \sum_{i \in S_s} & p_s(i | S_s, \wb_{s+1}) p_s(0 | S_s, \wb_{s+1}) \| x_{si}\|_{H_s^{-1}}^2
        \\
        &\leq 2 \sum_{s=1}^{t} \log \left( 1 + p_s(i | S_s, \wb_{s+1})  p_s(0 | S_s, \wb_{s+1}) \| x_{si}\|_{H_s^{-1}}^2  \right)
        \\
        &\leq 2 \sum_{s=1}^{t} \log \left( \frac{\det (H_{s+1})}{\det(H_s)} \right)
        \\
        &\leq 2 d \log \left( \frac{\operatorname{tr}(H_{t+1})}{d \lambda} \right)
        \leq 2d \log \left(1 + \frac{t}{d \lambda} \right).
    \end{align*}
    This proves the first inequality.

    To establish the proof for the second inequality,
    we return to Equation~\eqref{eq:hessian_for_elliptical_lowerbound}:
    \begin{align*}
        \Gcal_s(\wb_{s+1}) 
        &\succeq \sum_{i \in S_s} p_s(i | S_s, \wb_{s+1}) p_s(0 | S_s, \wb_{s+1})  x_{si} x_{si}^\top 
        \succeq \kappa  \sum_{i \in S_s} x_{si} x_{si}^\top,
    \end{align*}
    which implies that 
    \begin{align*}
        H_{t+1}
        = H_t + \Gcal_t(\wb_{t+1}) 
        \succeq H_t + \kappa \sum_{i \in S_t}  x_{ti}x_{ti}^\top
        .
    \end{align*}
    Since $\lambda \geq 1$, for all $t \geq 1$, we have $\kappa \max_{i \in S_t}   \| x_{ti} \|_{H_{t}^{-1}}^2 \leq \kappa$.
    We then conclude on the same way:
    \begin{align*}
        \sum_{s=1}^{t} \max_{i \in S_s} \| x_{si}\|_{H_s^{-1}}^2 
        &\leq \frac{2}{\kappa} \sum_{s=1}^{t} \log \left( 1 + \kappa\max_{i \in S_s} \| x_{si}\|_{H_s^{-1}}^2 \right)
        \\
        &\leq \frac{2}{\kappa} \sum_{s=1}^{t} \log \left( \frac{\det (H_{s+1})}{\det(H_s)} \right)
        \leq \frac{2}{\kappa} d \log \left( 1+ \frac{t}{d \lambda} \right),
    \end{align*}
    which proves the second inequality.
\end{proof}

\subsubsection{Proof of Lemma~\ref{lemma:revenue_second_pd}}
\label{app_subsubsec:proof_of_lemma:revenue_second_pd}
    \begin{proof}[Proof of Lemma~\ref{lemma:revenue_second_pd}]
        Let $i,j \in [K]$. 
        We first have
        \begin{align*}
            \frac{\partial Q}{\partial i} 
            = \frac{e^{u_i}}{v_0 + \sum_{k=1}^K e^{u_k}}
            - \frac{e^{u_i} \left( \sum_{k=1}^K e^{u_k} \right)  }{(v_0 + \sum_{k=1}^K e^{u_k})^2}
        \end{align*}
        Then, we get
        \begin{align*}
            &\frac{\partial^2 Q}{\partial i \partial j}    
            \\
            &= \frac{\mathbbm{1}_{i=j}e^{u_i}}{v_0 + \sum_{k=1}^K e^{u_k}}
            - \frac{e^{u_i} e^{u_j}}{(v_0 + \sum_{k=1}^K e^{u_k})^2}
            - \frac{\mathbbm{1}_{i=j}e^{u_i}  \left( \sum_{k=1}^K e^{u_k} \right) + e^{u_i}e^{u_j} }{(v_0 + \sum_{k=1}^K e^{u_k})^2}
            \\
            &+ \frac{e^{u_i}  \left( \sum_{k=1}^K e^{u_k} \right) 2e^{u_j} \left(v_0 + \sum_{k=1}^K e^{u_k} \right)  }{(v_0 + \sum_{k=1}^K e^{u_k})^4}
            \\
            &= \frac{\mathbbm{1}_{i=j}e^{u_i}}{v_0 + \sum_{k=1}^K e^{u_k}}
            - \frac{e^{u_i} e^{u_j}}{(v_0 + \sum_{k=1}^K e^{u_k})^2}
            - \frac{\mathbbm{1}_{i=j}e^{u_i}  \left( \sum_{k=1}^K e^{u_k} \right) \!+\! e^{u_i}e^{u_j} }{(v_0 + \sum_{k=1}^K e^{u_k})^2}
            + \frac{e^{u_i}  \left( \sum_{k=1}^K e^{u_k} \right) 2e^{u_j} }{(v_0 + \sum_{k=1}^K e^{u_k})^3}.
        \end{align*}
        Let $p_i(\ub) = \frac{e^{u_i}}{v_0 + \sum_{k=1}^K e^{u_k}}$ 
        and $p_0(\ub) = \frac{v_0}{v_0 + \sum_{k=1}^K e^{u_k}}$ .
        For $i=j$, we have
        \begin{align*}
            \left| \frac{\partial^2 Q}{\partial i \partial j} \right|
            &= \bigg| p_i(\ub) - p_i(\ub) p_j(\ub) 
            - p_i(\ub) \frac{ \sum_{k=1}^K e^{u_k} }{v_0 + \sum_{k=1}^K e^{u_k}}
            - p_i(\ub) p_j(\ub)
            \\
            &+ 2 p_i(\ub) p_j(\ub) \frac{ \sum_{k=1}^K e^{u_k} }{v_0 + \sum_{k=1}^K e^{u_k}} \bigg|
            \\
            &= \bigg| p_i(\ub)p_0(\ub) - 2p_i(\ub) p_j(\ub) + 2 p_i(\ub) p_j(\ub) \frac{ \sum_{k=1}^K e^{u_k} }{v_0 + \sum_{k=1}^K e^{u_k}} \bigg|
            \\
            &= \bigg| p_i(\ub)p_0(\ub)  - 2p_i(\ub) p_j(\ub) p_0(\ub) \bigg|
            \\
            &\leq 3 p_i(\ub)
        \end{align*}
        For $i \neq j$, we have
        \begin{align*}
            \left| \frac{\partial^2 Q}{\partial i \partial j} \right|
            &= \bigg| - p_i(\ub) p_j(\ub) - p_i(\ub) p_j(\ub) 
            +2 p_i(\ub) p_j(\ub) \frac{ \sum_{k=1}^K e^{u_k} }{v_0 + \sum_{k=1}^K e^{u_k}} \bigg|
            \\
            &= \bigg| - 2p_i(\ub) p_j(\ub)p_0(\ub)  \bigg|
            \\
            &\leq 2p_i(\ub) p_j(\ub).
        \end{align*}
        This concludes the proof.
    \end{proof}
%

\section{Proof of Lemma~\ref{lemma:online_confidence_set}} 
\label{app_sec:proof_lemma_confidence}
In this section, we provide the proof of Lemma~\ref{lemma:online_confidence_set}.
First, we present the main proof of Lemma~\ref{lemma:online_confidence_set}, followed by the proof of the technical lemma utilized within the main proof.
\subsection{Main Proof of Lemma~\ref{lemma:online_confidence_set}} \label{app_subsec:main_proof_lemma_confidence}
\begin{proof} [Proof of Lemma~\ref{lemma:online_confidence_set}]
The proof is similar to the analysis presented in~\citet{zhang2024online}. 
However, their MNL choice model is constructed using a shared context  $x_t$ and varying parameters across the choices $\wb_{1}^\star, \dots, \wb_K^\star$, whereas our approach considers an MNL choice model that shares the parameter $\wb^\star$ across the choices and has varying contexts for each item in the assortment $S$, $x_{t1}, \dots x_{ti_{|S|}}$. 
Moreover,~\citet{zhang2024online} only consider a fixed assortment size, whereas we consider a more general setting where the assortment size can vary in each round $t$.
We denote $K_t = |S_t|$ in the proof of Lemma~\ref{lemma:online_confidence_set}.
Note that $K_t \leq K$ for all $t \geq 1$.
    \begin{lemma} \label{lemma:online_parameter_gap_bound}
        Let the update rule be
        \begin{align*}
            \wb_{t+1} = \argmin_{\wb \in \mathcal{W}} \tilde{\ell}_{t}(\wb) + \frac{1}{2 \eta} \| \wb - \wb_{t} \|_{H_{t}}^2,
        \end{align*}
        where $\tilde{\ell}_{t}(\wb) = \ell_{t}(\wb_{t}) + \langle \wb - \wb_{t}, \nabla \ell_{t}(\wb_{t}) \rangle + \frac{1}{2} \|\wb - \wb_{t} \|^2_{\nabla^2 \ell_{t}(\wb_{t})}$ and $H_t = \lambda \Ib_d + \sum_{s=1}^{t-1} \Gcal_s(\wb_{s+1})$.
        Let $\eta = \frac{1}{2} \log (K+1) + 2$ and $\lambda > 0$.
        Then, we have
        \begin{align*}
            \| \wb_{t+1} - \wb^\star \|_{H_{t+1}}^2 
            &\leq  2 \eta\left(  \sum_{s=1}^t \ell_{s} (\wb^\star) - \sum_{s=1}^t \ell_{s}(\wb_{s+1})\right)
        + 4 \lambda 
        + 12\sqrt{2}  \eta  \sum_{s=1}^t \left\| \wb_{s+1} - \wb_{s} \right\|_2^2
        \\
        &- \sum_{s=1}^t \left\|  \wb_{s+1} - \wb_{s} \right\|_{H_{s}}^2.
            \numberthis \label{eq:online_parameter_gap_bound}
        \end{align*}
    \end{lemma}
We first bound the first term in Equation~\eqref{eq:online_parameter_gap_bound}.
For simplicity, we define the softmax function at round $t$ $\sigmab_t(\zb): \RR^{K_t} \rightarrow \RR^{K_t}$ as follows:
\begin{align}
    [\sigmab_t(\zb)]_{i} = \frac{\exp([\zb]_{i})}{ v_0+ \sum_{k =1}^{K_t}   \exp([\zb]_{k})}, \quad \forall i \in [K_t], 
    \label{eq:softmax}
\end{align}
where $[\cdot]_{i}$ denotes $i$'th element of the input vector. 
We denote the probability of choosing the outside option as $[\sigmab_t(\zb)]_{0} = \frac{v_0}{ v_0+ \sum_{k =1}^{K_t}   \exp([\zb]_{k})}$.
Although $[\sigmab_t(\zb)]_{0}$ is not the output of the softmax function $\sigmab_t(\zb)$, we represent it in a form similar to that in Equation~\eqref{eq:softmax} for simplicity.
Then, the user choice model in Equation~\eqref{eq:mnl_model} can be equivalently expressed as $p_t(i | S_t, \wb) = \left[\sigmab_t\left( (x_{tj}^\top \wb)_{j \in S_t}  \right)\right]_{i}$ for all $i \in [K_t]$ and $p_t(0 | S_t, \wb) = \left[\sigmab_t\left( (x_{tj}^\top \wb)_{j \in S_t}  \right)\right]_{0}$.
Furthermore, the loss function in  Equation~\eqref{eq:loss} can also be written as $\ell(\zb_t, \yb_t) = \sum_{k=0}^{K_t} \mathbf{1}\left\{ y_{ti} = 1 \right\} \cdot \log\left(\frac{1}{[\sigmab_t(\zb_t)]_k}\right)$.

Define a pseudo-inverse function of $\sigmab_t(\cdot)$ as $\sigmab_t^+ : \RR^{K_t} \rightarrow \RR^{K_t}$, where $[\sigmab_t^+(\qb)]_i = \log \left( q_i / (1- \| \qb \|_1)\right)$ for any $\qb \in \{ \pb \in [0,1]^{K_t} \mid \| \pb \|_1 < 1 \}$.
Then, inspired by the previous studies on binary
logistic bandit~\citep{faury2022jointly}, we decompose the regret into two terms by introducing an intermediate term.
\begin{align}
     \sum_{s=1}^t \ell_{s} (\wb^\star) - \sum_{s=1}^t \ell_{s}(\wb_{s+1})
     = \underbrace{\sum_{s=1}^t \ell_{s} (\wb^\star) 
     - \sum_{s=1}^t \ell (\tilde{\zb}_s, \yb_s)}_{(a)}
     + \underbrace{\sum_{s=1}^t \ell (\tilde{\zb}_s, \yb_s) 
     - \sum_{s=1}^t \ell_{s}(\wb_{s+1})}_{(b)},
     \label{eq:online_regret_intermediate}
\end{align}
where $\tilde{\zb}_s := \sigmab_s^+ \left( \EE_{\wb \sim P_s} \left[\sigmab_s\left( (x_{sj}^\top \wb)_{j \in S_s} \right) \right] \right)$, and $P_s := \mathcal{N}( \wb_s, cH_s^{-1} )$ is the Gaussian distribution with mean $\wb_s$ and covariance matrix $cH_s^{-1}$, where $c >0$ is a positive constant to be specified later.
We first show that the term $(a)$ is bounded by $\BigO\left( \log K (\log t)^2 \right)$ with high probability.
\begin{lemma} \label{lemma:online_regret_intermediate_(a)}
    Let $\delta \in (0,1]$. 
    Under Assumptions~\ref{assum:bounded_assumption}, for all $ t \in [T]$, with probability at least $1-\delta$, we have
    \begin{align*}
        \sum_{s=1}^t   \ell_{s} (\wb^\star) - \sum_{s=1}^t \ell(\tilde{\zb}_{s}, \yb_s) 
        &\leq 11 \cdot
        \left( 3\log ( 1 + (K + 1)t) + 3 \right) 
          \log \left( \frac{2\sqrt{1 + 2t}}{\delta} \right)   
        + 2.
    \end{align*}
\end{lemma}
Furthermore, we can bound the term $(b)$ by the following lemma.
\begin{lemma} \label{lemma:bound_(b)}
    For any $c > 0$, let $\lambda \geq \max\{ 2,  72 c d \}$. 
    Then, under Assumption~\ref{assum:bounded_assumption}, for all $t \geq 1$, we have
    \begin{align*}
        \sum_{s=1}^t \left( \ell(\tilde{\zb}_{s} , \yb_s) -  \ell_{s}(\wb_{s+1})\right)
        \leq \frac{1}{2c} \sum_{s=1}^t \| \wb_{s} - \wb_{s+1} \|_{H_{s}}^2  
        + \sqrt{6}c  d \log \left( 1 + \frac{t+1}{2\lambda}\right).
    \end{align*}
\end{lemma}
Now, we are ready to prove the Lemma~\ref{lemma:online_confidence_set}.
By combining Lemma~\ref{lemma:online_parameter_gap_bound}, Lemma~\ref{lemma:online_regret_intermediate_(a)}, and Lemma~\ref{lemma:bound_(b)}, we derive that
\begin{align*}
    &\| \wb_{t+1} - \wb^\star \|_{H_{t+1}}^2 
    \\
    &\leq 2 \eta\Bigg[\! 
    11 \cdot
        \left( 3\log ( 1 + (K + 1)t) + 3 \right) 
          \log \left( \frac{2\sqrt{1 + 2t}}{\delta} \right)   
        + 2
    + \sqrt{6}c  d \log \left( 1 + \frac{t+1}{2\lambda}\right)\!
    \Bigg]
    \\
    &+ \!4 \lambda\! 
    + 12\sqrt{2}  \eta  \sum_{s=1}^t \!\left\| \wb_{s+1} - \wb_{s} \right\|_2^2
    + \left( \frac{\eta}{c} \!-\!1 \right) \sum_{i=1}^t \left\|  \wb_{s+1} - \wb_{s} \right\|_{H_{s}}^2
    \\
    &\leq 2 \eta\Bigg[\! 
    11 \cdot
        \left( 3\log ( 1 + (K + 1)t) + 3 \right) 
          \log \left( \frac{2\sqrt{1 + 2t}}{\delta} \right)   
        + 2
    + \sqrt{6}c  d \log \left( 1 + \frac{t+1}{2\lambda}\right)\!
    \Bigg]
    \\
    &+ \!4 \lambda\! =: \beta'_{t+1}(\delta)^2 = \BigO \left(d (\log t \log K)^2 \right)
    \numberthis \label{eq:def_beta'}
    , 
\end{align*}
where the second inequality holds because by setting $c=7\eta/6$ and  $\lambda \geq \max \{84\sqrt{2} \eta, 84d\eta \}$, we obtain: 
\begin{align*}
    12\sqrt{2}  \eta  \sum_{s=1}^t & \!\left\| \wb_{s+1} - \wb_{s} \right\|_2^2
    + \left( \frac{\eta}{c} \!-\!1 \right) \sum_{s=1}^t \left\|  \wb_{s+1} - \wb_{s} \right\|_{H_{s}}^2
    \\
    &= 12\sqrt{2}  \eta  \sum_{s=1}^t \!\left\| \wb_{s+1} - \wb_{s} \right\|_2^2
    - \frac{1}{7} \sum_{s=1}^t \left\|  \wb_{s+1} - \wb_{s} \right\|_{H_{s}}^2
    \\
    &\leq \left( 12\sqrt{2}  \eta - \frac{\lambda}{7} \right)  \sum_{s=1}^t \!\left\| \wb_{s+1} - \wb_{s} \right\|_2^2
    \leq 0,
\end{align*}
where the first inequality holds since $H_s \succeq \lambda \mathbf{I}_d$.

By setting $\eta = \frac{1}{2} \log (K+1) +2$ and $\lambda = 84 \sqrt{2}d \eta$, we derive that
\begin{align*}
    \| \wb_{t} - \wb^\star \|_{H_{t}} 
    &\leq \| \wb_{t} - \wb_{t+1} \|_{H_{t}} 
    + \| \wb_{t+1} - \wb^\star \|_{H_{t}} 
    \\
    &\leq \| \wb_{t} - \wb_{t+1} \|_{H_{t}} 
    + \| \wb_{t+1} - \wb^\star \|_{H_{t+1}} 
    \\
    &\leq \| \wb_{t} - \wb_{t+1} \|_{H_{t}} 
    +  \beta'_{t+1}(\delta)
    \\
    &\leq \frac{2 \eta}{\lambda} 
    +  \beta'_{t+1}(\delta)
    := \beta_t(\delta)
    = \BigO \left( \sqrt{d} \log t \log K \right).
\end{align*}
where the third inequality follows from the definition of $\beta'_{t+1}(\delta)$ in Equation~\eqref{eq:def_beta'},
and the last inequality holds by Lemma~\ref{lemma:zhang_lemma20_improved}.

This concludes the proof of Lemma~\ref{lemma:online_confidence_set}.
\end{proof}

\begin{corollary} \label{corollary:B-bound_confidence}
    If the boundedness assumption on the parameter is relaxed to
    $\| \wb \|_2 \leq B$,
    then
    $\beta_t(\delta)
    = \BigO \left( B \sqrt{d} \log t \log K  
        + B^{3/2} \sqrt{d \log K }
        \right)$.
\end{corollary}
\begin{proof}[Proof of Corollary~\ref{corollary:B-bound_confidence}]
    When $\| \wb \|_2 \leq B$, following the same analysis as in Lemma~\ref{lemma:online_confidence_set}, we can set $\eta = \frac{1}{2} \log (K+1) + B + 1$,
    $c = 7 \eta/6$, and
    $\lambda = \BigO(dB \log K) $.
    Under these settings, we have
    \begin{align*}
        &\| \wb_{t+1} - \wb^\star \|_{H_{t+1}} 
        \\
        &\leq \sqrt{ 2 \eta\Bigg[\! 
        11 \cdot
            \left( 3\log ( 1 + (K + 1)t) + B + 2 \right) 
              \log \left( \frac{2\sqrt{1 + 2t}}{\delta} \right)   
            + 2
        + \sqrt{6}c  d \log \left( 1 + \frac{t+1}{2\lambda}\right)\!
        \Bigg] + \!4 B^2 \lambda\!
        }
        \\
        &=\BigO \left( B \sqrt{d} \log t \log K  
        + B^{3/2} \sqrt{d \log K }
        \right). 
    \end{align*}
    which concludes the proof.
\end{proof}

\subsection{Proofs of Lemmas for Lemma~\ref{lemma:online_confidence_set}} 
\label{app_subsec:proof_lemma:online_confidence_set}
\subsubsection{Proof of Lemma~\ref{lemma:online_parameter_gap_bound}  }
\label{app_subsubsec:proof_lemma:online_parameter_gap_bound}
\begin{proof}[Proof of Lemma~\ref{lemma:online_parameter_gap_bound}]
    Let $\tilde{\ell}_{s}(\wb) = \ell_{s}(\wb_{s}) + \langle \nabla \ell_{s}(\wb_{s}), \wb - \wb_{s}  \rangle + \frac{1}{2} \|\wb - \wb_{s} \|^2_{\nabla^2 \ell_{s}(\wb_{s})}$ be a second-order approximation of the original function $\ell_{s}(\wb)$ at the point $\wb_{s}$.
    The update rule in Equation~\eqref{eq:online_update} can also be expressed as
    \begin{align*}
        \wb_{s+1} = \argmin_{\wb \in \mathcal{W}} \tilde{\ell}_{s}(\wb) + \frac{1}{2 \eta} \| \wb - \wb_{s} \|_{H_{s}}^2.
    \end{align*}
    Then, by Lemma~\ref{lemma:loss_firstorder_decomposition}, we have
    \begin{align}
        \langle \nabla \tilde{\ell}_{s}(\wb_{s+1}), \wb_{s+1} - \wb^\star  \rangle
        \leq \frac{1}{2\eta} \left( \| \wb_{s} - \wb^\star  \|_{H_{s}}^2 - \| \wb_{s+1} - \wb^\star \|_{H_{s}}^2 - \| \wb_{s+1} - \wb_{s} \|_{H_{s}}^2  \right).
        \label{eq:concentration_firstorder_bound}
    \end{align}
    To utilize Lemma~\ref{lemma:zhang_lemma1}, we can rewrite the loss function as $\ell\left((x_{si}^\top \wb )_{i \in S_{s}}, \yb_s \right) = \ell_{s}(\wb)$.
    Consequently, according to Lemma~\ref{lemma:zhang_lemma1}, it follows that
    \begin{align}
        \ell_{s}(\wb_{s+1}) - \ell_{s}(\wb^\star)
        \leq \langle \nabla \ell_{s}(\wb_{s+1}), \wb_{s+1} - \wb^\star  \rangle
        -\frac{1}{\zeta} \| \wb_{s+1} - \wb^\star \|^2_{\nabla^2 \ell_{s}(\wb_{s+1})},
        \label{eq:concentration_lossgap_bound}
    \end{align}
    where $\zeta =  \log (K+1) + 4$.
    Then, by combining Equation~\eqref{eq:concentration_firstorder_bound} and~\eqref{eq:concentration_lossgap_bound}, we have
    \begin{align*}
        \ell_{s}(\wb_{s+1}) - \ell_{s}(\wb^\star)
        &\leq \langle \nabla  \ell_{s}(\wb_{s+1})- \nabla\tilde{\ell}_{s}(\wb_{s+1}) , \wb_{s+1} - \wb^\star  \rangle
        \\
        &+ \frac{1}{\zeta} \left( \| \wb_{s} - \wb^\star \|_{H_{s}}^2 
        - \| \wb_{s+1} - \wb^\star \|_{H_{s+1}}^2 
        - \| \wb_{s+1} - \wb_{s} \|_{H_{s}}^2  \right).            
    \end{align*}
    In above, we can further bound the first term of the right-hand side as:
    \begin{align*}
        \langle \nabla  \ell_{s}(\wb_{s+1}) - &\nabla\tilde{\ell}_{s}(\wb_{s+1}) , \wb_{s+1} - \wb^\star  \rangle
        \\
        &= \langle \nabla  \ell_{s}(\wb_{s+1}) 
        - \nabla \ell_{s} (\wb_{s}) 
        - \nabla^2 \ell_{s} (\wb_{s}) (\wb_{s+1} - \wb_{s})
        , \wb_{s+1} - \wb^\star
        \rangle
        \\
        &= \langle D^3 \ell_{s}(\xib_{s+1}) [\wb_{s+1} - \wb_{s}] (\wb_{s+1} - \wb_{s})
        , \wb_{s+1} - \wb^\star \rangle
        \\
        &\leq 3\sqrt{2} \| \wb_{s+1}  - \wb^\star \|_2 \| \wb_{s+1} - \wb_{s}\|_{\nabla^2 \ell_{s}(\xib_{s+1} )}^2
        \\
        &\leq  6\sqrt{2}\| \wb_{s+1} - \wb_{s}\|_{\nabla^2 \ell_{s}(\xib_{s+1} )}^2
        \\
        &\leq 6\sqrt{2} \| \wb_{s+1} - \wb_{s}\|_2^2
    \end{align*}
    where in the second equality, we apply the Taylor expansion by introducing  $\xib_{s+1}$, a convex combination of $\wb_{s+1}$ and $\wb_{s}$. 
    The first inequality follows from Lemma~\ref{lemma:tran_thm3} and Proposition~\ref{prop:self_concordant},
    the second inequality holds by Assumption~\ref{assum:bounded_assumption}, and the last inequality holds because 
    \begin{align*}
        \nabla^2 \ell_{s}(\xib_{s+1} )
        &=\Gcal_{s}(\xib_{s+1}) 
        \\
        &= \sum_{i \in S_{s}} p_{s}(i | S_{s}, \xib_{s+1}) x_{si} x_{si}^\top 
        -  \sum_{i \in S_{s}}  \sum_{j \in S_{s}} p_{s}(i | S_{s}, \xib_{s+1}) p_{s}(j | S_{s}, \xib_{s+1}) x_{si} x_{sj}^\top
        \\
        &= \!\!\sum_{i \in S_{s} \!\cup \{0\}} p_{s}(i | S_{s}, \xib_{s+1}) x_{si} x_{si}^\top 
        - \!\! \sum_{i \in S_{s}\!\cup \{0\}}  \sum_{j \in S_{s}\!\cup \{0\}} p_{s}(i | S_{s}, \xib_{s+1}) p_{s}(j | S_{s}, \xib_{s+1}) x_{si} x_{sj}^\top
        \\
        &= \EE_{i \sim p_s(\cdot | S_, \xib_{s+1})} \left[ x_{si} x_{si}^\top \right]
        - \EE_{i \sim p_s(\cdot | S_s, \xib_{s+1})} \left[ x_{si} \right]
        \left(\EE_{i \sim p_s(\cdot | S_s, \xib_{s+1})} \left[ x_{si}\right]\right)^\top 
        \\
        &\preceq \EE_{i \sim p_s(\cdot | S_s, \xib_{s+1})} \left[ x_{si} x_{si}^\top \right]
        \preceq \Ib_d,
    \end{align*}
    where the third equality holds by setting $x_{s0} = \mathbf{0}$ for all $s \geq 1$.

    Now, by taking the summation over $s$ and rearranging the terms, we obtain
    \begin{align*}
        &\| \wb_{t+1} - \wb^\star \|_{H_{t+1}}^2 
        \\
        &\leq \zeta \left( \sum_{s=1}^t  \ell_{s} (\wb^\star) - \sum_{s=1}^t  \ell_{s}(\wb_{s+1})\right)
        + \| \wb_{1} - \wb^\star \|_{H_{1}}^2
        + 6\sqrt{2} \zeta  \sum_{s=1}^t \left\| \wb_{s+1} - \wb_{s} \right\|_2^2
        \\
        &- \sum_{s=1}^t \left\|  \wb_{s+1} - \wb_{s} \right\|_{H_{s}}^2
        \\
        &\leq  \zeta\left(  \sum_{s=1}^t \ell_{s} (\wb^\star) - \sum_{s=1}^t \ell_{s}(\wb_{s+1})\right)
        + 4 \lambda 
        + 6\sqrt{2} \zeta  \sum_{s=1}^t \left\| \wb_{s+1} - \wb_{s} \right\|_2^2
        - \sum_{s=1}^t \left\|  \wb_{s+1} - \wb_{s} \right\|_{H_{s}}^2,
    \end{align*}    
    where the last inequality holds since $\| \wb_{1} - \wb^\star \|_{H_{1}}^2 \leq \lambda  \| \wb_{1} - \wb^\star \|_2^2 \leq 4 \lambda$.
    Plugging in $\zeta = 2 \eta$, we conclude the proof.
\end{proof}
\subsubsection{Proof of Lemma~\ref{lemma:online_regret_intermediate_(a)}  }
\label{app_subsubsec:proof_lemma:online_regret_intermediate_(a)}
\begin{proof}[Proof of Lemma~\ref{lemma:online_regret_intermediate_(a)}]
    Since the norm of $\tilde{\zb}_s = \sigmab_s^+ \left( \EE_{\wb \sim P_s} \left[\sigmab_s\left( (x_{sj}^\top \wb)_{j \in S_s} \right) \right] \right)$ is unbounded in general, as suggested by~\citet{foster2018logistic}, we use the smoothed version $\tilde{\zb}^{\mu}_{s} = \sigmab_s^+\left( \operatorname{smooth}^\mu_s \EE_{\wb \sim P_s} \left[\sigmab_s\left( (x_{sj}^\top \wb)_{j \in S_s} \right) \right] \right)$ as an intermediate-term, where the smooth function is defined by $ \operatorname{smooth}^\mu_s(\qb) = (1-\mu)\qb + \mu \mathbf{1}/(K_s + 1)$, where $\mathbf{1} \in \RR^{K_s}$ is an all one vector.

    Note that $\tilde{\zb}_{s}^\mu =\sigmab_s^+ ( \operatorname{smooth}^\mu_s( \sigmab_s (\tilde{\zb}_{s})))$ by the definition of  the pseudo inverse function $\sigmab_s^+$ such that $\sigmab_s^+ ( \sigmab_s (\qb)  ) = \qb$ for any $\qb \in \{ \pb \in [0,1]^{K_s} \mid \| \pb\|_1 < 1\}$. 
    Then, by Lemma~\ref{lemma:zhang_lemma17}, we have
    \begin{align*}
        \sum_{s=1}^t \ell (\tilde{\zb}_{s}^\mu, \yb_s)
        - \sum_{s=1}^t \ell (\tilde{\zb}_{s}, \yb_s)
        \leq 2 \mu t,
        \quad
        \text{and}
        \quad
        \|\tilde{\zb}_{s}^\mu \|_{\infty} \leq \log ( 1 + (K + 1)/ \mu).
        \numberthis \label{eq:I2_A_first_bound}
    \end{align*}
    Hence, to prove the lemma, we need only to bound the gap between the loss of $\wb^\star$ and $\tilde{\zb}_{s}^\mu$.
    To enhance clarity in our presentation, let $\ell(\zb_{s}^\star, \yb_s )  =\ell_{s} (\wb^\star) $, where $\zb_{s}^\star = \left(x_{sj}^\top \wb^\star \right)_{j \in S_s} \in \RR^{K_s}$.
    Then, we have
    \begin{align*}
        \sum_{s=1}^t \ell_{s} (\wb^\star) - \sum_{s=1}^t \ell(\tilde{\zb}_{s}^\mu, \yb_s) 
        &= \sum_{s=1}^t \ell(\zb_{s}^\star, \yb_s ) - \sum_{s=1}^t \ell(\tilde{\zb}_{s}^\mu, \yb_s)  
        \\
        &\leq  \sum_{s=1}^t  \langle \nabla_z\ell(\zb_{s}^\star, \yb_s ), \zb_{s}^\star - \tilde{\zb}_{s}^\mu \rangle
        - \sum_{s=1}^t \frac{1}{c_{\mu}} \| \zb_{s}^\star- \tilde{\zb}_{s}^\mu\|_{\nabla_z^2 \ell(\zb_{s}^\star, \yb_s ) }^2
        \\
        &= \sum_{s=1}^t  \langle \sigmab_s(\zb_s^\star) - \yb_s, \zb_{s}^\star - \tilde{\zb}_{s}^\mu \rangle
        - \sum_{s=1}^t \frac{1}{c_{\mu}} \| \zb_{s}^\star- \tilde{\zb}_{s}^\mu\|_{\nabla \sigmab_s (\zb_{s}^\star) }^2
        ,
        \numberthis  \label{eq:I2_A_intermediate_term}
    \end{align*}
    where $c_{\mu} = \log (K+1) + 2\log ( 1 + (K + 1)/ \mu) + 2$, the inequality holds by Lemma~\ref{lemma:zhang_lemma1}, and the last equality holds by a direct calculation of the first order and Hessian of the logistic loss as follows:
    \begin{align*}
        \nabla_z \ell(\zb_s, \yb_s) = \sigmab_s(\zb_s) - \yb_s, 
        \quad
        \nabla_z^2 \ell(\zb_s, \yb_s) = \operatorname{diag}(\sigmab_s(\zb_s)) - \sigmab_s(\zb_s)\sigmab_s(\zb_s)^\top.
    \end{align*}

    We first bound the first term of the right-hand side.
    Define $\mathbf{d}_{s} = (\zb_{s}^\star - \tilde{\zb}_{s}^\mu )/(c_\mu + 1)$.
    Let $\mathbf{d}'_s $ be $\mathbf{d}_{s}$ extended with zero padding.
    Specifically, we define $\mathbf{d}'_s = [\mathbf{d}_{s}^\top, 0, \dots, 0]^\top\in \RR^K$, 
    where the zeros are appended to increase the dimension of  $\mathbf{d}_{s}$ to $K$.
    Similarly, we also extend $ \sigmab_s (\zb_{s}^\star) - \yb_s$ with zero padding and define $\varepsilonb_s = [ (\sigmab_s (\zb_{s}^\star) - \yb_s)^\top, 0, \dots, 0 ]^\top \in \RR^K$.
 
    Then, one can easily verify that $\|\mathbf{d}'_{s} \|_{\infty} \leq 1$ since $\| \zb_{s}^\star \|_{\infty} \leq \max_{i \in S_s} \| x_{si} \|_2 \| \wb^\star \|_2 \leq 1$ and $\| \tilde{\zb}_{s}^\mu \|_{\infty} \leq  \log ( 1 + (K + 1)/ \mu)$.
    On the other hand, $\mathbf{d}_{s}'$ is $\mathcal{F}_{s}$-measurable since $\zb_{s}^\star $ and $ \tilde{\zb}_{s}^\mu $ are independent of $\yb_s$.
    Moreover, we have $ \|\mathbf{d}'_{s}\|^2_{\EE[\varepsilonb_s  \varepsilonb_s^\top \mid \mathcal{F}_{s}]} = \|\mathbf{d}_{s}\|^2_{\EE\left[ (\sigmab_s(\zb_s^\star) - \yb_s) (\sigmab_s(\zb_s^\star) - \yb_s)^\top  \mid \mathcal{F}_{s} \right]} =  \|\mathbf{d}_{s}\|^2_{\nabla \sigmab_s (\zb_{s}^\star)}$  and $\| \sigmab_s (\zb_{s}^\star) - \yb_s\|_1 \leq 2$.
    Thus, by Lemma~\ref{lemma:zhang_lemma15}, with probability at least $1-\delta$, for any $t \geq 1$, we have
    \begin{align*}
         &\sum_{s=1}^t  \langle  \sigmab_s (\zb_{s}^\star) - \yb_s, \zb_{s}^\star - \tilde{\zb}_{s}^\mu \rangle
         = (c_\mu + 1) \sum_{s=1}^t \langle \sigmab_s (\zb_{s}^\star) - \yb_s, \mathbf{d}_{s} \rangle 
         \\
         &= (c_\mu + 1) \sum_{s=1}^t \langle \varepsilonb_s, \mathbf{d}'_{s} \rangle 
         \\
         &\leq (c_\mu + 1) \sqrt{\tilde{\lambda} + \sum_{s=1}^t \| \mathbf{d}_{s} \|_{\nabla \sigmab_s (\zb_{s}^\star)}^2 }
         \left(
            \frac{\sqrt{\tilde{\lambda}}}{4} + \frac{4}{\sqrt{\tilde{\lambda}}}\log 
         \cdot \left( \frac{2\sqrt{ 1  +  \frac{1}{\tilde{\lambda}} \sum_{s=1}^t \| \mathbf{d}_{s} \|_{\nabla \sigmab_s (\zb_{s}^\star)}^2}}{\delta} \right)  
         \right)
         \\
         &\leq  (c_\mu + 1) \sqrt{\tilde{\lambda} + \sum_{s=1}^t \| \mathbf{d}_{s} \|_{\nabla \sigmab_s (\zb_{s}^\star)}^2 }
         \cdot 
         \left(
            \frac{\sqrt{\tilde{\lambda}}}{4} 
            + \frac{4}{\sqrt{\tilde{\lambda}}} \log \left( \frac{2\sqrt{1 + 2t}}{\delta} \right)   
         \right),
         \numberthis  \label{eq:I2_A_self_normal}
    \end{align*}
    where 
    in
    the second inequality,
    we set  $\tilde{\lambda} > 1$, and the last inequality holds 
    because $  \| \mathbf{d}_{s} \|_{\nabla \sigmab_s (\zb_{s}^\star)}^2 =\mathbf{d}_{s}^\top \nabla \sigmab_s (\zb_{s}^\star) \mathbf{d}_{s} \leq 2$.
    Then, combining Equation~\eqref{eq:I2_A_intermediate_term} and~\eqref{eq:I2_A_self_normal}, we obtain
    \begin{align*}
        \sum_{s=1}^t &\ell_{s} (\wb^\star) - \sum_{s=1}^t \ell(\tilde{\zb}_{s}^\mu, \yb_s)  
        \\
        &\leq  (c_\mu + 1) \sqrt{\tilde{\lambda}  + \sum_{s=1}^t \| \mathbf{d}_{s} \|_{\nabla \sigmab_s (\zb_{s}^\star)}^2 }
        \cdot 
         \left(
            \frac{\sqrt{\tilde{\lambda}}}{4} 
            + \frac{4}{\sqrt{\tilde{\lambda}}} \log \left( \frac{2\sqrt{1 + 2t}}{\delta} \right)   
         \right)
         - \sum_{s=1}^t \frac{1}{c_{\mu}} \| \zb_{s}^\star- \tilde{\zb}_{s}^\mu\|_{\nabla \sigmab_s (\zb_{s}^\star) }^2
         \\
         &\leq  (c_\mu + 1) \sqrt{\tilde{\lambda}  + \sum_{s=1}^t \| \mathbf{d}_{s} \|_{\nabla \sigmab_s (\zb_{s}^\star)}^2 }
         \cdot 
         \left(
            \frac{\sqrt{\tilde{\lambda}}}{4} 
            + \frac{4}{\sqrt{\tilde{\lambda}}} \log \left( \frac{2\sqrt{1 + 2t}}{\delta} \right)   
         \right)
         - (c_\mu + 1)\sum_{s=1}^t  \| \mathbf{d}_{s} \|_{\nabla \sigmab_s (\zb_{s}^\star) }^2
         \\
         &\leq  (c_\mu + 1) \left(  \tilde{\lambda}  + \sum_{s=1}^t \| \mathbf{d}_{s} \|_{\nabla \sigmab_s (\zb_{s}^\star)}^2 \right)
         +  (c_\mu + 1) \left(
            \frac{\sqrt{\tilde{\lambda}}}{4} 
            + \frac{4}{\sqrt{\tilde{\lambda}}} 
            \log \left( \frac{2\sqrt{1 + 2t}}{\delta} \right)   
         \right)^2
         \\
         &- (c_\mu + 1)\sum_{s=1}^t  \| \mathbf{d}_{s} \|_{\nabla \sigmab_s (\zb_{s}^\star) }^2
         \\
         &= (c_\mu + 1) \left( \frac{17}{16} \tilde{\lambda}
         + 2  \log \left( \frac{2\sqrt{1 + 2t}}{\delta} \right)  
         + \frac{16}{\tilde{\lambda}} 
         \left(\log \left( \frac{2\sqrt{1 + 2t}}{\delta} \right) \right)^2
         \right),
         \numberthis \label{eq:I2_A_second_bound}
    \end{align*}
    where the third inequality holds due to the AM-GM inequality.
    Finally, combining Equation~\eqref{eq:I2_A_first_bound} and~\eqref{eq:I2_A_second_bound}, by setting $\mu = 1/t$ and $\tilde{\lambda} = \frac{16}{\sqrt{17}} \log \left( \frac{2\sqrt{1 + 2t}}{\delta} \right)$, we have
    \begin{align*}
        \sum_{s=1}^t \left(\ell_{s} (\wb^\star) - \ell(\tilde{\zb}_{s}, \yb_s)  \right)
        &\leq (c_\mu + 1) 
         (2\sqrt{17} +2 )
         \log \left( \frac{2\sqrt{1 + 2t}}{\delta} \right)   
        + 2\mu t
        \\
        &\leq 11 \cdot
        \left( 3\log ( 1 + (K + 1)t) + 3 \right) 
          \log \left( \frac{2\sqrt{1 + 2t}}{\delta} \right)   
        + 2
    \end{align*}
    where the last inequality holds by the definition of $c_\mu = \log (K+1) + 2\log ( 1 + (K + 1)/ \mu) + 2$.
    This concludes the proof.     
\end{proof}

\subsubsection{Proof of Lemma~\ref{lemma:bound_(b)}  }
\label{app_subsubsec:proof_lemma:lemma:bound_(b)}
\begin{proof}[Proof of Lemma~\ref{lemma:bound_(b)}]
    The proof with an observation from Proposition 2 in~\citet{foster2018logistic},
    which notes that $\tilde{\zb}_{s}$ is an aggregation forecaster for the logistic function. Hence, it satisfies 
    \begin{align}
        \ell(\tilde{\zb}_{s}, \yb_s)
        \leq -\log \left( \EE_{\wb \sim  P_s}\left[ e^{-\ell_{s}(\wb)} \right] \right)
        = -\log \left( \frac{1}{Z_{s}} \int_{\RR^d} e^{-L_{s}(\wb)} \dd \wb \right),
        \label{eq:ztilde_loss_upper_ln}
    \end{align}
    where $L_{s}(\wb) := \ell_{s}(\wb) + \frac{1}{2c} \| \wb - \wb_{s} \|_{H_{s}}^2$ and $Z_{s} := \sqrt{(2\pi)^d c |H_{s}^{-1}|}$.

    Then, by the quadratic approximation, we get
    \begin{align}
        \tilde{L}_{s}(\wb) = L_{s}(\wb_{s+1}) 
        + \langle \nabla L_{s}(\wb_{s+1}) , \wb - \wb_{s+1} \rangle
        + \frac{1}{2c}\| \wb - \wb_{s+1}\|_{H_{s}}^2.
        \label{eq:L_s_tilde_quadratic}
    \end{align}
    Applying Lemma~\ref{lemma:zhang_lemma18} and considering the fact that
     $\ell_{s}$ is $3\sqrt{2}$-self-concordant-like function by Proposition~\ref{prop:self_concordant},  we have
    \begin{align}
          L_{s}(\wb) \leq \tilde{L}_{s}(\wb) + e^{18\| \wb - \wb_{s+1}\|_2^2 } \| \wb - \wb_{s+1} \|_{\nabla^2 \ell_{s}(\wb_{s+1})}^2.
          \label{eq:L_s_upperbound}
    \end{align}
    We define the function $\tilde{f}_{s+1}: \mathcal{W} \rightarrow \RR$ as
    \begin{align*}
        \tilde{f}_{s+1} (\wb)
        = \exp\left( -\frac{1}{2c} \| \wb - \wb_{s+1} \|_{H_{s}}^2 -  e^{18\| \wb - \wb_{s+1}\|_2^2 } \| \wb - \wb_{s+1} \|_{\nabla^2 \ell_{s}(\wb_{s+1})}^2   \right).
    \end{align*}
    Then, we can establish a lower bound for the expectation in Equation~\eqref{eq:ztilde_loss_upper_ln} as follows:
    \begin{align*}
        \EE_{\wb \sim  P_s}\left[ e^{-\ell_{s}(\wb)} \right]
        &= \frac{1}{Z_{s}} \int_{\RR^d} \exp(-L_{s}(\wb)) \dd \wb
        \\
        &\geq \frac{1}{Z_{s}}  \int_{\RR^d} \exp(-\tilde{L}_{s}(\wb) - e^{18\| \wb - \wb_{s+1}\|_2^2 } \| \wb - \wb_{s+1} \|_{\nabla \ell_{s}(\wb_{s+1})}^2 ) \dd \wb
        \\
        &= \frac{\exp(-L_{s}(\wb_{s+1}) )}{Z_{s}}   \int_{\RR^d} \tilde{f}_{s+1}(\wb) \cdot \exp(-\langle \nabla L_{s} (\wb_{s+1}), \wb - \wb_{s+1} \rangle) \dd \wb,
        \numberthis \label{eq:exp_loss_lowerbound}
    \end{align*}
    where the first inequality holds by Equation~\eqref{eq:L_s_upperbound} and the last equality holds by Equation~\eqref{eq:L_s_tilde_quadratic}.
    We define $\tilde{Z}_{s+1} = \int_{\RR^d} \tilde{f}_{s+1} (\wb)  \dd \wb \leq + \infty$.
    Moreover, we denote the distribution whose density function is $\tilde{f}_{s+1} (\wb)/\tilde{Z}_{s+1} $ as $\tilde{P}_{s+1}$.
    Then, we can rewrite Equation~\eqref{eq:exp_loss_lowerbound} as follows:
    \begin{align*}
        \EE_{\wb \sim  P_s}\left[ e^{-\ell_{s}(\wb)} \right]
        &\geq \frac{\exp(-L_{s}(\wb_{s+1}) ) \tilde{Z}_{s+1}}{Z_{s}} \EE_{\wb \sim \tilde{P}_{s+1}} \left[ \exp(-\langle \nabla L_{s} (\wb_{s+1}), \wb - \wb_{s+1} \rangle) \right]
        \\
        &\geq \frac{\exp(-L_{s}(\wb_{s+1}) ) \tilde{Z}_{s+1}}{Z_{s}}
        \exp \left( - \EE_{\wb \sim \tilde{P}_{s+1}} \left[\langle \nabla L_{s} (\wb_{s+1}), \wb - \wb_{s+1} \rangle \right]   \right)
        \\
        &= \frac{\exp(-L_{s}(\wb_{s+1}) ) \tilde{Z}_{s+1}}{Z_{s}},
        \numberthis \label{eq:exp_loss_lowerbound2}
    \end{align*}
    where the second inequality follows from Jensen's inequality, and the equality holds because $\tilde{P}_{s+1}$ is symmetric around $\wb_{s+1}$, thus $\EE_{\wb \sim \tilde{P}_{s+1}} \left[\langle \nabla L_{s} (\wb_{s+1}), \wb - \wb_{s+1} \rangle \right] = 0$.

    By plugging Equation~\eqref{eq:exp_loss_lowerbound2} into Equation~\eqref{eq:ztilde_loss_upper_ln}, we have
    \begin{align}
        \ell(\tilde{\zb}_{s}, \yb_s)
        \leq L_{s}(\wb_{s+1}) + \log Z_{s} - \log \tilde{Z}_{s+1}.
        \label{eq:ztilde_loss_upper_ln_intermediate}
    \end{align}
    In the above, we can bound the last term, $-\log \tilde{Z}_{s+1}$, by
    \begin{align*}
        - \log \tilde{Z}_{s+1}
        &= -\log \left( \int_{\RR^d}  \exp\left( -\frac{1}{2c} \| \wb - \wb_{s+1} \|_{H_{s}}^2 -  e^{18\| \wb - \wb_{s+1}\|_2^2 } \| \wb - \wb_{s+1} \|_{\nabla^2 \ell_{s}(\wb_{s+1})}^2   \right) \dd \wb \right)
        \\
        &= -\log \left( \widehat{Z}_{s+1} \cdot \EE_{\wb \sim \widehat{P}_{s+1} } \left[ \exp \left(  -  e^{18\| \wb - \wb_{s+1}\|_2^2 } \| \wb - \wb_{s+1} \|_{\nabla^2 \ell_{s}(\wb_{s+1})}^2  \right)  \right]  \right)
        \\
        &\leq -\log  \widehat{Z}_{s+1} + \EE_{\wb \sim \widehat{P}_{s+1} }  \left[  e^{18\| \wb - \wb_{s+1}\|_2^2 } \| \wb - \wb_{s+1} \|_{\nabla^2 \ell_{s}(\wb_{s+1})}^2  \right]
        \\
        &=  -\log  Z_{s} + \EE_{\wb \sim \widehat{P}_{s+1} }  \left[  e^{18\| \wb - \wb_{s+1}\|_2^2 } \| \wb - \wb_{s+1} \|_{\nabla^2 \ell_{s}(\wb_{s+1})}^2  \right],
        \numberthis \label{eq:ztilde_loss_upper_ln_intermediate2}
    \end{align*}
    where $ \widehat{P}_{s+1}  = \mathcal{N}(\wb_{s+1}, c H_{s}^{-1})$ and $\widehat{Z}_{s+1} = \int_{\RR^d} \exp\left(  -\frac{1}{2c} \| \wb - \wb_{s+1} \|_{H_{s}}^2 \right) \dd \wb$.
    In Equation~\eqref{eq:ztilde_loss_upper_ln_intermediate2}, the inequality holds due to Jensen’s inequality, and the last inequality is by the fact that $\widehat{Z}_{s+1} =  \int_{\RR^d} \exp\left(  -\frac{1}{2c} \| \wb - \wb_{s+1} \|_{H_{s}}^2 \right) \dd \wb =  \sqrt{(2\pi)^d c |H_{s}^{-1}|} = Z_{s}$.

    By applying the Cauchy-Schwarz inequality, we can further bound the second term on the right-hand side of Equation~\eqref{eq:ztilde_loss_upper_ln_intermediate2} by
    \begin{align*}
        \EE_{\wb \sim \widehat{P}_{s+1} }  &\left[  e^{18\| \wb - \wb_{s+1}\|_2^2 } \| \wb - \wb_{s+1} \|_{\nabla^2 \ell_{s}(\wb_{s+1})}^2  \right]
        \\
        &\leq \underbrace{\sqrt{ \EE_{\wb \sim \widehat{P}_{s+1} }  \left[  e^{36\| \wb - \wb_{s+1}\|_2^2 }\right] }}_{\texttt{(a)-1}}
        \underbrace{\sqrt{  \EE_{\wb \sim \widehat{P}_{s+1} }  \left[  \| \wb - \wb_{s+1} \|_{\nabla^2 \ell_{s}(\wb_{s+1})}^4  \right] }}_{\texttt{(a)-2}}.
        \numberthis \label{eq:expectation_decomposition}
    \end{align*}
    Note that, since $\widehat{P}_{s+1} = \mathcal{N}(\wb_{s+1}, c H_{s}^{-1})$, there exist orthogonal bases $\eb_1, \dots, \eb_d \in \RR^d$ such that $\wb - \wb_{s+1}$ follows the same distribution as 
    \begin{align}
        \sum_{j=1}^d \sqrt{c \lambda_j \left( H_{s}^{-1} \right)} X_j \eb_j,
        \quad
        \text{where }
        X_j \stackrel{i.i.d.}{\sim} \mathcal{N}(0,1), 
        \forall j \in [d],
        \label{eq:basis_expression}
    \end{align}
    and $\lambda_j \left( H_{s}^{-1} \right)$  denotes the $j$-th largest eigenvalue of $H_{s}^{-1}$.
    Then, we can bound the term \texttt{(a)-1} in Equation~\eqref{eq:expectation_decomposition} as follows:
    \begin{align*}
        \sqrt{ \EE_{\wb \sim \widehat{P}_{s+1} }  \left[  e^{36\| \wb - \wb_{s+1}\|_2^2 }\right] }
        &= \sqrt{\EE_{X_j} \left[  \prod_{j=1}^d e^{36c \lambda_j \left( H_{s}^{-1} \right) X_j^2}  \right] }
        \leq \sqrt{ \prod_{j=1}^d\EE_{X_j} \left[   e^{ \frac{36c}{\lambda} X_j^2}  \right] }
        \\
        &= \left( \EE_{X \sim \chi^2} \left[ e^{\frac{36c}{\lambda} X} \right] \right)^{\frac{d}{2}}
        \leq  \EE_{X \sim \chi^2} \left[ e^{ \frac{18cd}{\lambda} X} \right],
    \end{align*}
    where the first inequality holds since  $\lambda_j \left( H_{s}^{-1} \right) \leq \frac{1}{\lambda}$. 
    In the second equality, $ \chi^2$ denotes the chi-square distribution, and the last inequality is due to Jensen’s inequality.
    By setting $\lambda \geq 72c d$, we get
    \begin{align}
        \sqrt{ \EE_{\wb \sim \widehat{P}_{s+1} }  \left[  e^{36\| \wb - \wb_{s+1}\|_2^2 }\right] }
        \leq \EE_{X \sim \chi^2} \left[ e^{\frac{X}{4}} \right] 
        \leq \sqrt{2},
        \label{eq:term (a)-1}
    \end{align}    
    where the last inequality holds due to the fact that the moment-generating function for $\chi^2$-distribution is bounded by $\EE_{X \sim \chi^2} [e^{tX}] \leq 1/\sqrt{1-2t} $ for all $t \leq 1/2$.

    Now, we bound the term \texttt{(a)-2} in Equation~\eqref{eq:expectation_decomposition}.
    \begin{align*}
        \sqrt{  \EE_{\wb \sim \widehat{P}_{s+1} }  \left[  \| \wb - \wb_{s+1} \|_{\nabla^2 \ell_{s}(\wb_{s+1})}^4  \right] }
        &= \sqrt{  \EE_{\wb \sim  \mathcal{N}(0, c H_{s}^{-1}) }  \left[  \| \wb  \|_{\nabla^2 \ell_{s}(\wb_{s+1})}^4  \right] }
        \\
        &= \sqrt{  \EE_{\wb \sim  \mathcal{N}(0, c \bar{H}_{s}^{-1}) }  \left[  \| \wb  \|_2^4  \right] },
    \end{align*}
    where $\bar{H}_{s} = ( \nabla^2 \ell_{s}(\wb_{s+1}) )^{-1/2} H_{s} (\nabla^2 \ell_{s}(\wb_{s+1}))^{-1/2}$.
    Let $\bar{\lambda}_j = \lambda_j \left( c \bar{H}_{s}^{-1} \right)$ be the $j$-th largest eigenvalue of the matrix $c \bar{H}_{s}^{-1}$.
    Then, conducting an analysis similar to that in Equation~\eqref{eq:basis_expression} yields that
    \begin{align*}
        \sqrt{  \EE_{\wb \sim  \mathcal{N}(0, c \bar{H}_{s}^{-1}) }  \left[  \| \wb  \|_2^4  \right] }
        &= \sqrt{\EE_{X_j \sim \mathcal{N}(0,1)} \left[ \left\| \sum_{j=1}^d \sqrt{\bar{\lambda}_j} X_j \eb_j  \right\|_2^4 \right] }
        \\
        &= \sqrt{\EE_{X_j \sim \mathcal{N}(0,1)} \left[ \left( \sum_{j=1}^d \bar{\lambda}_j X_j^2\right)^2 \right] }
        \\
        &= \sqrt{ \sum_{j=1}^d \sum_{j'=1}^d \bar{\lambda}_j \bar{\lambda}_{j'} \EE_{X_j, X_{j'} \sim \mathcal{N}(0,1)} \left[X_j^2 X_{j'}^2\right] }
        \\
        &\leq \sqrt{3 \sum_{j=1}^d \sum_{j'=1}^d \bar{\lambda}_j \bar{\lambda}_{j'}}
        = \sqrt{3}c \operatorname{Tr} \left(\bar{H}_{s}^{-1}\right),
    \end{align*}
    where the inequality holds due to $\EE_{X_j, X_{j'} \sim \mathcal{N}(0,1)} [X_j^2 X_{j'}^2] \leq 3$ for all $j, j' \in [d]$, and the last equality holds because $\sum_{j=1}^d \bar{\lambda}_j = \operatorname{Tr}\left(c \bar{H}_{s}^{-1} \right)$.
    Here, $\operatorname{Tr}(A)$ denotes the trace of the matrix $A$.

    We define the matrix $M_{s+1} := \lambda \Ib_d/2 + \sum_{\tau=1}^s \nabla^2 \ell_{\tau} (\wb_{\tau+1})$.
    Under the condition $\lambda \geq 2$, for any $s \in [T]$ and $\wb \in \mathcal{W}$, we have $\nabla^2 \ell_{s} (\wb) \preceq \Ib_d \leq \frac{\lambda}{2} \Ib_d $.
    Thus, we have $H_{s} \succeq M_{s+1}$.
    Then, we can bound the trace as follows:
    \begin{align*}
         \operatorname{Tr} \left(\bar{H}_{s}^{-1}\right)
         &= \operatorname{Tr}\left( H_{s}^{-1} \nabla^2 \ell_s(\wb_{s+1}) \right)
         \leq \operatorname{Tr} \left( M_{s+1}^{-1}  \nabla^2 \ell_s(\wb_{s+1}) \right) 
         \\
         &= \operatorname{Tr}  \left( M_{s+1}^{-1} (M_{s+1}- M_{s}) \right)
         \leq \log \frac{\operatorname{det}(M_{s+1})}{\operatorname{det}(M_{s})},
    \end{align*}
    where the last inequality holds by Lemma 4.5 of~\citet{hazan2016introduction}.
    Therefore we can bound the term \texttt{(a)-2} as
    \begin{align}
        \sqrt{  \EE_{\wb \sim \widehat{P}_{s+1} }  \left[  \| \wb - \wb_{s+1} \|_{\nabla^2 \ell_{s}(\wb_{s+1})}^4  \right] } 
        \leq \sqrt{3}c \log \frac{\operatorname{det}(M_{s+1})}{\operatorname{det}(M_{s})}.
        \label{eq:term (a)-2}
    \end{align}
    By plugging Equation~\eqref{eq:term (a)-1} and~\eqref{eq:term (a)-2} into Equation~\eqref{eq:expectation_decomposition}, we have
    \begin{align}
        \EE_{\wb \sim \widehat{P}_{s+1} }  \left[  e^{18\| \wb - \wb_{s+1}\|_2^2 } \| \wb - \wb_{s+1} \|_{\nabla^2 \ell_{s}(\wb_{s+1})}^2  \right]
        \leq \sqrt{6}c \log \frac{\operatorname{det}(M_{s+1})}{\operatorname{det}(M_{s})}.
        \label{eq:term (a)-1 and (a)-2}
    \end{align}
    Combining Equation~\eqref{eq:ztilde_loss_upper_ln_intermediate},~\eqref{eq:ztilde_loss_upper_ln_intermediate2}, and~\eqref{eq:term (a)-1 and (a)-2}, and taking summation over $s$, we derive that
    \begin{align*}
        \sum_{s=1}^t \ell(\tilde{\zb}_{s}, \yb_s)
        &\leq \sum_{s=1}^t L_{s}(\wb_{s+1}) +  \sqrt{6}c \sum_{s=1}^t \log \frac{\operatorname{det}(M_{s+1})}{\operatorname{det}(M_{s})}
        \\
        &=\sum_{s=1}^t  \ell_{s}(\wb_{s+1}) + \frac{1}{2c} \sum_{s=1}^t\| \wb_{s} - \wb_{s+1} \|_{H_{s}}^2  +  \sqrt{6}c \sum_{s=1}^t \log \frac{\operatorname{det}(M_{s+1})}{\operatorname{det}(M_{s})}
        \\
        &=\sum_{s=1}^t  \ell_{s}(\wb_{s+1}) + \frac{1}{2c} \sum_{s=1}^t\| \wb_{s} - \wb_{s+1} \|_{H_{s}}^2  
        +  \sqrt{6}c \log \left( \frac{\operatorname{det}(M_{s+1})}{\operatorname{det}\left(\frac{\lambda}{2} \Ib_d\right)}   \right) 
        \\
        &\leq \sum_{s=1}^t  \ell_{s}(\wb_{s+1}) + \frac{1}{2c} \sum_{s=1}^t \| \wb_{s} - \wb_{s+1} \|_{H_{s}}^2  
        + \sqrt{6}c \cdot d \log \left( 1 + \frac{t+1}{2\lambda}\right),
    \end{align*}
    By rearranging the terms, we conclude the proof.
\end{proof}

\subsection{Technical Lemmas for Lemma~\ref{lemma:online_confidence_set}} 
\label{app_sec:technical_leamms}

\begin{lemma}[Proposition 4.1 of~\citealt{campolongo2020temporal}]
\label{lemma:loss_firstorder_decomposition}
    Let the $\wb_{t+1}$ be the solution of the update rule
    \[
    \wb_{t+1} = \arg\min_{\wb \in \mathcal{V}} \eta_t \ell_t(\wb) + D_{\psi}(\wb, \wb_t),
    \]
    where $\mathcal{V} \subseteq \mathcal{W} \subseteq \mathbb{R}^d$ is a non-empty convex set and $D_{\psi}(\wb_1, \wb_2) = \psi(\wb_1) - \psi(\wb_2) - \langle \nabla\psi(\wb_2), \wb_1 - \wb_2 \rangle$ is the Bregman Divergence w.r.t. a strictly convex and continuously differentiable function $\psi : \mathcal{W} \rightarrow \mathbb{R}$. Further supposing $\psi(\wb)$ is $1$-strongly convex w.r.t. a certain norm $\|\cdot\|$ in $\mathcal{W}$, then there exists a $\gb'_t \in \partial\ell_t(\wb_{t+1})$ such that
    \begin{align*}
        \langle \eta_t \gb'_t, \wb_{t+1} - \ub \rangle 
        \leq \langle \nabla \psi(\wb_t) - \nabla \psi(\wb_{t+1}), \wb_{t+1} - \ub \rangle     
    \end{align*}
    for any $\ub \in \mathcal{W}$.
\end{lemma}
\begin{lemma}[Lemma 15 of \citealt{zhang2024online}] \label{lemma:zhang_lemma15}
Let $\{\mathcal{F}_t\}_{t=1}^{\infty}$ be a filtration. 
Let $\{\zb_t\}_{t=1}^{\infty}$ be a stochastic process in $\mathcal{B}_2(K) = \{\zb \in \mathbb{R}^K \mid \|\zb\|_{\infty} \leq 1\}$ such that $\zb_t$ is $\mathcal{F}_t$ measurable. 
Let $\{\boldsymbol{\varepsilonb}_t\}_{t=1}^{\infty}$ be a martingale difference sequence such that $\varepsilonb_t \in \mathbb{R}^K$ is $\mathcal{F}_{t+1}$ measurable. 
Furthermore, assume that, conditional on $\mathcal{F}_t$, we have $\|\varepsilonb_{t}\|_1 \leq 2$ almost surely. 
Let $\Sigma_t = \EE[\varepsilonb_t \varepsilonb_t^\top | \mathcal{F}_t]$. 
and $\lambda > 0$.
Then, for any $t \geq 1$ define
    \begin{align*}
    U_t = \sum_{s=1}^{t-1} \langle \varepsilonb_s, \zb_s \rangle \quad 
    \text{and} 
    \quad H_t = \lambda + \sum_{s=1}^{t-1} \|\zb_s \|_{\Sigma_s}^2,        
    \end{align*}
    Then, for any $\delta \in (0,1]$, we have
    \begin{align*}
    \Pr\left[ \exists t \geq 1, U_t \geq \sqrt{H_t} 
    \left( 
        \frac{\sqrt{\lambda}}{4} + \frac{4}{\sqrt{\lambda}} \log \left( \sqrt{\frac{H_t}{\lambda}} \right) + \frac{4}{\sqrt{\lambda}} \log\left(\frac{2}{\delta}\right) 
    \right)
    \right] 
    \leq \delta.        
    \end{align*}
\end{lemma}
\begin{lemma}[Lemma 1 of \citealt{zhang2024online}] \label{lemma:zhang_lemma1}
Let $C>0$, $\mathbf{a} \in [-C,C]^K$, $\yb \in \RR^{K+1}$ be a one-hot vector and $\mathbf{b} \in \RR^K$.
Then, we have
\begin{align*}
    \ell(\mathbf{a}, \yb) \geq 
    \ell(\mathbf{b}, \yb) + \nabla \ell(\mathbf{b}, \yb)^\top (\mathbf{a} - \mathbf{b})
    + \frac{1}{\log(K+1) + 2(C+1)} (\mathbf{a} - \mathbf{b})^\top \nabla^2 \ell(\mathbf{b},\yb)  (\mathbf{a} - \mathbf{b}).
\end{align*}
\end{lemma}
\begin{lemma}[Lemma 17 of \citealt{zhang2024online}] \label{lemma:zhang_lemma17}
    Let $\zb \in \mathbb{R}^K$ be a $K$-dimensional vector.
    Let $\ell(\zb, \yb) = \sum_{k=0}^{K} \mathbf{1}\{ y_{i}  = 1\} \cdot \log\left(\frac{1}{[\sigmab(\zb)]_k}\right)$, where $\yb = [y_{0}, \dots, y_{K}]^\top \in \RR^{K+1}$,
    and
    the softmax function $\sigmab(\zb): \RR^{K} \rightarrow \RR^{K}$ is defined as  $[\sigmab(\zb)]_{i} = \frac{\exp([\zb]_{i})}{v_0 +  \sum_{k=1}^K   \exp([\zb]_{k})}$ for all $i \in [K]$,
    and $[\sigmab(\zb)]_{0} = \frac{ v_0 }{v_0 +  \sum_{k=1}^K   \exp([\zb]_{k})}$.
    Define $\zb^{\mu} := \sigmab^+\left(\operatorname{smooth}_\mu(\sigmab(\zb))\right)$, where $\operatorname{smooth}_\mu(\qb) = (1 - \mu)\qb + \mu \mathbf{1}/(K+1)$. Then, for $\mu \in [0, 1/2]$, we have
    \begin{align*}
        \ell(\zb^{\mu}, \yb) - \ell(\zb, \yb) \leq 2\mu
    \end{align*}
    We also have $\|\zb^{\mu}\|_{\infty} \leq \log( 1 + (K + 1)/ \mu)$.
\end{lemma}
\begin{lemma}[Lemma 18 of \citealt{zhang2024online}] \label{lemma:zhang_lemma18}
    Let $L_t(\wb) = \ell_t(\wb) + \frac{1}{2c} \| \wb - \wb_t \|_{H_t}^2$. 
    Assume that $\ell_t$ is a $M$-self-concordant-like function.
    Then, for any $\wb,  \wb_t \in \mathcal{W}$, the quadratic approximation $\tilde{L}_t(\wb) = L_t(\wb_{t+1}) 
            + \langle \nabla^2L_t(\wb_{t+1}) , \wb - \wb_{t+1} \rangle
            + \frac{1}{2c}\| \wb - \wb_{t+1}\|_{H_t}^2$ satisfies
    \begin{align*}
         L_t(\wb) \leq \tilde{L}_t(\wb) + e^{M^2\| \wb - \wb_{t+1}\|_2^2 } \| \wb - \wb_{t+1} \|_{\nabla \ell_t(\wb_{t+1})}^2.
    \end{align*}
\end{lemma}
\begin{lemma} \label{lemma:zhang_lemma20_improved}
    Let $\wb_{t+1}  = \argmin_{\wb \in \mathcal{W}} 
    \langle 
        \nabla \ell_t (\wb_t), \wb
    \rangle
    + \frac{1}{2 \eta} 
    \|\wb - \wb_t \|_{\tilde{H}_t}^2 
    $.
    Then, we have
    \begin{align*}
        \| \wb_{t+1} - \wb_t \|_{H_t}
        \leq 2\eta \| \nabla \ell_t (\wb_t) \|_{H_t^{-1}}
        \leq \frac{2\eta}{\lambda} \| \nabla \ell_t (\wb_t) \|_2
        \leq \frac{4 \eta}{\lambda}.
    \end{align*}
\end{lemma}
\begin{proof} [Proof of Lemma~\ref{lemma:zhang_lemma20_improved}]
    We make a slight improvement to Lemma 20 of~\citet{zhang2024online}. 
    \begin{lemma} [Lemma 20 of \citealt{zhang2024online}]
         Under the same conditions as Lemma~\ref{lemma:zhang_lemma20_improved}, the following holds:
        \begin{align*}
            \| \wb_{t+1} - \wb_t \|_{H_t}
            \leq 2\eta \| \nabla \ell_t (\wb_t) \|_{H_t^{-1}}
            \leq \frac{2\eta}{\lambda} \| \nabla \ell_t (\wb_t) \|_2
            \leq \frac{2\eta \sqrt{K}}{\lambda}.
        \end{align*}
    \end{lemma}
    In the proof of Lemma 20 by~\citet{zhang2024online}, the bound for $ \| \nabla \ell_t (\wb_t) \|_2$ was given as $ \| \nabla \ell_t (\wb_t) \|_2 \leq 2 \sqrt{K}$.
    However, a tighter bound can be established: $ \| \nabla \ell_t (\wb_t) \|_2 \leq 2$.
    Using this tighter bound, we can refine and derive the improved result.
\end{proof}

\section{Proofs of Theorem~\ref{theorem:lower_bound_non_uniform}} 
\label{app_sec:proof_for_lower_non_uniform}
In this section, we provide the proof of Theorem~\ref{theorem:lower_bound_non_uniform}.
\subsection{Adversarial Rewards Construction}
\label{app_subsec:lower_adversarial_setting_non_uniform}
We follow a similar context construction as in Section~\ref{app_subsec:lower_adversarial_setting}, but now consider the case where all context vectors are distinct and the rewards are non-uniform.
Accordingly, the total number of items is $N = \binom{d}{d/4}$.
Given $\wb_V$, we define a unique item $i^\star \in [N]$ as the one that maximizes $x_i^\top \wb_V$, i.e., $x_{i^\star} = x_V$.
In addition, let $i' \in [N] \setminus {i^\star}$ denote the item with the second-largest value of $x_i^\top \wb_V$.
At each round $t$, we assign a reward of $1$ to $i^\star$ when $t$ is odd, and a reward of $1$ to one randomly selected item other than $i^\star$ when $t$ is even.
For simplicity, let $\Tcal^\star \subseteq [T]$ denote the set of rounds corresponding to the former case, i.e., when the reward $1$ is assigned to $i^\star$.
Specifically, for $t \in \Tcal^\star$, the rewards are constructed as follows:
\begin{align*}
    r_{ti} = 
    \begin{cases}
    1, & \text{for } i = i^\star\\
    \nu, & \text{for } i = i' \\
    \gamma, & \text{otherwise},
    \end{cases}
\end{align*}
where $\nu$ and $\gamma$ are defined as
\begin{align*}
    \nu 
    = \frac{e^{d/4 - 0.5}}{v_0 + e^{d/4 - 0.5} },
    \numberthis  \label{eq:def_nu}
\end{align*}
and
\begin{align*}
    \gamma = \min_{S \in \mathcal{S}} \frac{\min_{i \in S} \exp( x_i^\top \wb_V ) \nu }{v_0 + \min_{i \in S} \exp( x_i^\top \wb_V )}
    = \frac{\nu}{v_0 + 1} < 1.
\end{align*}
Moreover, for rounds $t \notin \Tcal^\star$, let $\hat{i} \in [N] \setminus {i^\star}$ denote the randomly selected item (other than $i^\star$) that receives a reward of $1$.
The rewards are then defined as follows:
\begin{align*}
    r_{ti} = 
    \begin{cases}
    1, & \text{for } i = \hat{i}\\
    \nu, & \text{for } i = i' \\
    \gamma, & \text{otherwise}.
    \end{cases}
\end{align*}
%

\subsection{Main Proof of Theorem~\ref{theorem:lower_bound_non_uniform}}
\label{app_subsec:lower_bound_non_uniform_main}
\begin{proof}[Proof of Theorem~\ref{theorem:lower_bound_non_uniform}]
We begin the proof by introducing useful lemmas.
\begin{lemma} 
\label{lemma:opt_S}
Let $R_t(S^\star_t, \wb_V) = \frac{\sum_{i \in S^\star_t} \exp( x_{i}^\top \wb_V ) r_{ti} }{v_0 + \sum_{j \in S^\star_t} \exp(x_{j}^\top \wb_V) }$.
Then, we have
    \begin{align*}
        S^\star_t =
        \begin{cases}
            \{ i^\star \} & \text{for } t \in \Tcal^\star 
            \\
            \{ \hat{i}, i' \}  & \text{for } t \notin \Tcal^\star.
        \end{cases}
    \end{align*}
\end{lemma}
Lemma~\ref{lemma:opt_S} establishes that any policy which relies solely on  \textit{rewards}—without estimating the underlying parameter—incurs a regret of order $\Omega(T)$ in this instance.
In particular, the reward distributions over rounds in $\Tcal^\star$ and $[T] \setminus \Tcal^\star$ are identical up to item permutations, making them indistinguishable based solely on the given rewards.
Consequently, reward-based heuristics—such as selecting the top-1 or top-2 items, or choosing items whose rewards exceed a fixed threshold—necessarily suffer $\Omega(T)$ regret on $\Tcal^\star$, $[T] \setminus \Tcal^\star$, or both.
For instance, the top-1 strategy incurs zero regret for $t \in \Tcal^\star$ but $\Omega(T)$ regret for $t \notin \Tcal^\star$, since the optimal assortment size in that case is 2.
Conversely, the top-2 strategy suffers $\Omega(T)$ regret for $t \in \Tcal^\star$ and zero regret for $t \notin \Tcal^\star$.
Similarly, a threshold-based policy that selects items whose rewards exceed a fixed value yields a fixed-size assortment for all $t$ (since the reward distributions are identical across rounds), and therefore necessarily incurs $\Omega(T)$ regret in one or both cases.
Therefore, we can safely exclude such reward-based heuristics from consideration and focus our analysis on the rounds in $\Tcal^\star$.
\begin{lemma} 
\label{lemma:non_uniform_revenue_upperbound}
For any $t \in \Tcal^\star$ and $i^\star \notin S_t$, we have
    \begin{align*}
        R_t(S_t, \wb_V) \leq \frac{\max_{i \in S_t}\exp(x_i^\top \wb_V) }{v_0 + \max_{i \in S_t}\exp(x_i^\top \wb_V)}.
    \end{align*}
\end{lemma}
Lemma~\ref{lemma:non_uniform_revenue_upperbound} shows that, for $t \in \Tcal^\star$ and $i^\star \notin S_t$, the expected revenue of $S_t$ can be upper bounded by that of a singleton set whose reward has been replaced with $1$.

Now, we are ready to provide the proof of Theorem~\ref{theorem:lower_bound_non_uniform}.
Let $x_{U_1}, \dots, x_{U_L}$ be the distinct feature vectors contained in assortments $S_t$ with $U_1, \dots, U_L \in \mathcal{V}_{d/4}$.
Let $\tilde{U}_t$ be the (unique) item among $U_1, \dots, U_L$ that maximizes $x_U^\top \wb_V$, i.e., $\tilde{U}_t = \argmax_{U \in \{U_1, \dots, U_L \}} x_U^\top \wb_V$, where $\wb_V$ is the underlying parameter.
Then, we have
\begin{align*}
    \sum_{t=1}^T &R_t(S^\star_t, \wb_V) - R_t(S_{t}, \wb_V) 
    \\
    &\geq \sum_{t \in \Tcal^\star} R_t(S^\star_t, \wb_V) - R_t(S_{t}, \wb_V) 
    \\
    &=  \sum_{t \in \Tcal^\star} \frac{\exp(x_V^\top \wb_V )}{v_0 + \exp(x_V^\top \wb_V )} - R_t(S_t, \wb_V) 
    \\
    &\geq  \sum_{t \in \Tcal^\star} 
    \mathbf{1} \{ i^\star \notin S_t \}
    \left(
    \frac{\exp(x_V^\top \wb_V )}{v_0 + \exp(x_V^\top \wb_V )} - \frac{\max_{i \in S_t}\exp(x_i^\top \wb_V) }{v_0 + \max_{i \in S_t}\exp(x_i^\top \wb_V)} \right)
    \\
    &=  \sum_{t \in \Tcal^\star} 
    \mathbf{1} \{ i^\star \notin S_t \}
    \left(
    \frac{\exp(x_V^\top \wb_V )}{v_0 + \exp(x_V^\top \wb_V )} - \frac{\exp(x_{\tilde{U}_t}^\top \wb_V) }{v_0 + \exp(x_{\tilde{U}_t}^\top \wb_V)}
    \right),
\end{align*}
where the first equality holds by Lemma~\ref{lemma:opt_S} (and recall that $x_{i^\star} = x_V$), 
and
the second inequality holds by Lemma~\ref{lemma:non_uniform_revenue_upperbound}.
Then, by applying Lemma~\ref{lemma:lower_revenue_gap} with $K=1$, we can further bound the regret as follows:
\begin{align*}
    \sum_{t=1}^T R_t(S^\star_t, \wb_V) - R_t(S_{t}, \wb_V) 
    \geq \frac{v_0}{(v_0 + e)^2} \sum_{t \in \Tcal^\star} 
    \frac{\delta \epsilon}{2 \sqrt{d}},
\end{align*}
where $\epsilon \in (0, 1/d\sqrt{d})$ and $\delta = d/4 - | \tilde{U}_t \cap V |$.
Note that for $i^\star \in S_t$, we have $\tilde{U}_t = V$, and hence we get $\delta = 0$.

The remainder of the analysis follows directly from the proof of Theorem~\ref{thm:lower_bound} with $K = 1$ and $\Tcal^\star = T/2$.
Therefore, setting $v_0 = \Theta(1)$, we obtain
\begin{align*}
    \sup_{\wb} \EE_{\wb}^{\pi} \left[ \Regret(\wb) \right]
    &\geq \frac{1}{|\mathcal{V}_{d/4}|} \sum_{V \in \mathcal{V}_{d/4}} \EE_{\wb_V}^{\pi} \sum_{t=1}^T R_t(S^\star_t, \wb_V) - R_t(S_{t}, \wb_V) 
    = \Omega \left( d\sqrt{T} \right).
\end{align*}
This concludes the proof of Theorem~\ref{theorem:lower_bound_non_uniform}.
\end{proof}

\subsection{Proofs of Lemmas for Theorem~\ref{theorem:lower_bound_non_uniform}}
\subsubsection{Proof of Lemma~\ref{lemma:opt_S}}
\label{app_subsubsec:proof_of_lemma:opt_S}
\begin{proof}[Proof of Lemma~\ref{lemma:opt_S}]
    First, for any $t \in [T]$ and all $ i \in S^\star_t$, 
    we show that $ r_{ti} \geq R_t(S^\star_t, \wb_V)$.
    We prove this by contradiction.
    Assume that there exists $i \in S^\star_t$ such that $r_{ti} < R_t(S^\star_t, \wb_V)$.
    Then, removing the item $i$ from the assortment $ S^\star_t$ yields higher expected revenue.
    This contradicts the optimality of $S^\star_t$.
    Thus, for any $t \in [T]$, we have
    \begin{align*}
        r_{ti} \geq R_t(S^\star_t, \wb_V), \quad \forall i \in S^\star_t.
        \numberthis \label{eq:opt_S_property}
    \end{align*}
    Hence, 
    for $t \in \Tcal^\star$, we have  $i^\star \in S_t^\star$, since $r_{i^\star} = 1 >  R_t(S^\star_t, \wb_V)$.
    Furthermore, for any other item $i \neq i^\star $,  by definition of $\nu$ in Equation~\eqref{eq:def_nu}, we have 
    \begin{align*}
        r_{ti} \leq 
        \nu
        =\frac{e^{d/4 - 0.5}}{v_0 + e^{d/4 - 0.5} }
        < \frac{e^{d/4 }}{v_0 + e^{d/4 } }
        =  \frac{\exp(x_V^\top \wb_V) }{v_0 + \exp(x_V^\top \wb_V)} 
        = R_t(\{i^\star\}, \wb_V).
    \end{align*}
    Therefore, 
    by Equation~\eqref{eq:opt_S_property}, 
    only the item $i^\star$ is included in the optimal assortment for all $t \in \Tcal^\star$.

    We next analyze the case $t \notin \Tcal^\star$. 
    Using the same reasoning as above and applying Equation~\eqref{eq:opt_S_property}, we obtain that $\hat{i} \in S_t^\star$, since $r_{t\hat{i}} = 1 > R_t(S_t^\star, \wb_V)$.
    Moreover, we have
    \begin{align*}
        r_{ti'} = \nu = \frac{e^{d/4 - 0.5}}{v_0 + e^{d/4 - 0.5} }
        >\frac{e^{d/4 - 1}}{v_0 + e^{d/4 - 1} }
        =  R_t(\{\hat{i}\}, \wb_V),
    \end{align*}
    which implies $R_t(\{\hat{i}\}, \wb_V)
        <  R_t(\{\hat{i}, i'\}, \wb_V)$.
    And for any $i \notin \{ \hat{i}, i' \}$,
    \begin{align*}
        r_{ti} = \gamma
        = \frac{\nu}{v_0 + 1}
        < \frac{1}{v_0 + 1}
        < R_t(\{\hat{i}\}, \wb_V)
        <  R_t(\{\hat{i}, i'\}, \wb_V).
    \end{align*}
    Therefore,  by Equation~\eqref{eq:opt_S_property}, we can conclude that  $S^\star_t = \{ \hat{i}, i' \}  $.
    This completes the proof.
\end{proof}
\subsubsection{Proof of Lemma~\ref{lemma:non_uniform_revenue_upperbound}}
\begin{proof}[Proof of Lemma~\ref{lemma:non_uniform_revenue_upperbound}]
    We provide a proof by considering the following cases:
    
    \textbf{Case 1. } $i' \in S_t$ and $i^\star \notin S_t$.

    Recall that, by the construction of rewards, we have
    \begin{align*}
        \gamma 
        =  \min_{S \in \mathcal{S}} \frac{\min_{i \in S} \exp( x_i^\top \wb_V ) \nu }{v_0 + \min_{i \in S} \exp( x_i^\top \wb_V )}
        \leq \frac{\min_{i \in S_t} \exp( x_i^\top \wb_V ) \nu }{v_0 + \min_{i \in S_t} \exp( x_i^\top \wb_V )}
        \leq \frac{\exp(x_{i'}^\top \wb_V) \nu}{v_0 + \exp(x_{i'}^\top \wb_V)}.
        \numberthis 
        \label{eq:gamma_upperbound}
    \end{align*}
    This implies that 
    \begin{align*}
        & \left\{ \sum_{i \in S_t \setminus \{i'\} } \exp(x_i^\top \wb_V) \right\} 
        \frac{\left(v_0 + \exp(x_{i'}^\top \wb_V) \right)  \nu}{ v_0 + 1}
        \leq \left\{ \sum_{i \in S_t \setminus \{i'\} } \exp(x_i^\top \wb_V) \right\} \exp(x_{i'}^\top \wb_V) \nu
        \\ 
        &\Leftrightarrow \left( \exp(x_{i'}^\top \wb_V) + 
        \frac{\sum_{i \in S_t \setminus \{i'\} } \exp(x_i^\top \wb_V)}{v_0 + 1}  \right) \left(v_0 + \exp(x_{i'}^\top \wb_V) \right) 
        \\
        &\quad\,\,\leq \exp(x_{i'}^\top \wb_V) \left( v_0 + \exp(x_{i'}^\top \wb_V) +  \sum_{i \in S_t \setminus \{i'\} } \exp(x_i^\top \wb_V) \right)
        \\
        &\Leftrightarrow \frac{ \exp(x_{i'}^\top \wb_V) + \sum_{i \in S_t \setminus \{i'\} } \exp(x_i^\top \wb_V)/(v_0 + 1)  }{v_0 + \sum_{i \in S_t} \exp(x_i^\top \wb_V)}
        \leq \frac{\exp(x_{i'}^\top \wb_V)}{v_0 + \exp(x_{i'}^\top \wb_V)}
        .
        \numberthis \label{eq:gamma_revenue}
    \end{align*}
    Therefore, we have
    \begin{align*}
        R(S_t, \wb_V) 
        &= \frac{\sum_{i \in S_t} \exp(x_i^\top \wb_V) r_{ti}}{v_0 + \sum_{i \in S_t} \exp(x_i^\top \wb_V)}
        = \frac{ \exp(x_{i'}^\top \wb_V) \nu + \sum_{i \in S_t \setminus \{i'\} } \exp(x_i^\top \wb_V) \cdot \nu/(v_0 + 1)  }{v_0 + \sum_{i \in S_t} \exp(x_i^\top \wb_V)}
        \\
        &\leq \frac{\exp(x_{i'}^\top \wb_V) \nu}{v_0 + \exp(x_{i'}^\top \wb_V)}
        \leq \frac{\max_{i \in S_t}\exp(x_i^\top \wb_V) }{v_0 + \max_{i \in S_t}\exp(x_i^\top \wb_V)},
    \end{align*}
    where the first inequality holds by Equation~\eqref{eq:gamma_revenue}.

    \textbf{Case 2. } $i', i^\star \notin S_t$.

    Let us return to Equation~\eqref{eq:gamma_upperbound}.
    Since $\frac{v_0 + \sum_{i \in S_t} \exp (x_i^\top \wb_V) }{\sum_{i \in S_t} \exp (x_i^\top \wb_V)} \geq 1$, we have
    \begin{align*}
        \gamma 
        \leq \frac{\min_{i \in S_t} \exp( x_i^\top \wb_V ) }{v_0 + \min_{i \in S_t} \exp( x_i^\top \wb_V )}
        \leq \frac{\min_{i \in S_t} \exp( x_i^\top \wb_V ) }{v_0 + \min_{i \in S_t} \exp( x_i^\top \wb_V )}
        \cdot
        \frac{v_0 + \sum_{i \in S_t} \exp (x_i^\top \wb_V) }{\sum_{i \in S_t} \exp (x_i^\top \wb_V)}
        ,
    \end{align*}
    which is equivalent to
    \begin{align*}
        \frac{\sum_{i \in S_t} \exp (x_i^\top \wb_V) \gamma}{v_0 + \sum_{i \in S_t} \exp (x_i^\top \wb_V) } 
        \leq \frac{\min_{i \in S_t} \exp( x_i^\top \wb_V ) }{v_0 + \min_{i \in S_t} \exp( x_i^\top \wb_V )}
        .
    \end{align*}
    Hence, we get
    \begin{align*}
        R(S_t, \wb_V) 
        &=  \frac{\sum_{i \in S_t} \exp (x_i^\top \wb_V) \gamma}{v_0 + \sum_{i \in S_t} \exp (x_i^\top \wb_V) } 
        \leq \frac{\min_{i \in S_t} \exp( x_i^\top \wb_V ) }{v_0 + \min_{i \in S_t} \exp( x_i^\top \wb_V )}
        \\
        &\leq \frac{\max_{i \in S_t}\exp(x_i^\top \wb_V) }{v_0 + \max_{i \in S_t}\exp(x_i^\top \wb_V)}
        .
    \end{align*}
    This concludes the proof.
\end{proof}

\section{Proofs of Theorem~\ref{theorem:upper_bound_non_uniform}} 
\label{app_sec:upper_bound_non_uniform}
In this section, we provide the proof of Theorem~\ref{theorem:upper_bound_non_uniform}.
Since we now consider the case of non-uniform rewards, the sizes of both the chosen assortment $S_t$, and the optimal assortment, $S_t^\star$ are no longer fixed at $K$.

We begin the proof by introducing additional useful lemmas.
Lemma~\ref{lemma:optimizm}  shows that $\tilde{R}_{t}(S_t)$, defined in Equation~\eqref{eq:opt_revenue}, is an upper bound of the true expected revenue of
the optimal assortment, $R_{t}(S_t^\star, \wb^\star) $.
\begin{lemma} [Lemma 4 in~\citealt{oh2021multinomial}] \label{lemma:optimizm}
Let 
$\tilde{R}_{t}(S) = \frac{\sum_{i \in S} \exp( \alpha_{ti} ) r_{ti} }{v_0 + \sum_{j \in S} \exp(\alpha_{tj}) }$.
And suppose $S_t = \argmax_{S \in \mathcal{S}} \tilde{R}_{t}(S)$.
If for every item $i \in S_t^\star$, $\alpha_{ti} \geq x_{ti}^\top \wb^\star$, then for all $t \geq 1$, the following inequalities hold:
    \begin{align*}
        R_{t}(S_{t}^\star, \wb^\star) \leq \tilde{R}_t(S_{t}^\star)
        \leq \tilde{R}_t(S_{t}).
    \end{align*}
\end{lemma}
Note that Lemma~\ref{lemma:optimizm} does not claim that the expected revenue is a monotone function in general. 
Instead, it specifically states that the value of the expected revenue, when associated with the optimal assortment, increases with an increase in the MNL parameters~\citep{agrawal2019mnl, oh2021multinomial}.

Lemma~\ref{lemma:increasing_R} shows that $\tilde{R}_{t}(S_t)$ increases as the utilities of items in $S_t$ increase.

\begin{lemma} \label{lemma:increasing_R}
    Let 
    $\tilde{R}_{t}(S) = \frac{\sum_{i \in S} \exp( \alpha_{ti} ) r_{ti} }{v_0 + \sum_{j \in S} \exp(\alpha_{tj}) }$ and $S_t = \argmax_{S \in \mathcal{S}} \tilde{R}_{t}(S)$.
    Assume $\alpha'_{ti} \geq \alpha_{ti} \geq 0$ for all $i \in [N]$.
    Then, we have
    \begin{align*}
        \tilde{R}_{t}(S_t) 
        \leq \frac{\sum_{i \in S_t} \exp( \alpha'_{ti} ) r_{ti} }{v_0 + \sum_{j \in S_t} \exp(\alpha'_{tj}) }.
    \end{align*}
\end{lemma}

Furthermore, we provide a novel elliptical potential Lemma~\ref{lemma:elliptical_x_tilde} for the \textit{centralized} context vectors $\tilde{x}_{ti}$.
\begin{lemma}   \label{lemma:elliptical_x_tilde}
Let $H_t = \lambda \Ib_d + \sum_{s=1}^{t-1} \Gcal_s(\wb_{s+1})$, where $\Gcal_s(\wb) = \sum_{i \in S_s} p_s(i | S_s, \wb) x_{si} x_{si}^\top -  \sum_{i \in S_s}  \sum_{j \in S_s} p_s(i | S_s, \wb) p_s(j | S_s, \wb) x_{si} x_{sj}^\top$.
Define $\tilde{x}_{si} = x_{si} - \EE_{ j \sim p_s(\cdot | S_s, \wb_{s+1}) }[x_{sj}]$.
Suppose $\lambda \geq 2$.
Then the following statements hold true:
\begin{enumerate}[label={(\arabic*)}]

    \item $\sum_{s=1}^{t} \sum_{i \in S_s} p_s(i | S_s, \wb_{s+1}) \| \tilde{x}_{si}\|_{H_s^{-1}}^2
    \leq 2d \log \left( 1+ \frac{t}{d \lambda} \right)$,
    
    \item $\sum_{s=1}^{t} \max_{i \in S_s} \| \tilde{x}_{si}\|_{H_s^{-1}}^2
    \leq \frac{2}{\kappa} d \log \left( 1+ \frac{t}{d \lambda} \right)$.

\end{enumerate}
\end{lemma}
Now, we prove the Theorem~\ref{theorem:upper_bound_non_uniform}.

%
\subsection{Proof of Theorem~\ref{theorem:upper_bound_non_uniform}}
\label{app_subsec:proof_theorem:upper_bound_non_uniform}
\begin{proof}[Proof of Theorem~\ref{theorem:upper_bound_non_uniform}]
    Let $\alpha'_{ti} = x_{ti}^\top \wb^\star + 2 \beta_t(\delta) \| x_{ti} \|_{H_t^{-1}}$.
    If $\wb^\star \in \mathcal{C}_t(\delta)$, then, by Lemma~\ref{lemma:utility}, we have
    \begin{align*}
        \alpha_{ti} 
        \leq x_{ti}^\top \wb^\star + 2 \beta_t(\delta) \| x_{ti} \|_{H_t^{-1}}
        = \alpha'_{ti}.
    \end{align*}
    We denote $\dbtilde{R}_t(S_{t}) = \frac{\sum_{i \in S_t} \exp( \alpha'_{ti} ) r_{ti} }{v_0 + \sum_{j \in S_t} \exp(\alpha'_{tj}) }$.
    Then, we can bound the regret as follows:
    \begin{align*}
        \sum_{t=1}^T  R_{t}(S_{t}^\star, \wb^\star) -  R_{t}(S_{t}, \wb^\star) 
        \leq \sum_{t=1}^T  \tilde{R}_t(S_{t}) -  R_{t}(S_{t}, \wb^\star)  
        \leq \sum_{t=1}^T  \dbtilde{R}_t(S_{t}) -  R_{t}(S_{t}, \wb^\star),  
    \end{align*}
    where
    the first inequality holds by Lemma~\ref{lemma:optimizm} and the last inequality holds by Lemma~\ref{lemma:increasing_R}.
    
    Now, we define $\tilde{Q}:\RR^{|S_t|} \rightarrow \RR$, such that for all $\ub = (u_1, \dots, u_{|S_t|})^\top \in \RR^{|S_t|}$, $\tilde{Q}(\ub) = \sum_{k=1}^{|S_t|} \frac{\exp(u_k)r_{ti_k} }{v_0 + \sum_{j=1}^{|S_t|} \exp(u_j)}$.
    Let $S_t = \{i_1, \dots, i_{|S_t|} \}$.
    Moreover, for all $t \geq 1$, let $\ub_t = (u_{ti_1}, \dots u_{ti_{|S_t|}} )^\top = (\alpha'_{ti_1} , \dots, \alpha'_{ti_{{|S_t|}}} )^\top$ 
    and $\ub^\star_t = (u_{ti_1}^\star, \dots u_{ti_{|S_t|}}^\star )^\top = (x_{ti_1}^\top \wb^\star, \dots, x_{ti_{{|S_t|}}}^\top \wb^\star)^\top$.
    Then, by applying a second order Taylor expansion, we obtain
    \begin{align*}
        \sum_{t=1}^T \dbtilde{R}_t(S_{t}) -  R_{t}(S_{t}, \wb^\star)  
        &=  \sum_{t=1}^T  \tilde{Q}(\ub_t) - \tilde{Q}(\ub^\star_t) 
        \\
        &=  \underbrace{  \sum_{t=1}^T \nabla \tilde{Q}(\ub_t^\star)^\top (\ub_t - \ub^\star_t) }_{\texttt{(C)}}
        + \underbrace{\frac{1}{2} \sum_{t=1}^T (\ub_t - \ub^\star_t)^\top \nabla^2 \tilde{Q}(\bar{\ub}_t) (\ub_t - \ub^\star_t)}_{\texttt{(D)}},
    \end{align*}
    where $\bar{\ub}_t = (\bar{u}_{ti_1}, \dots, \bar{u}_{ti_{|S_t|}})^\top \in \RR^{|S_t|}$ is the convex combination of $\ub_t$ and $\ub^\star_t$.
    
    We first bound the term \texttt{(C)}.
    \begin{align*}
        & \sum_{t=1}^T \nabla \tilde{Q}(\ub_t^\star)^\top (\ub_t - \ub^\star_t) 
        \\
        &=  \sum_{t=1}^T \sum_{i \in S_t} \frac{\exp(x_{ti}^\top \wb^\star )r_{ti}}{v_0 + \sum_{k\in S_t} \exp(x_{tk}^\top \wb^\star ) } (u_{ti} - u_{ti}^\star ) 
        -  \sum_{j \in S_t}   \frac{\exp(x_{tj}^\top \wb^\star ) r_{tj} \sum_{i \in S_t} \exp(x_{ti}^\top \wb^\star )}{(v_0 + \sum_{k\in S_t} \exp(x_{tk}^\top \wb^\star ))^2 } (u_{ti} - u_{ti}^\star )   
        \\
        &=  \sum_{t=1}^T \sum_{i \in S_t} p_t(i | S_t, \wb^\star) r_{ti} (u_{ti} - u_{ti}^\star )
        - \sum_{i \in S_t}  \sum_{j \in S_t}  p_t(i | S_t, \wb^\star) r_{ti} p_t(j | S_t, \wb^\star)  (u_{tj} - u_{tj}^\star )
        \\
        &=  \sum_{t=1}^T \sum_{i \in S_t} p_t(i | S_t, \wb^\star)  r_{ti} \left( (u_{ti} - u_{ti}^\star ) -  \sum_{j \in S_t} p_t(j | S_t, \wb^\star)  (u_{tj} - u_{tj}^\star )  \right)
        \\
        &=  \sum_{t=1}^T \sum_{i \in S_t} p_t(i | S_t, \wb^\star)  r_{ti} \left( 2 \beta_t(\delta) \| x_{ti} \|_{H_t^{-1}} -  \sum_{j \in S_t} p_t(j | S_t, \wb^\star)  2 \beta_t(\delta) \| x_{tj} \|_{H_t^{-1}} \right) 
        \\
        &= 2 \sum_{t=1}^T  \beta_t(\delta) \sum_{i \in S_t} p_t(i | S_t, \wb^\star)  r_{ti} \left(  \| x_{ti} \|_{H_t^{-1}} -  \sum_{j \in S_t} p_t(j | S_t, \wb^\star) \| x_{tj} \|_{H_t^{-1}} \right).
    \end{align*}
    Let $x_{t0} = \mathbf{0}$.
    Then, we can further bound the right-hand side as follows:
    \begin{align*}
        2 \sum_{t=1}^T \beta_t(\delta) &\sum_{i \in S_t} p_t(i | S_t, \wb^\star)  r_{ti} \left(  \| x_{ti} \|_{H_t^{-1}} -  \sum_{j \in S_t} p_t(j | S_t, \wb^\star) \| x_{tj} \|_{H_t^{-1}} \right)
        \\
        &= 2 \sum_{t=1}^T \beta_t(\delta) \sum_{i \in S_t} p_t(i | S_t, \wb^\star)  r_{ti} \left(  \| x_{ti} \|_{H_t^{-1}} -  \sum_{j \in S_t\cup \{0\}} p_t(j | S_t, \wb^\star) \| x_{tj} \|_{H_t^{-1}} \right)
        \\
        &= 2 \sum_{t=1}^T \beta_t(\delta) \sum_{i \in S_t} p_t(i | S_t, \wb^\star)  r_{ti} \left(  \| x_{ti} \|_{H_t^{-1}} -  \EE_{j \sim p_t(\cdot | S_t, \wb^\star)} \left[ \| x_{tj} \|_{H_t^{-1}}  \right] \right)
        \\
        &\leq 2 \sum_{t=1}^T \beta_t(\delta) \sum_{i \in S_t^{+}} p_t(i | S_t, \wb^\star)   \left( \| x_{ti} \|_{H_t^{-1}} -  \EE_{j \sim p_t(\cdot | S_t, \wb^\star)} \left[ \| x_{tj} \|_{H_t^{-1}}  \right] \right)
        \\
        &\leq 2 \beta_T(\delta) \sum_{t=1}^T  \sum_{i \in S_t^{+}} p_t(i | S_t, \wb^\star)   \left( \| x_{ti} \|_{H_t^{-1}} -  \EE_{j \sim p_t(\cdot | S_t, \wb^\star)} \left[ \| x_{tj} \|_{H_t^{-1}}  \right] \right)
        \\
        &\leq  2  \beta_T(\delta) \sum_{t=1}^T \sum_{i \in S_t^{+}} p_t(i | S_t, \wb^\star)   \left( \| x_{ti} \|_{H_t^{-1}}  -  \left\|  \EE_{j \sim p_t(\cdot | S_t, \wb^\star)} \left[ x_{tj} \right]   \right\|_{H_t^{-1}}      \right)
        \\
        &\leq   2  \beta_T(\delta) \sum_{t=1}^T \sum_{i \in S_t^{+}} p_t(i | S_t, \wb^\star)    \| x_{ti} - \EE_{j \sim p_t(\cdot | S_t, \wb^\star)} \left[ x_{tj} \right]  \|_{H_t^{-1}}    
        \\
        &\leq   2  \beta_T(\delta) \sum_{t=1}^T \sum_{i \in S_t} p_t(i | S_t, \wb^\star)    \left\| x_{ti} - \EE_{j \sim p_t(\cdot | S_t, \wb^\star)} \left[ x_{tj} \right]  \right\|_{H_t^{-1}}   
        ,
    \end{align*}
    where, in the first inequality, we define $S_{t}^{+} \subseteq S_{t}$ as the subset of items in $S_t$ such that the term $ \| x_{ti} \|_{H_t^{-1}} -  \EE_{j \sim p_t(\cdot | S_t, \wb^\star)} \left[ \| x_{tj} \|_{H_t^{-1}}  \right] \geq 0$ and $r_{ti} \in [0,1]$,
    the second inequality holds because $\beta_1(\delta) \leq \dots \leq \beta_T(\delta)$, 
    the third inequality holds due to Jensen's inequality, 
    and the second-to-last inequality holds due to the fact that $\| \ab \|  = \| \ab - \bb + \bb\| \leq \| \ab-\bb \| + \|\bb\|$ for any vectors $\ab, \bb \in \RR^d$.

    For simplicity, we denote $\EE_{\wb}[x_{tj}] = \EE_{j \sim p_t(\cdot | S_t, \wb)}[x_{tj}]$.
    Let $\bar{x}_{ti} = x_{ti} - \EE_{ \wb^\star}[x_{tj}]$ and $\tilde{x}_{ti} = x_{ti} - \EE_{ \wb_{t+1}}[x_{tj}]$.
    Then, we have
    \begin{align*}
        &\sum_{t=1}^T \sum_{i \in S_t} p_t(i | S_t, \wb^\star)    \| x_{ti} - \EE_{j \sim p_t(\cdot | S_t, \wb^\star)} \left[ x_{tj} \right]  \|_{H_t^{-1}}    
        = \sum_{t=1}^T \sum_{i \in S_t} p_t(i | S_t, \wb^\star) \| \bar{x}_{ti} \|_{H_t^{-1}}
        \\
        &\leq \sum_{t=1}^T \sum_{i \in S_t} p_t(i | S_t, \wb^\star) \| \bar{x}_{ti} - \tilde{x}_{ti} \|_{H_t^{-1}} 
        + \sum_{t=1}^T \sum_{i \in S_t} p_t(i | S_t, \wb^\star) \| \tilde{x}_{ti} \|_{H_t^{-1}} 
        \\
        &= \sum_{t=1}^T \sum_{i \in S_t} p_t(i | S_t, \wb^\star) \| \bar{x}_{ti} - \tilde{x}_{ti} \|_{H_t^{-1}} 
        + \sum_{t=1}^T \sum_{i \in S_t} \left(  p_t(i | S_t, \wb^\star) - p_t(i | S_t, \wb_{t+1}) \right)\| \tilde{x}_{ti} \|_{H_t^{-1}}
        \\
        &+ \sum_{t=1}^T \sum_{i \in S_t} p_t(i | S_t, \wb_{t+1}) \| \tilde{x}_{ti} \|_{H_t^{-1}}
        ,
        \numberthis \label{eq:non_unform_bouns_fist_decomposition}
    \end{align*}
    where the inequality holds by the triangle inequality.
    Now, we bound the terms on the right-hand side of Equation~\eqref{eq:non_unform_bouns_fist_decomposition} individually.
    For the first term, we have
    \begin{align*}
        &\sum_{t=1}^T \sum_{i \in S_t} p_t(i | S_t, \wb^\star) \| \bar{x}_{ti} - \tilde{x}_{ti} \|_{H_t^{-1}} 
        =   \sum_{t=1}^T \sum_{i \in S_t} p_t(i | S_t, \wb^\star) 
        \left\| \EE_{\wb_{t+1}}[x_{tj}]
         -  \EE_{\wb^\star}[x_{tj}]  \right\|_{H_t^{-1}}
         \\
         &= \sum_{t=1}^T \sum_{i \in S_t} p_t(i | S_t, \wb^\star) 
        \left\| \sum_{j \in S_t} \left( p_t(j | S_t, \wb_{t+1}) -   p_t(j | S_t, \wb^\star) \right) x_{tj}  \right\|_{H_t^{-1}}
        ,
    \end{align*}
    where the last equality holds due to the setting of $x_{t0} = \mathbf{0}$.
    By the mean value theorem, there exists $\xib_t = (1-c) \wb^\star + c \wb_{t+1}$ for some $c \in (0,1)$ such that
    \begin{align*} 
        &\left\| \sum_{j \in S_t} \left( p_t(j | S_t, \wb_{t+1}) -   p_t(j | S_t, \wb^\star) \right) x_{tj}  \right\|_{H_t^{-1}}
        = \left\| \sum_{j \in S_t} \nabla p_t(j | S_t, \xib_t)^\top (\wb_{t+1} - \wb^\star)  x_{tj}  \right\|_{H_t^{-1}}
        \\
        &\leq  \sum_{j \in S_t} \left| \nabla p_t(j | S_t, \xib_t)^\top (\wb_{t+1} - \wb^\star) \right| \left\|x_{tj}  \right\|_{H_t^{-1}}
        \\
        &= \sum_{j \in S_t} \left| \left( p_t(j | S_t, \xib_t) x_{tj} - p_t(j | S_t, \xib_t) \sum_{k \in S_t} p_t(k | S_t, \xib_t) x_{tk}   \right)^\top  (\wb_{t+1} - \wb^\star) \right| \left\|x_{tj}  \right\|_{H_t^{-1}}
        \\
        &\leq \sum_{j \in S_t}   p_t(j | S_t, \xib_t) \left| x_{tj}^\top (\wb_{t+1} - \wb^\star) \right|    \left\|x_{tj}  \right\|_{H_t^{-1}}
        \\
        &+ \sum_{j \in S_t}   p_t(j | S_t, \xib_t) \left\|x_{tj}  \right\|_{H_t^{-1}} \sum_{k \in S_t}  p_t(k | S_t, \xib_t) \left| x_{tk}^\top  (\wb_{t+1} - \wb^\star) \right| 
        \\
        &\leq \sum_{j \in S_t}  p_t(j | S_t, \xib_t)  \|\wb_{t+1} - \wb^\star\|_{H_t}    \left\|x_{tj}  \right\|_{H_t^{-1}}^2
        \\
        &+ \sum_{j \in S_t}   p_t(j | S_t, \xib_t) \left\|x_{tj}  \right\|_{H_t^{-1}} \sum_{k \in S_t}  p_t(k | S_t, \xib_t) \| x_{tk}\|_{H_t^{-1}}  \| \wb_{t+1} - \wb^\star \|_{H_t}
        \\
        &\leq \beta_{t+1}(\delta)  \sum_{j \in S_t}  p_t(j | S_t, \xib_t)      \left\|x_{tj}  \right\|_{H_t^{-1}}^2
        +
        \beta_{t+1}(\delta) \left( \sum_{j \in S_t}   p_t(j | S_t, \xib_t) \left\|x_{tj}  \right\|_{H_t^{-1}} \right)^2
        \\
        &\leq 2  \beta_{t+1}(\delta)  \sum_{j \in S_t}  p_t(j | S_t, \xib_t)      \left\|x_{tj}  \right\|_{H_t^{-1}}^2
        \\
        &
        \leq  2  \beta_{t+1}(\delta)  \max_{j \in S_t}  \left\|x_{tj}  \right\|_{H_t^{-1}}^2
        ,
    \end{align*}
    where the fourth inequality holds by Lemma~\ref{lemma:online_confidence_set} 
    and the second-to-last inequality holds 
    due to Jensen's inequality.
    Hence, we have
    \begin{align*}
        \sum_{t=1}^T \sum_{i \in S_t} p_t(i | S_t, \wb^\star) \| \bar{x}_{ti} - \tilde{x}_{ti} \|_{H_t^{-1}}
        &\leq  2     \sum_{t=1}^T \beta_{t+1}(\delta)  \sum_{i \in S_t} p_t(i | S_t, \wb^\star)
         \max_{j \in S_t}  \left\|x_{tj}  \right\|_{H_t^{-1}}^2
         \\
        &\leq 2 \beta_{T+1}(\delta)  \sum_{t=1}^T  \max_{i \in S_t}  \left\|x_{ti}  \right\|_{H_t^{-1}}^2
        \\
        &\leq  \frac{4}{\kappa} \beta_{T+1}(\delta) d \log \left( 1 + \frac{T}{d\lambda} \right)
        , \numberthis \label{eq:non_unform_bouns_fist_decomposition_1}
    \end{align*}
    where the last inequality holds by Lemma~\ref{lemma:elliptical}.
    Using similar reasoning, we can bound the second term of Equation~\eqref{eq:non_unform_bouns_fist_decomposition}.
    By the mean value theorem, there exists $\xib'_t = (1-c') \wb^\star + c' \wb_{t+1}$ for some $c' \in (0,1)$ such that
    \begin{align*}
        &\sum_{i \in S_t} \left(  p_t(i | S_t, \wb^\star) - p_t(i | S_t, \wb_{t+1}) \right)\| \tilde{x}_{ti} \|_{H_t^{-1}}
        = \sum_{i \in S_t}  \nabla p_t(i | S_t, \xib'_t)^\top (\wb^\star - \wb_{t+1})  \| \tilde{x}_{ti} \|_{H_t^{-1}}
        \\
        &=  \sum_{i \in S_t} \left( p_t(i|S_t, \xib'_t) x_{ti} - p_t(i|S_t, \xib'_t) \sum_{k \in S_t} p_t(k|S_t, \xib'_t) x_{tk}\right)^\top  (\wb^\star - \wb_{t+1})  \| \tilde{x}_{ti} \|_{H_t^{-1}}
        \\
        &\leq \beta_{t+1}(\delta) \sum_{i \in S_t}  p_t(i|S_t, \xib'_t) \| x_{ti} \|_{H_t^{-1}}  \| \tilde{x}_{ti} \|_{H_t^{-1}}
        \\
        &+  \beta_{t+1}(\delta) \sum_{i \in S_t} p_t(i|S_t, \xib'_t)   \| \tilde{x}_{ti} \|_{H_t^{-1}}  \sum_{k \in S_t} p_t(k|S_t, \xib'_t) \| x_{tk}\|_{H_t^{-1}}
        \\
        &\leq  \beta_{t+1}(\delta) \max_{i \in S_t} \| x_{ti} \|_{H_t^{-1}}  \| \tilde{x}_{ti} \|_{H_t^{-1}}
        +  \beta_{t+1}(\delta) \max_{i \in S_t}  \| \tilde{x}_{ti} \|_{H_t^{-1}}  \max_{k \in S_t} \| x_{tk}\|_{H_t^{-1}}
        .
    \end{align*}
    Then, by applying the AM-GM inequality to each term, we obtain 
    \begin{align*}
        &\beta_{t+1}(\delta) \max_{i \in S_t} \| x_{ti} \|_{H_t^{-1}}  \| \tilde{x}_{ti} \|_{H_t^{-1}}
        +  \beta_{t+1}(\delta) \max_{i \in S_t}  \| \tilde{x}_{ti} \|_{H_t^{-1}}  \max_{k \in S_t} \| x_{tk}\|_{H_t^{-1}}
        \\
        &\leq  \beta_{t+1}(\delta) \max_{i \in S_t} \frac{\| x_{ti} \|_{H_t^{-1}}^2 +  \| \tilde{x}_{ti} \|_{H_t^{-1}}^2}{2}
        +  \beta_{t+1}(\delta) \frac{ \left(\max_{i \in S_t}  \| \tilde{x}_{ti} \|_{H_t^{-1}} \right)^2 + \left(  \max_{k \in S_t} \| x_{tk}\|_{H_t^{-1}} \right)^2 }{2}
        \\
        &=  \beta_{t+1}(\delta) \max_{i \in S_t} \frac{\| x_{ti} \|_{H_t^{-1}}^2 +  \| \tilde{x}_{ti} \|_{H_t^{-1}}^2}{2}
        +  \beta_{t+1}(\delta) \frac{ \max_{i \in S_t}  \| \tilde{x}_{ti} \|_{H_t^{-1}}^2 +  \max_{k \in S_t} \| x_{tk}\|_{H_t^{-1}}^2 }{2}
        \\
        &\leq  2\beta_{t+1}(\delta) \max \left\{ \max_{i \in S_t}\| x_{ti} \|_{H_t^{-1}}^2,  \max_{i \in S_t}  \| \tilde{x}_{ti} \|_{H_t^{-1}}^2 \right\}
    \end{align*}
    where the equality holds since $(\max_i a_i)^2 = \max_i a_i^2$ for any $a_i \geq 0$.
    Thus, by Lemma~\ref{lemma:elliptical_x_tilde} (or Lemma~\ref{lemma:elliptical}), we get
    \begin{align*}
        \sum_{t=1}^T & \sum_{i \in S_t} \left(  p_t(i | S_t, \wb^\star) - p_t(i | S_t, \wb_{t+1}) \right)\| \tilde{x}_{ti} \|_{H_t^{-1}}
        \\
        &\leq  2\beta_{t+1}(\delta) \max \left\{ \max_{i \in S_t}\| x_{ti} \|_{H_t^{-1}}^2,  \max_{i \in S_t}  \| \tilde{x}_{ti} \|_{H_t^{-1}}^2 \right\}
        \leq  \frac{4}{\kappa} \beta_{T+1}(\delta) d \log \left( 1 + \frac{T}{d\lambda} \right),
        , \numberthis \label{eq:non_unform_bouns_fist_decomposition_2}
    \end{align*}
    Finally, we bound the third term of Equation~\eqref{eq:non_unform_bouns_fist_decomposition}.
    By the Cauchy-Schwarz inequality, we have
    \begin{align*}
        \sum_{t=1}^T \sum_{i \in S_t} p_t(i | S_t, \wb_{t+1}) \| \tilde{x}_{ti} \|_{H_t^{-1}} 
        &\leq \sqrt{\sum_{t=1}^T \sum_{i \in S_t} p_t(i | S_t, \wb_{t+1})} \sqrt{\sum_{t=1}^T \sum_{i \in S_t} p_t(i | S_t, \wb_{t+1}) \| \tilde{x}_{ti} \|_{H_t^{-1}}^2 }
        \\
        &\leq \sqrt{T} \sqrt{2d \log\left( 1 + \frac{T}{d\lambda} \right)}
        , \numberthis \label{eq:non_unform_bouns_fist_decomposition_3}
    \end{align*}
    where the last inequality holds by Lemma~\ref{lemma:elliptical_x_tilde}.
    Plugging Equation~\eqref{eq:non_unform_bouns_fist_decomposition_1},~\eqref{eq:non_unform_bouns_fist_decomposition_2}, and~\eqref{eq:non_unform_bouns_fist_decomposition_3} into Equation~\eqref{eq:non_unform_bouns_fist_decomposition}, we get
    \begin{align*}
        \sum_{t=1}^T \sum_{i \in S_t} &p_t(i | S_t, \wb^\star)    \| x_{ti} - \EE_{j \sim p_t(\cdot | S_t, \wb^\star)} \left[ x_{tj} \right]  \|_{H_t^{-1}}   
        \\
        &\leq \sqrt{T} \sqrt{2d \log\left( 1 + \frac{T}{d\lambda} \right)}
        +  \frac{8}{\kappa} \beta_{T+1}(\delta) d \log \left( 1 + \frac{T}{d\lambda} \right)
    \end{align*}
    Thus, we can bound the term~\texttt{(c)} as follows: 
    \begin{align*}
         \sum_{t=1}^T \nabla \tilde{Q}(\ub_t^\star)^\top (\ub_t - \ub^\star_t) 
         \leq 2 \beta_{T}(\delta) \sqrt{T} \sqrt{2d \log\left( 1 + \frac{T}{d\lambda} \right)}
        +  \frac{16}{\kappa} \beta_{T}(\delta) \beta_{T+1}(\delta) d \log \left( 1 + \frac{T}{d\lambda} \right)
        , \numberthis \label{eq:non_uniform_first}
    \end{align*}
    Now, we bound the term \texttt{(D)}.
    Define $Q:\RR^{|S_t|} \rightarrow \RR$, such that for all $\ub = (u_1, \dots, u_{|S_t|}) \in \RR^{|S_t|}$, $Q(\ub) = \sum_{i=1}^{|S_t|} \frac{\exp(u_i) }{v_0 + \sum_{j=1}^{|S_t|} \exp(u_j)}$.
    Then, we have $\left| \frac{\partial^2 \tilde{Q}}{\partial i \partial j} \right| 
    \leq  \left| \frac{\partial^2 Q}{\partial i \partial j} \right| $ since $r_{ti} \in [0,1]$.
    By following the similar reasoning from Equation~\eqref{eq:term_B} to Equation~\eqref{eq:term_B_last} in Section~\ref{app_subsec:main_proof_thm_upper}, we have
    \begin{align*}
        \frac{1}{2} \sum_{t=1}^T (\ub_t - \ub^\star_t)^\top \nabla^2 \tilde{Q}(\bar{\ub}_t) (\ub_t - \ub^\star_t)
        &= \frac{1}{2} \sum_{t=1}^T \sum_{i \in S_t} \sum_{j \in S_t}(u_{ti} - u_{ti}^\star ) \frac{\partial^2 \tilde{Q}}{\partial i \partial j} (u_{tj} - u_{tj}^\star )
        \\
        &\leq \frac{1}{2} \sum_{t=1}^T \sum_{i \in S_t} \sum_{j \in S_t} |u_{ti} - u_{ti}^\star | \left| \frac{\partial^2 Q}{\partial i \partial j} \right| |u_{tj} - u_{tj}^\star |
        \\
        &\leq  10 \beta_T(\delta)^2  \sum_{t=1}^T \max_{i \in S_t }  \| x_{ti} \|_{H_t^{-1}}^2 
        . \numberthis \label{eq:non_uniform_second}
    \end{align*}
    where the first inequality holds because  $\left| \frac{\partial^2 \tilde{Q}}{\partial i \partial j} \right| 
    \leq  \left| \frac{\partial^2 Q}{\partial i \partial j} \right| $.
    Combining Equation~\eqref{eq:non_uniform_first} and ~\eqref{eq:non_uniform_second}, we derive that  
    \begin{align*}
         \Regret(\wb^\star)
         &\leq 
         2 \beta_{T}(\delta) \sqrt{T} \sqrt{2d \log\left( 1 + \frac{T}{d\lambda} \right)}
        +  \frac{16}{\kappa} \beta_{T}(\delta) \beta_{T+1}(\delta) d \log \left( 1 + \frac{T}{d\lambda} \right)
         \\
         &+ 10 \beta_T(\delta)^2  \sum_{t=1}^T \max_{i \in S_t }  \| x_{ti} \|_{H_t^{-1}}^2
         \\
         &= \BigOTilde \left( d\sqrt{T} + \frac{1}{\kappa}d^2 \right)
         ,
    \end{align*}
    where $\beta_{T}(\delta) =\BigO \left( \sqrt{d} \log T \log K \right)$.
    This concludes the proof of Theorem~\ref{theorem:upper_bound_non_uniform}.
\end{proof}

\begin{remark} \label{remark:B-bound_upper_non-uniform}
    If the boundedness assumption on the parameter is relaxed to
    $\| \wb \|_2 \leq B$,
    since
    $\beta_t(\delta)
    = \BigO \left( B \sqrt{d} \log t \log K  
        + B^{3/2} \sqrt{d \log K }
        \right)$ (refer Corollary~\ref{corollary:B-bound_confidence}),
    we have 
    $\Regret(\wb^\star)
    = \BigOTilde \left ( B^{3/2} d\sqrt{T}  + \frac{1}{\kappa} B^3 d^2 \right).
    $
\end{remark}

\subsection{Proofs of Lemmas for Theorem~\ref{theorem:upper_bound_non_uniform}} 
\label{app_subsec:useful_lemmas_cor1}

\subsubsection{Proof of Lemma~\ref{lemma:increasing_R}}
\label{app_subsubsec:proof_of_lemma:increasing_R}
\begin{proof}[Proof of Lemma~\ref{lemma:increasing_R}]
    We prove the result by first showing that for any $i \in S_t$, we have $r_{ti} \geq \tilde{R}_t(S_t)$.
    Suppose that there exists $i \in S_t$ for which $r_{ti} < \tilde{R}_t(S_t)$.
    Removing item $i$ from the assortment $ S_t$ results in a higher expected revenue. 
    Consequently, $S_t \neq \argmax_{S \in \Scal} \tilde{R}_t(S)$, which contradicts the optimality of $S_t$.
    Thus, we have
    \begin{align*}
        r_{ti} \geq \tilde{R}_t(S_t), \quad \forall i \in S_t.
    \end{align*}
    If we increase $\alpha_{ti}$ to $\alpha_{ti}'$ for all $i \in S_t$, the probability of selecting the outside option decreases. 
    In other words, the sum of probabilities of choosing any $i \in S_t$ increases.
    Since $r_{ti} \geq \tilde{R}_t(S_t)$ for all $i \in S_t$, this results in an increase in revenue.
    Hence, we get
    \begin{align*}
        \tilde{R}_{t}(S_t) 
        = \frac{\sum_{i \in S_t} \exp( \alpha_{ti} ) r_{ti} }{v_0 + \sum_{j \in S_t} \exp(\alpha_{tj}) }
        \leq \frac{\sum_{i \in S_t} \exp( \alpha'_{ti} ) r_{ti} }{v_0 + \sum_{j \in S_t} \exp(\alpha'_{tj}) }.
    \end{align*}
    This concludes the proof.
\end{proof}

\subsubsection{Proof of Lemma~\ref{lemma:elliptical_x_tilde}}
\label{app_subsubsec:proof_of_lemma:elliptical_x_tilde}
\begin{proof}[Proof of Lemma~\ref{lemma:elliptical_x_tilde}]
    For notational simplicity, let  $\EE_{\wb}[x_{tj}] = \EE_{j \sim p_t(\cdot | S_t, \wb)}[x_{tj}]$.
    Let $x_{t0} = \mathbf{0}$.
    We can rewrite $\Gcal_s(\wb)$ in the following way:
    \begin{align*}
        &\Gcal_s(\wb_{s+1}) 
        \\
        &= \sum_{i \in S_s} p_s(i | S_s, \wb_{s+1}) x_{si} x_{si}^\top -  \sum_{i \in S_s}  \sum_{j \in S_s} p_s(i | S_s, \wb_{s+1}) p_s(j | S_s, \wb_{s+1}) x_{si} x_{sj}^\top \numberthis \label{eq:hessian_for_elliptical}
        \\
        &= \sum_{i \in S_s \cup \{0\}} p_s(i | S_s, \wb_{s+1}) x_{si} x_{si}^\top -  \sum_{i \in S_s \cup \{0\}}  \sum_{j \in S_s \cup \{0\}} p_s(i | S_s, \wb_{s+1}) p_s(j | S_s, \wb_{s+1}) x_{si} x_{sj}^\top
        \\
        &= \EE_{\wb_{s+1}}[x_{si}x_{si}^\top] 
        - \EE_{\wb_{s+1}}[x_{si}] \left(\EE_{\wb_{s+1}}[x_{si}] \right)^\top
        \\
        &= \EE_{\wb_{s+1}}\left[(x_{si} - \EE_{\wb_{s+1}}[x_{sm}]) (x_{si} - \EE_{\wb_{s+1}}[x_{sm}])^\top\right] 
        \\
        &= \EE_{\wb_{s+1}} [\tilde{x}_{si}\tilde{x}_{si}^\top]
        = \sum_{i \in S_s \cup \{ 0\} } p_s(i | S_s, \wb_{s+1}) \tilde{x}_{si}\tilde{x}_{si}^\top 
        \succeq \sum_{i \in S_s  } p_s(i | S_s, \wb_{s+1}) \tilde{x}_{si}\tilde{x}_{si}^\top
        .        
    \end{align*}
    This means that
    \begin{align*}
        H_{t+1} = H_t + \Gcal_t(\wb_{t+1}) 
        \succeq H_t +  \sum_{i \in S_t} p_t(i | S_t, \wb_{t+1}) \tilde{x}_{ti}\tilde{x}_{ti}^\top.
        \numberthis \label{eq:hessian_for_elliptical_p}
    \end{align*}
    Hence, we can derive that
    \begin{align*}
        \det \left( H_{t+1} \right)
        \geq \det \left( H_t \right) \left( 1 +  \sum_{i \in S_t}  p_t(i | S_t, \wb_{t+1}) \| \tilde{x}_{ti} \|_{H_{t}^{-1}}^2 \right).
    \end{align*}
    Since $\lambda \geq 1$, for all $t \geq 1$ we have $\sum_{i \in S_t}  p_t(i | S_t, \wb_{t+1}) \| \tilde{x}_{ti} \|_{H_{t}^{-1}}^2 \leq  \frac{1}{\lambda} \max_{i \in S_t} \| \tilde{x}_{ti} \|_2 \leq 1$.
    Then, using the fact that $z \leq 2 \log ( 1 + z)$ for any $z \in [0,1]$, we get
    \begin{align*}
        \sum_{s=1}^{t} \sum_{i \in S_s} p_s(i | S_s, \wb_{s+1}) \| \tilde{x}_{si}\|_{H_s^{-1}}^2 
        &\leq 2 \sum_{s=1}^{t} \log \left( 1 + p_s(i | S_s, \wb_{s+1}) \| \tilde{x}_{si}\|_{H_s^{-1}}^2  \right)
        \\
        &\leq 2 \sum_{s=1}^{t} \log \left( \frac{\det (H_{s+1})}{\det(H_s)} \right)
        \\
        &= 2 \log \left( \frac{\det (H_{t+1})}{\det(H_1)} \right)
        \\
        &\leq 2 d \log  \left( \frac{\tr(H_{t+1}) }{d \lambda}  \right)
        \leq 2d \log \left(1 + \frac{t}{d \lambda} \right).
    \end{align*}
    This proves the first inequality.

    To show the second inequality, we come back to Equation~\eqref{eq:hessian_for_elliptical_p}.
    By the definition of $\kappa$, we get
    \begin{align*}
        H_{t+1}
        &= H_t + \Gcal_t(\wb_{t+1}) 
        = H_t +  \sum_{i \in S_t} p_t(i | S_t, \wb_{t+1}) \tilde{x}_{ti}\tilde{x}_{ti}^\top 
        \\
        &\succeq H_t + \kappa \sum_{i \in S_t}  \tilde{x}_{ti}\tilde{x}_{ti}^\top
        . 
    \end{align*}
    Thus, we obtain that
    \begin{align*}
        \det \left( H_{t+1} \right)
        \geq \det \left( H_t \right) \left( 1 +  \kappa \sum_{i \in S_t}  \| \tilde{x}_{ti} \|_{H_{t}^{-1}}^2 \right)
        \geq \det \left( H_t \right) \left( 1 +  \kappa \max_{i \in S_t}  \| \tilde{x}_{ti} \|_{H_{t}^{-1}}^2 \right).
    \end{align*}
    Since $\lambda \geq 1$, for all $t \geq 1$ we have $\kappa \max_{i \in S_t}  \| \tilde{x}_{ti} \|_{H_{t}^{-1}}^2 \leq  \frac{\kappa}{\lambda} \| \tilde{x}_{ti} \|_2 \leq \kappa$.
    We then reach the conclusion in the same manner:
    \begin{align*}
        \sum_{s=1}^{t} \max_{i \in S_s} \| \tilde{x}_{si}\|_{H_s^{-1}}^2
        &\leq \frac{2}{\kappa} \sum_{s=1}^{t} \log \left( 1 + \kappa\max_{i \in S_s} \| \tilde{x}_{si}\|_{H_s^{-1}}^2 \right)
        \\
        &\leq  \frac{2}{\kappa} \sum_{s=1}^{t} \log \left( \frac{\det (H_{s+1})}{\det(H_s)} \right)
        \\
        &= \frac{2}{\kappa} \log \left(\frac{\det (H_{t+1})}{\det(H_1)} \right)
        \\
        &\leq \frac{2}{\kappa} d \log  \left( \frac{\tr(H_{t+1}) }{d \lambda}  \right)
        \leq \frac{2}{\kappa} d \log \left( 1+ \frac{t}{d \lambda} \right).
    \end{align*}
    This proves the second inequality.
\end{proof}
\section{Proof of Proposition~\ref{prop:lower_instance}}
\label{app_sec:proof_lower_instance}
\subsection{Main Proof of Proposition~\ref{prop:lower_instance}}
\label{app_subsec:proof_lower_instance_main}
\begin{proof} [Proof of Proposition~\ref{prop:lower_instance}]
    We construct the parameters and features in the same way as in Section~\ref{app_subsec:lower_adversarial_setting}.
    Recall that, by Proposition~\ref{prop:lower_S_tilde}, it is sufficient to bound $\sum_{i \in S^\star} p(i | S^\star, \wb_V) - \sum_{i \in \tilde{S}_t} p(i | \tilde{S}_t, \wb_V) $.
    We denote $\tilde{U}_t$ as the unique $U^\star \in \mathcal{V}_{d/4}$ in $\tilde{S}_t$.
    We define a function $\mu: \RR \rightarrow [0,1]$ such that for any $z \in \RR$, $\mu(z) = \frac{K \exp ( z )}{ v_0 + K \exp(z)} $.
    Since all items in $S^\star$ (and in $\tilde{S}_t$) are identical, we can express $\sum_{i \in S^\star} p(i | S^\star, \wb_V)$ (and $\sum_{i \in \tilde{S}_t} p(i | \tilde{S}_t, \wb_V) $) as follows:
    \begin{align*}
        \sum_{i \in S^\star} p(i | S^\star, \wb_V)
        &= \frac{K \exp (x_V^\top \wb_V)}{v_0 + K \exp (x_V^\top \wb_V)}
        = \mu( x_V^\top \wb_V ),
        \\
        \sum_{i \in \tilde{S}_t} p(i | \tilde{S}_t, \wb_V) 
        &= \frac{K \exp (x_{\tilde{U}_t}^\top \wb_V)}{v_0 + K \exp (x_{\tilde{U}_t}^\top \wb_V)}
        = \mu( x_{\tilde{U}_t}^\top \wb_V ).
    \end{align*}
    Given $\wb_V$, we define $\kappa^\star_t(\wb_V) := \sum_{i \in S^\star} p_t(i | S^\star, \wb_V) p_t(0 | S^\star, \wb_V)$.
    Since the context vectors $\{x_{ti} \}$ are constructed to be invariant across rounds $t$, $\kappa^\star_t(\wb_V)$ is also independent of $t$, i.e., $\kappa^\star_1(\wb_V) = \dots = \kappa^\star_T(\wb_V)$.
    Therefore, we omit the index $t$ for simplicity.
    Note that, in our instance construction, $\kappa^\star(\wb_V) = \mu ( x_V^\top \wb_V )\left(1-\mu ( x_V^\top \wb_V ) \right) = \dot{\mu} ( x_V^\top \wb_V ) $.

    We hypothesize that for all $V \in \mathcal{V}_{d/4}$, the regret is dominated by $d \sqrt{\kappa^\star(\wb_V) T}$. 
    If this assumption does not hold, then by definition, there exists some $V \in \mathcal{V}_{d/4}$ such that $\EE_{V} \left[
        \Regret (\wb_V)
        \right] = \Omega \left( d \sqrt{\kappa^\star(\wb_V) T} \right)$, thereby completing the proof.

    \textbf{Hypothesis.} There exists a constant $C>0$ such that:
    \begin{align*}
        \EE_{V} \left[
        \Regret (\wb_V)
        \right]
        \leq C \cdot d \sqrt{\kappa^\star(\wb_V) T}
        ,
        \quad \,
        \forall V \in \mathcal{V}_{d/4}.
        \numberthis \label{eq:hypothesis}
    \end{align*}

    Additionally, we set $\wb^\star = \argmax_{\wb_V} \kappa^\star(\wb_V) $, thus, we have $\kappa^\star = \kappa^\star (\wb^\star) = \max_{\wb_V} \kappa^\star(\wb_V)$.
    
    To establish an instance-dependent lower bound for $ \mu( x_V^\top \wb_V ) -  \mu( x_{\tilde{U}_t}^\top \wb_V )$, we use the following lemma in place of Lemma~\ref{lemma:lower_revenue_gap}:
    \begin{lemma} \label{lemma:instance_lower_revenue_gap}
        Suppose $\epsilon \in (0, 1/d\sqrt{d})$ and define $\delta := d/4 - | \tilde{U}_t \cap V |$.
        Then, we have
        \begin{align*}
            \mu( x_V^\top \wb_V ) -  \mu( x_{\tilde{U}_t}^\top \wb_V )
            \geq \frac{\kappa^\star(\wb_V) }{3} \cdot \frac{\delta \epsilon}{\sqrt{d}}
            .
        \end{align*}
    \end{lemma}
    For any $j \in V$, define random variables $\tilde{M}_j := \sum_{t=1}^T \mathbf{1} \{ j \in \tilde{U}_t \}$.
    Then, by Lemma~\ref{lemma:instance_lower_revenue_gap}, for all $V \in \mathcal{V}_{d/4}$, we get
    \begin{align*}
        \EE_{V} 
        \sum_{t=1}^T
        \left[
        \sum_{i \in S^\star}
        p(i | S^\star, \wb_V) - \sum_{i \in \tilde{S}_t} p(i | \tilde{S}_t, \wb_V)
        \right]
        \geq \frac{\kappa^\star(\wb_V) }{3} \cdot \frac{ \epsilon}{\sqrt{d}} 
        \left( \frac{dT}{4} - \sum_{j \in V} \EE_{V}[\tilde{M}_j] \right).
        \numberthis \label{eq:instance_lower_expected_revenue_gap}
    \end{align*}
    Additionally, we define $\mathcal{V}_{d/4}^{(j)} := \{ V \in \mathcal{V}_{d/4} : j \in V \}$ and $\mathcal{V}_{d/4 -1}:= \{ V \subseteq [d] : |V| = d/4 -1 \}$.
    By averaging both sides of Equation~\eqref{eq:instance_lower_expected_revenue_gap} with respect to all $V \in \mathcal{V}_{d/4}$, and by following reasoning similar to that in the proof of Theorem~\ref{thm:lower_bound}, we get
    \begin{align*}
        \frac{1}{\left|\mathcal{V}_{d/4} \right|} \sum_{V \in \mathcal{V}_{d/4}} &\EE_{V} 
        \sum_{t=1}^T
        \left[
        \sum_{i \in S^\star} p(i | S^\star, \wb_V) - \sum_{i \in \tilde{S}_t} p(i | \tilde{S}_t, \wb_V)
        \right]
        \\
        &\geq \frac{\kappa^\star(\wb_V) }{3} \cdot \frac{ \epsilon}{\sqrt{d}}  
        \left( \frac{dT}{6} - \max_{V \in \mathcal{V}_{d/4 -1}} \sum_{j \notin V} 
        \left| \EE_{V \cup \{j\}}[ \tilde{M}_j]
        - \EE_{V} [ \tilde{M}_j] \right| \right).
        \numberthis \label{eq:instance_lower_expected_revenue_gap_before_pinsker}
    \end{align*}
    For simplicity, let $P = \mathbb{P}_V$ and $Q = \mathbb{P}_{V \cup \{j\}}$.
    Then, we can bound the term $\left| \EE_{V \cup \{j\}}[ \tilde{M}_j] - \EE_{V} [ \tilde{M}_j] \right|$ in Equation~\eqref{eq:instance_lower_expected_revenue_gap_before_pinsker} for any $V \in \mathcal{V}_{d/4 -1}$.
    \begin{align*}
        \left| \EE_{P}[ \tilde{M}_j] - \EE_{Q} [ \tilde{M}_j\right] |
        &\leq \sum_{t=0}^T t \cdot \left| P[\tilde{M}_j =t ] - Q [\tilde{M}_j = t] \right|
        \\
        &\leq T \cdot \| P-Q \|_{\operatorname{TV}} 
        \leq T \cdot \sqrt{\frac{1}{2} \operatorname{KL}(P \| Q)},
        \numberthis \label{eq:instance_lower_pinskers}
    \end{align*}
    where $\| P-Q \|_{\operatorname{TV}} = \sup_{A} |P(A) - Q(A) |$ | is the total variation distance between $P$ and $Q$, $\operatorname{KL}(P \| Q) = \int (\log \dd P / \dd Q)\dd P$ is s the Kullback-Leibler (KL) divergence between $P$ and $Q$, and the last inequality holds by Pinsker's inequality.
    We can derive the instance-dependent bound for
    the KL divergence term in Equation~\eqref{eq:instance_lower_pinskers} by the following lemma:
    \begin{lemma} \label{lemma:instance_lower_KL}
        For any $V \in \mathcal{V}_{d/4 - 1}$ and $j \in [d]$, there exists a positive constant $C_{\operatorname{KL}} >0$ such that
        \begin{align*}
            \sum_{j=1}^d \operatorname{KL}(P_{V} \| Q_{V \cup \{j\}}) 
            \leq  C_{\operatorname{KL}} \cdot
        \epsilon^2
        \kappa^\star(\wb_V) T.
        \end{align*}
    \end{lemma}
    Plugging Equation~\eqref{eq:instance_lower_pinskers} into Equation~\eqref{eq:instance_lower_expected_revenue_gap_before_pinsker}, we get
    \begin{align*}
        \frac{1}{\left|\mathcal{V}_{d/4} \right|} \sum_{V \in \mathcal{V}_{d/4}} &\EE_{V} 
         \sum_{t=1}^T
        \left[
        \sum_{i \in S^\star} p(i | S^\star, \wb_V) - \sum_{i \in \tilde{S}_t} p(i | \tilde{S}_t, \wb_V)
        \right]
        \\
        &\geq
        \frac{\kappa^\star(\wb_V) }{3} \cdot \frac{ \epsilon}{\sqrt{d}}  
        \left( \frac{dT}{6} - 
        T \sum_{j=1}^d \sqrt{\frac{1}{2} \operatorname{KL}(P \| Q)}
        \right)
        \\
        &\geq 
        \frac{\kappa^\star(\wb_V) }{3} \cdot \frac{ \epsilon}{\sqrt{d}}  
        \left( \frac{dT}{6} - 
        T \sqrt{d}
        \cdot \sqrt{\frac{1}{2}  \sum_{j=1}^d  \operatorname{KL}(P \| Q)}
        \right)
        \\
        &\geq 
        \frac{\kappa^\star(\wb_V) }{3} \cdot \frac{ \epsilon}{\sqrt{d}}  
        \left( \frac{dT}{6} - 
        T \sqrt{d}
        \cdot \sqrt{\frac{1}{2}  
         C_{\operatorname{KL}} \cdot
        \epsilon^2
        \kappa^\star(\wb_V) T
        }
        \right)
        ,
    \end{align*}
    where the second inequality is due to the Cauchy-Schwartz inequality,
    and the last inequality is by Lemma~\ref{lemma:instance_lower_KL}.
    
    By setting $\epsilon = \sqrt{\frac{d}{72 C_{\operatorname{KL}} \cdot  \kappa^\star(\wb_V) T } }$, we have
    \begin{align*}
        \frac{1}{\left|\mathcal{V}_{d/4} \right|} \sum_{V \in \mathcal{V}_{d/4}} &\EE_{V}
        \sum_{t=1}^T
        \left[
        \sum_{i \in S^\star} p(i | S^\star, \wb_V) - \sum_{i \in \tilde{S}_t} p(i | \tilde{S}_t, \wb_V)\right]
        \\
        &\geq
        \frac{\kappa^\star(\wb_V) }{3} \cdot \frac{ 1}{\sqrt{d}}  
        \cdot
         \sqrt{\frac{d}{72 C_{\operatorname{KL}} \cdot  \kappa^\star(\wb_V) T   } }
        \cdot
         \frac{dT}{12} 
         \\
         &= \Omega \left(
            d \sqrt{ \kappa^\star(\wb_V) T }
         \right).
    \end{align*}
    Recall that by construction, 
    $\kappa^\star = \max_{\wb_V} \kappa^\star(\wb_V)$.
    Thus, by taking the maximum over $\wb_V$, we complete the proof of Proposition~\ref{prop:lower_instance}.
\end{proof}
%
\subsection{Proofs of Lemmas for Proposition~\ref{prop:lower_instance}}
\subsubsection{Proof of Lemma~\ref{lemma:instance_lower_revenue_gap}}
\begin{proof} [Proof of Lemma~\ref{lemma:instance_lower_revenue_gap}]
    For simplicity, let $x = x_{V}$ and $\hat{x} = x_{\tilde{U}_t}$.
    Then, by the mean value theorem,  we have
    \begin{align*}
        \mu( x^\top \wb_V ) -  \mu( \hat{x}^\top \wb_V )
        &= \int_{v=0}^1 \dot{\mu} \left( x^\top \wb_V + v (\hat{x} - x)^\top \wb_V  \right)
        \left( x - \hat{x} \right)^\top \wb_V
        \\
        &\geq \frac{\dot{\mu} (x^\top \wb_V) }{ 1 + \left| (x - \hat{x})^\top \wb_V  \right| }
        \left( x - \hat{x} \right)^\top \wb_V
        \\
        &\geq \frac{\dot{\mu} (x^\top \wb_V) }{ 3 }
        \left( x - \hat{x} \right)^\top \wb_V
        \\
        &= \frac{\kappa^\star(\wb_V) }{3} \left( x - \hat{x} \right)^\top \wb_V
        \\
        &\geq \frac{\kappa^\star(\wb_V) }{3} \cdot \frac{\delta \epsilon}{\sqrt{d}}
        ,
    \end{align*}
    where 
    the first inequality holds by Lemma~\ref{lemma:lemma7_abelille},
    the second inequality holds by the bounded assumption (Assumption~\ref{assum:bounded_assumption}),
    and the last inequality holds by the definition of $\delta = d/4 - | \tilde{U}_t \cap V |$.
\end{proof}
\subsubsection{Proof of Lemma~\ref{lemma:instance_lower_KL}}
\begin{proof} [Proof of Lemma~\ref{lemma:instance_lower_KL}]
    Consider a fixed round $t$, an assortment $\tilde{S}_t$, and the set $\tilde{U}_t$.
    Let $U = \tilde{U}_t$.
    Define $m_j(\tilde{S}_t) := \sum_{x_U \in \tilde{S}_t}\mathbf{1} \{j \in U \} /K$, which  captures the average presence of $j \in U$ across the assortment $\tilde{S}_t$.
    For simplicity, we denote $p =   \frac{  K \exp(x_{U}^\top \wb_{V} )}{v_0 + K \exp(x_{U}^\top \wb_{V } )}$ and $q =  \frac{  K \exp(x_{U}^\top \wb_{V \cup \{j\} } )}{v_0 + K \exp(x_{U}^\top \wb_{V \cup \{j\} } )}$.
    Then, we get
    \begin{align*}
        \operatorname{KL}\left( \mathbb{P}_{V} (\cdot | \tilde{S}_t ) \| \mathbb{P}_{V \cup \{j\}} (\cdot |  \tilde{S}_t) \right)
        &\leq \chi^2\! \left(\mathbb{P}_{V} (\cdot | \tilde{S}_t ) \| \mathbb{P}_{V \cup \{j\}} (\cdot |  \tilde{S}_t) \right)
        \\
        &=  \chi^2\! \left(
        \operatorname{Bernoulli} \left( p\right) 
        \| \operatorname{Bernoulli} \left( q\right) 
        \right)
        \\
        &= \frac{(p-q)^2}{q} + \frac{ (p-q)^2 }{1-q}
        \\
        &= \frac{(p-q)^2}{q(1-q)}
        \\
        &= \frac{ \left( \mu (x_U^\top \wb_V) - \mu (x_U^\top \wb_{V \cup \{j\}} ) \right)^2 }{\dot{\mu} (x_U^\top \wb_{V \cup \{j\}} )}
        \\
        &= \frac{ \left(\dot{\mu} (x_U^\top \bar{\wb}_{V,j} )\right)^2 }{\dot{\mu} (x_U^\top \wb_{V \cup \{j\}} )} 
        \left( x_U^\top \left( \wb_V - \wb_{V \cup \{j\} } \right) \right)^2
        ,
    \end{align*}
    where in the first inequality, we used $\operatorname{KL} \leq \chi^2$ (\citet{tsybakov2008nonparametric}, Chapter 2), where $\chi^2$ is a chi-square divergence,
    the second equality holds by the expression of the chi-square divergence for Bernoulli random variables,
    and the last equality holds by the mean value theorem, where $\bar{\wb}_{V,j}$ is a convex combination of $\wb_V$ and $\wb_{V \cup \{j\} }$.
    Then, by Lemma~\ref{lemma:lemma9_abelille}, we can further bound the last term as follows:
    \begin{align*}
         \operatorname{KL}\left( \mathbb{P}_{V} (\cdot | \tilde{S}_t ) \| \mathbb{P}_{V \cup \{j\}} (\cdot |  \tilde{S}_t) \right)
        &\leq \dot{\mu}  (x_U^\top \wb_{V} ) e^{3\left| x_U^\top \left( \wb_V -  \wb_{V \cup \{j\} } \right) \right| } 
         \left( x_U^\top \left( \wb_V - \wb_{V \cup \{j\} } \right) \right)^2
         \\
         &\leq  \dot{\mu}  (x_U^\top \wb_{V} )  e^{3/d^2} 
         \left( x_U^\top \left( \wb_V - \wb_{V \cup \{j\} } \right) \right)^2
         \\
         &\leq  \dot{\mu}  (x_U^\top \wb_{V} )  e^{1/4}
         \left( x_U^\top \left( \wb_V - \wb_{V \cup \{j\} } \right) \right)^2
         \\
         &\leq  \dot{\mu}  (x_U^\top \wb_{V} )  e^{1/4}
         \cdot \frac{m_j(\tilde{S}_t) \epsilon^2}{d}
         ,
    \end{align*}
    where the second-to-the last inequality holds based on the assumption that $d \geq 4$,
    and the last inequality holds since $m_j(\tilde{S}_t)\leq 1$.

    Now, let us consider $t$ varying over the round $t \in [T]$. Then, we obtain 
    \begin{align*}
        \sum_{j=1}^d  \operatorname{KL}(P_{V} \| Q_{V \cup \{j\}}) 
        &= \sum_{j=1}^d \sum_{t=1}^T \EE_{V} \left[ \operatorname{KL}\left( \mathbb{P}_{V} (\cdot | \tilde{S}_t ) \| \mathbb{P}_{V \cup \{j\}} (\cdot |  \tilde{S}_t) \right) \right]
        \\
        &\leq e^{1/4} \cdot
        \frac{\epsilon^2}{d}
        \sum_{j=1}^d   \sum_{t=1}^T
        \EE_{V} \left[
        \dot{\mu}  (x_{\tilde{U}_t}^\top \wb_{V} )  
         \cdot m_j(\tilde{S}_t)
         \right]
        \\
        &= e^{1/4} \cdot
        \frac{\epsilon^2}{d}
         \EE_{V} \left[ 
         \sum_{t=1}^T \left(
        \dot{\mu}  (x_{\tilde{U}_t}^\top \wb_{V} )  
         \sum_{j=1}^d m_j(\tilde{S}_t) \right)  \right]
        \\
        &\leq  e^{1/4} \cdot
        \frac{\epsilon^2}{4}
        \EE_{V} \left[
         \sum_{t=1}^T 
        \dot{\mu}  (x_{\tilde{U}_t}^\top \wb_{V} )  
        \right]
        \\
        &\leq e^{1/4} \cdot
        \frac{\epsilon^2}{4}
        \left( 
        \dot{\mu}  (x_{V}^\top \wb_{V} ) T 
        + 
        \EE_{V} \left[
        \Regret (\wb_V)
        \right]
        \right)
        \\
        &= e^{1/4} \cdot
        \frac{\epsilon^2}{4}
        \left( 
        \kappa^\star(\wb_V) T 
        + 
        \EE_{V} \left[
        \Regret (\wb_V)
        \right]
        \right)
        \\
        &\leq  C_{\operatorname{KL}} \cdot
        \frac{\epsilon^2}{2}
        \left( 
        \kappa^\star(\wb_V) T 
        + 
        d \sqrt{ \kappa^\star(\wb_V) T }
        \right)
        ,
    \end{align*}
     the equality follows from the chain rule of relative entropy (cf. Exercise 14.11 of~\citet{lattimore2020bandit}),
     the second inequality holds because $\sum_{j=1}^d   m_j(\tilde{S}_t) \leq \frac{d}{4}$,
     second-to-the last inequality holds by Lemma~\ref{lemma:kappa_star_bound},
     and the last inequality holds by the hypothesis (Equation~\eqref{eq:hypothesis}) with $ C_{\operatorname{KL}} > 0$.
     
     Furthermore, by the definition of $\kappa$ (Assumption~\ref{assum:kappa}), we know that $ \kappa^\star(\wb_V) \geq \kappa$.
     Hence,
     given the assumption that $T \geq  d^2 / \kappa \geq  d^2 / \kappa^\star (\wb_V)$,
     we derive that
     \begin{align*}
         \sum_{j=1}^d  \operatorname{KL}(P_{V} \| Q_{V \cup \{j\}}) 
        &\leq  C_{\operatorname{KL}} \cdot
        \epsilon^2
        \kappa^\star(\wb_V) T,
     \end{align*}
     which concludes the proof.
\end{proof}

\subsection{Technical Lemmas for Proposition~\ref{prop:lower_instance}}
\begin{lemma} [Lemma 7 of~\citet{abeille2021instance}]
    \label{lemma:lemma7_abelille}
    Let $f$ be a strictly increasing function such that $|\ddot{f}| \leq \dot{f}$, and let $\mathcal{Z}$ be any bounded interval of $\RR$.
    Then, for all $z_1, z_2 \in \mathcal{Z}$:
    \begin{align*}
        \int_{v=0}^1 \dot{f} \left( z_1 + v(z_2 - z_1) \right) d v 
        \geq \frac{\dot{f}(z)}{1 + |z_1 - z_2|}, \quad \text{for}\,\, z \in \{ z_1, z_2\}.
    \end{align*}
\end{lemma}
\begin{lemma} [Lemma 9 of~\citet{abeille2021instance}]
    \label{lemma:lemma9_abelille}
    Let $f$ be a strictly increasing function such that $|\ddot{f}| \leq \dot{f}$, and let $\mathcal{Z}$ be any bounded interval of $\RR$.
    Then, for all $z_1, z_2 \in \mathcal{Z}$:
    \begin{align*}
        \dot{f}(z_2) \exp (- |z_2 - z_1|)
        \leq \dot{f}(z_1)
        \leq \dot{f}(z_2)  \exp (|z_2 - z_1|).
    \end{align*}
\end{lemma}
\begin{lemma} [Lemma 11 of~\citet{perivier2022dynamic}]
    \label{lemma:kappa_star_bound}
    Let $\kappa_t^\star := \sum_{i \in S_t^\star} p_t(i | S_t^\star, \mathbf{w}^\star)p_t(0 | S_t^\star, \mathbf{w}^\star)$.
    Then, we have
    \begin{align*}
        \sum_{t=1}^T
        \sum_{i \in S_t} p_t(i | S_t, \mathbf{w}^\star)p_t(0 | S_t, \mathbf{w}^\star)
        \leq \sum_{t=1}^T \kappa_t^\star + \Regret(\wb^\star).
    \end{align*}
\end{lemma}
%

\section{Proof of Proposition~\ref{prop:upper_instance}}
\label{app_sec:proof_upper_instance}
In this section, we provide the proof of Proposition~\ref{prop:upper_instance}.
We introduce useful lemmas to support the proof.
\begin{lemma} [Theorem 4 of~\citet{tran2015composite}]
    \label{lemma:hessian_lower}
    Let $f : \RR^K \rightarrow \RR$ be a $M_f$-self-concordant-like function
    and let $x,y \in \operatorname{dom}(f)$, then:
    \begin{align*}
        e^{-M_f \| y-x \|_2} \nabla^2 f(x) \preceq \nabla^2 f(y).
    \end{align*}
\end{lemma}
\begin{proof} [Proof of Proposition~\ref{prop:upper_instance}]
    We begin the proof with Equation~\eqref{eq:eff_term_A} from the proof of Theorem~\ref{thm:upper_bound}: 
    \begin{align*}
        \sum_{t=1}^T \nabla Q(\mathbf{u}_t^\star)^\top (\mathbf{u}_t - \mathbf{u}_t^\star)
    \leq 2 \beta_T(\delta) \sum_{t=1}^T \sum_{i \in S_t} p_t(i | S_t, \mathbf{w}^\star)p_t(0 | S_t, \mathbf{w}^\star) \| x_{ti} \|_{H_t^{-1}}.
        \numberthis \label{eq:upper_instance_first}
    \end{align*}
    We then divide the total rounds into two disjoint sets, $J_1$ and $J_2$, such that  $J_1 \bigcup J_2 = [T]$.
    Specifically, let $J_1 = \{ t \in [T] | \sum_{i \in S_t} p_t(i | S_t, \mathbf{w}^\star)p_t(0 | S_t, \mathbf{w}^\star) \geq \sum_{i \in S_t} p_t(i | S_t, \mathbf{w}_{t+1})p_t(0 | S_t, \mathbf{w}_{t+1})\}$ and $J_2 = [T] \setminus J_1$.
    For a better presentation, we also define that:
    \begin{align*}
        g_t(S ; \mathbf{w}) &:= \sum_{i \in S} p_t(i | S, \mathbf{w})p_t(0 | S, \mathbf{w}) \| x_{ti} \|_{H_t^{-1}}.
    \end{align*}
    Then, we can rewrite the right-hand side of Equation~\eqref{eq:upper_instance_first} as follows:
    \begin{align*}
         2 \beta_T(\delta) \sum_{t=1}^T \sum_{i \in S_t} &p_t(i | S_t, \mathbf{w}^\star)p_t(0 | S_t, \mathbf{w}^\star) \| x_{ti} \|_{H_t^{-1}}
         \\
         &=
         2 \beta_T(\delta) \sum_{t=1}^T g_t(S_t; \mathbf{w}^\star)
         \\
         &= 2 \beta_T(\delta) \sum_{t \in J_1} g_t(S_t; \mathbf{w}^\star) + 2 \beta_T(\delta) \sum_{t \in J_2} g_t(S_t; \mathbf{w}^\star).
    \end{align*}
    To bound $\sum_{t \in J_1} g_t(S_t; \mathbf{w}^\star)$, by the mean value theorem, we get
    \begin{align*}
        \sum_{i \in J_1} g_t(S_t; \mathbf{w}^\star)
    = \sum_{i \in J_1} g_t(S_t; \mathbf{w}_{t+1}) + \sum_{i \in J_1} \nabla_{\mathbf{w}} g_t(S_t; \bar{\mathbf{w}}_t)^\top (\mathbf{w}^\star - \mathbf{w}_{t+1}),
        \numberthis \label{eq:upper_instance_J_1}
    \end{align*}
    where $\bar{\mathbf{w}}_t$ is the convex combination of $\mathbf{w}^\star$ and $\mathbf{w}_{t+1}$.  
    The first term of Equation~\eqref{eq:upper_instance_J_1} can be bound by  
    \begin{align*}
        &\sum_{t \in J_1} g_t(S_t; \mathbf{w}_{t+1})
        \\
        &\leq \sqrt{\sum_{t\in J_1} \sum_{i \in S_t} p_t(i | S_t, \mathbf{w}_{t+1})p_t(0 | S_t, \mathbf{w}_{t+1})} \sqrt{\sum_{t \in J_1} \sum_{i \in S_t} p_t(i | S_t, \mathbf{w}_{t+1})p_t(0 | S_t, \mathbf{w}_{t+1}) \| x_{ti}\|_{H_t^{-1}}^2 }
        \\
        &\leq \sqrt{\sum_{t\in J_1} \sum_{i \in S_t} p_t(i | S_t, \mathbf{w}^\star)p_t(0 | S_t, \mathbf{w}^\star)} \cdot \BigOTilde \left( \sqrt{d} \right)  
        \leq \sqrt{\sum_{t=1}^T \kappa_t^\star + \Regret(\wb^\star)} \cdot \BigOTilde \left( \sqrt{d} \right) 
        \numberthis \label{eq:upper_instance_J_1_1}
    ,
    \end{align*}
    where the second-to-the last inequality holds by the definition of $J_1$ and the elliptical potential lemma (Lemma~\ref{lemma:elliptical}), and the last inequality holds by the following Lemma:
    
    Moreover, the second term of Equation~\eqref{eq:upper_instance_J_1} can be bounded as follows:
    \begin{align*}
    &\left|\sum_{i \in J_1}\nabla_{\mathbf{w}} g_t(S_t; \bar{\mathbf{w}}_t)^\top (\mathbf{w}^\star - \mathbf{w}_{t+1}) \right|
    \\
    &=\Bigg|\sum_{t \in J_1} p_t(0 | S_t, \bar{\mathbf{w}}_t) \sum_{i \in S_t}p_t(i | S_t, \bar{\mathbf{w}}_t) x_{ti}^\top (\mathbf{w}^\star - \mathbf{w}_{t+1}) \| x_{ti} \|_{H_t^{-1}}
    \\
    &- 2 p_t(0 | S_t, \bar{\mathbf{w}}_t) \sum_{i \in S_t}p_t(i | S_t, \bar{\mathbf{w}}_t) \| x_{ti} \|_{H_t^{-1}} \sum_{j \in S_t}p_t(j | S_t, \bar{\mathbf{w}}_t) x_{tj}^\top (\mathbf{w}^\star - \mathbf{w}_{t+1}) \Bigg| 
    \\
    &\leq \beta_{T}(\delta) \sum_{t \in J_1} \sum_{i \in S_t}p_t(i | S_t, \bar{\mathbf{w}}_t)  \| x_{ti} \|_{H_t^{-1}}^2
    + 2 \beta_{T}(\delta) \sum_{t \in J_1} \left( \sum_{i \in S_t}p_t(i | S_t, \bar{\mathbf{w}}_t)  \| x_{ti} \|_{H_t^{-1}} \right)^2
    \\
    &\leq \beta_{T}(\delta) \sum_{t \in J_1} \max_{i \in S_t} \| x_{ti} \|_{H_t^{-1}}^2
    + 2 \beta_{T}(\delta) \sum_{t \in J_1} \left( \max_{i \in S_t} \| x_{ti} \|_{H_t^{-1}} \right)^2
    \\
    &\leq 3 \beta_{T}(\delta) \sum_{t=1}^T \max_{i \in S_t} \| x_{ti} \|_{H_t^{-1}}^2 = \BigOTilde \left(d^{3/2}/\kappa \right),
    \numberthis \label{eq:upper_instance_J_1_2}
    \end{align*}
    where the first inequality holds by Lemma~\ref{lemma:online_confidence_set},
    and the last inequality holds by Lemma~\ref{lemma:elliptical}.
    Combining Equation~\eqref{eq:upper_instance_J_1_1} and~\eqref{eq:upper_instance_J_1_2}, we can bound the term $\sum_{t \in J_1} g_t(S_t; \mathbf{w}^\star)$ as follows:
    \begin{align*}
        \sum_{t \in J_1} g_t(S_t; \mathbf{w}^\star)
        \leq \BigOTilde \left( \sqrt{d}  \cdot  \sqrt{ \sum_{t=1}^T \kappa_t^\star + \Regret(\wb^\star) } + d^{3/2}/\kappa  \right).
        \numberthis \label{eq:upper_instance_J_1_final}
    \end{align*}
    Now, we bound $\sum_{t \in J_2} g_t(S_t; \mathbf{w}^\star)$.
    We define $H_t^\star := \lambda \mathbf{I}_d + \sum_{s=1}^{t-1} \mathcal{G}_s(\wb^\star)$.
    Recall that $\nabla^2 \ell_s(\wb) = \mathcal{G}_s(\wb)$ and $\ell_s$ is $3 \sqrt{2}$-self-concordant-like function (Proposition~\ref{prop:self_concordant}).
    Then, by Lemma~\ref{lemma:hessian_lower}, we get $H_t \succeq H_t^\star e^{-3\sqrt{2}} $.
    With this fact, we can now bound the term $\sum_{t \in J_2} g_t(S_t; \mathbf{w}^\star)$:
    \begin{align*}
        \sum_{t \in J_2} g_t(S_t; \mathbf{w}^\star)
        &\leq \sqrt{\sum_{t \in J_2} \sum_{i \in S_t} p_t(i | S_t, \mathbf{w}^\star)p_t(0 | S_t, \mathbf{w}^\star)} \sqrt{\sum_{t \in J_2} \sum_{i \in S_t} p_t(i | S_t, \mathbf{w}^\star)p_t(0 | S_t, \mathbf{w}^\star) \| x_{ti} \|_{H_t^{-1}}^2 }
        \\
        &\leq \sqrt{\sum_{t=1}^T \kappa_t^\star + \Regret} \sqrt{ \sum_{t=1}^T \sum_{i \in S_t} p_t(i | S_t, \mathbf{w}^\star)p_t(0 | S_t, \mathbf{w}^\star) \| x_{ti} \|_{H_t^{-1}}^2 }
        \\
        &\leq \sqrt{\sum_{t=1}^T \kappa_t^\star + \Regret}  \sqrt{ e^{3\sqrt{2}} \sum_{t=1}^T \sum_{i \in S_t} p_t(i | S_t, \mathbf{w}^\star)p_t(0 | S_t, \mathbf{w}^\star) \| x_{ti} \|_{ \left(H_t^\star\right) ^{-1}}^2 }
        \\
        &\leq \tilde{\mathcal{O}}\left( \sqrt{d} \cdot \sqrt{\sum_{t=1}^T \kappa_t^\star + \Regret}  \right),
        \numberthis \label{eq:upper_instance_J_2_final}
    \end{align*}
    where the second inequality follows from Lemma~\ref{lemma:kappa_star_bound},
    second-to-the last inequality holds because  $H_t \succeq H_t^\star e^{-3\sqrt{2}} $,
    and the last inequality is obtained by a slight modification of Lemma~\ref{lemma:elliptical} with $\wb_1, \dots, \wb_{T} = \wb^\star$.

    Now, by plugging Equation~\eqref{eq:upper_instance_J_1_final} and~\eqref{eq:upper_instance_J_2_final} into Equation~\eqref{eq:upper_instance_first}, and using the fact that $\beta_T(\delta) = \BigOTilde(\sqrt{d})$, we obtain that
    \begin{align*}
        2 \beta_T(\delta) \sum_{t=1}^T \sum_{i \in S_t} p_t(i | S_t, \mathbf{w}^\star)p_t(0 | S_t, \mathbf{w}^\star) \| x_{ti} \|_{H_t^{-1}}
        =  \BigOTilde \left( d  \sqrt{ \sum_{t=1}^T \kappa_t^\star + \Regret(\wb^\star) } + d^{3/2}/\kappa  \right).
    \end{align*}
    Furthermore, by applying the same analysis as in the proof of Theorem~\ref{thm:upper_bound} in order to bound the second-order term in the Taylor expansion (term \texttt{(B)} in Equation~\eqref{eq:upper_uniform_regret_decompose}), we get
    \begin{align*}
        \Regret(\wb^\star) =  \BigOTilde \left( d  \sqrt{ \sum_{t=1}^T \kappa_t^\star + \Regret(\wb^\star) } + d^2/\kappa  \right).
    \end{align*}
    By solving the above equation, we conclude the proof of Proposition~\ref{prop:upper_instance}.
\end{proof}

\section{Experiment Details and Additional Results}
\label{app_sec:experimat_details}
\begin{figure*}[htp!]
    \centering
    \begin{subfigure}[b]{0.32\textwidth}
        \includegraphics[width=\textwidth, trim=0mm 0mm 11mm 5mm, clip]{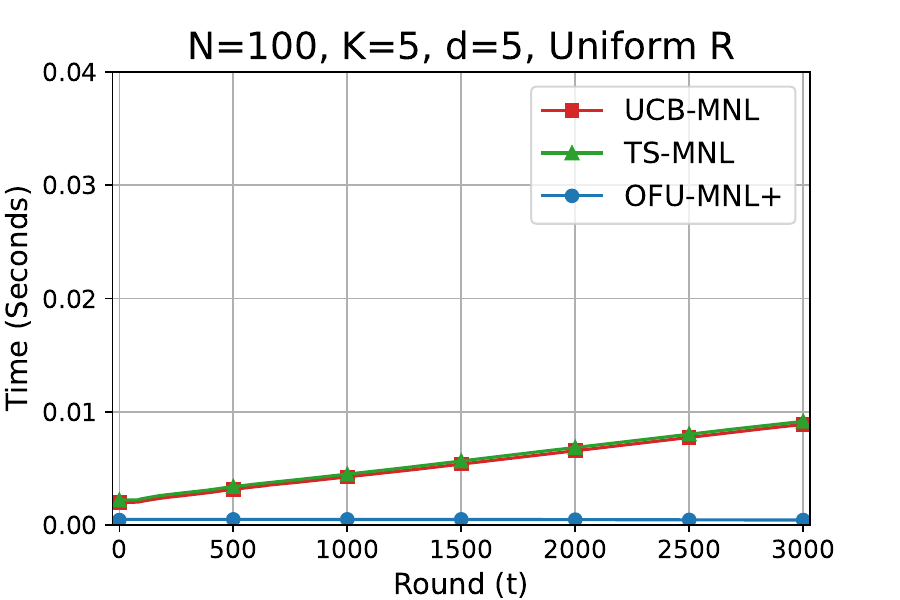}
        \label{fig:N=100_K=5_d=5_dist=0_r}
    \end{subfigure}
    \begin{subfigure}[b]{0.32\textwidth}
        \includegraphics[width=\textwidth, trim=0mm 0mm 11mm 5mm, clip]{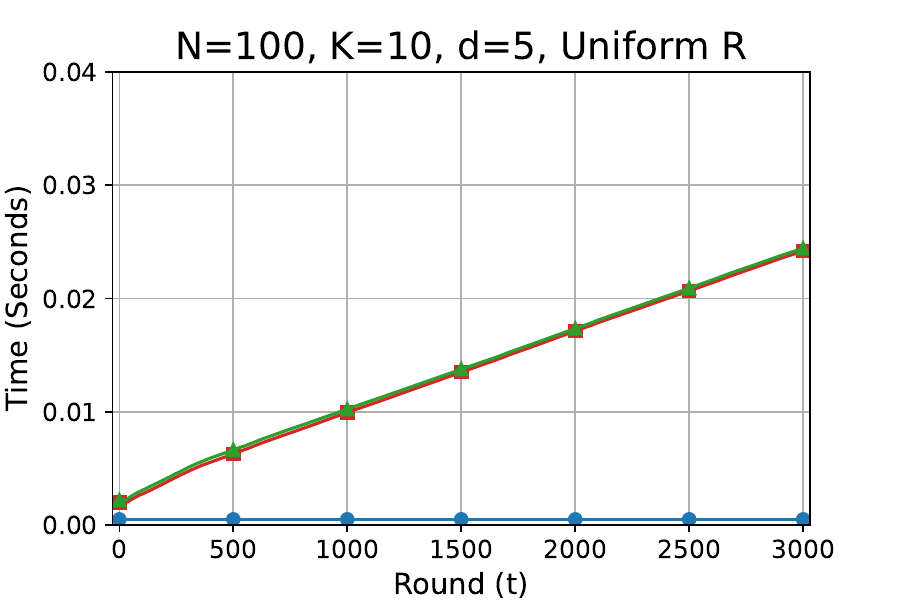}
        \label{fig:N=100_K=10_d=5_dist=0_r}
    \end{subfigure}
    \begin{subfigure}[b]{0.32\textwidth}
        \includegraphics[width=\textwidth, trim=0mm 0mm 11mm 5mm, clip]{figures/mnlBandit_N=100_K=15_d=5_dist=0_runtime.pdf}
        \label{fig:N=100_K=15_d=5_dist=0_r}
    \end{subfigure}
    
    \begin{subfigure}[b]{0.32\textwidth}
        \includegraphics[width=\textwidth, trim=0mm 0mm 11mm 5mm, clip]{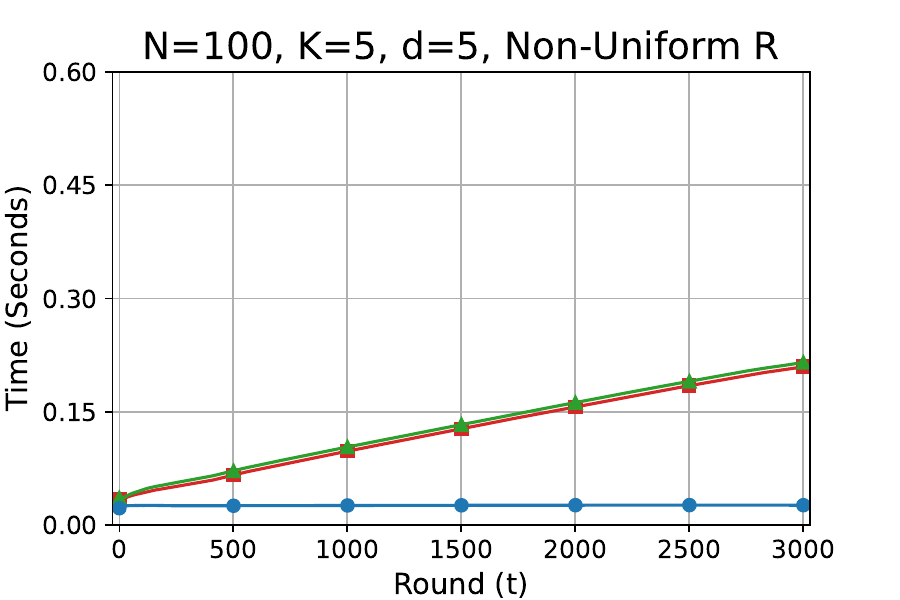}
        \label{fig:N=100_K=5_d=5_dist=0_r_nu}
    \end{subfigure}
    \begin{subfigure}[b]{0.32\textwidth}
        \includegraphics[width=\textwidth, trim=0mm 0mm 11mm 5mm, clip]{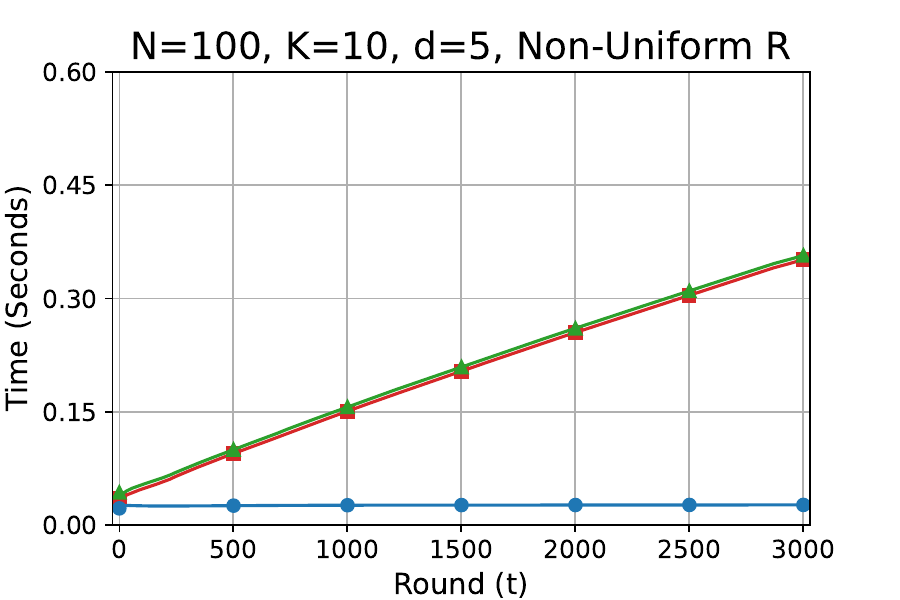}
        \label{fig:N=100_K=10_d=5_dist=0_r_nu}
    \end{subfigure}
    \begin{subfigure}[b]{0.32\textwidth}
        \includegraphics[width=\textwidth, trim=0mm 0mm 11mm 5mm, clip]{figures/mnlBandit_N=100_K=15_d=5_dist=0_runtime_nu.pdf}
        \label{fig:N=100_K=15_d=5_dist=0_r_nu}
    \end{subfigure}
    \caption{Runtime per round under uniform rewards (first row) and non-uniform rewards (second row). }
    \label{fig:runtime}
\end{figure*}
\begin{figure*}[h]
    \centering
    \begin{subfigure}[b]{0.32\textwidth}
        \includegraphics[width=\textwidth, trim=0mm 0mm 11mm 5mm, clip]{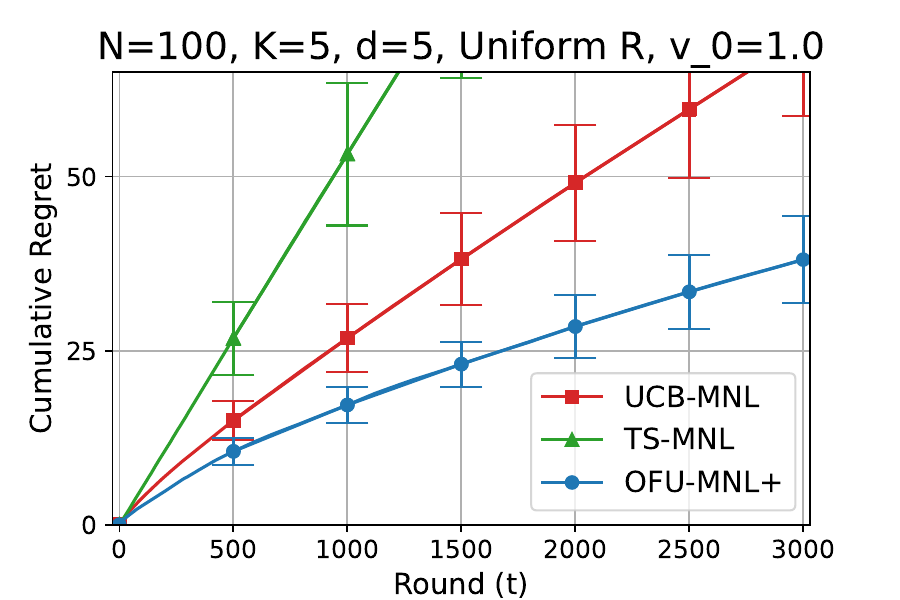}
        \label{fig:v_0=5}
    \end{subfigure}
    \begin{subfigure}[b]{0.32\textwidth}
        \includegraphics[width=\textwidth, trim=0mm 0mm 11mm 5mm, clip]{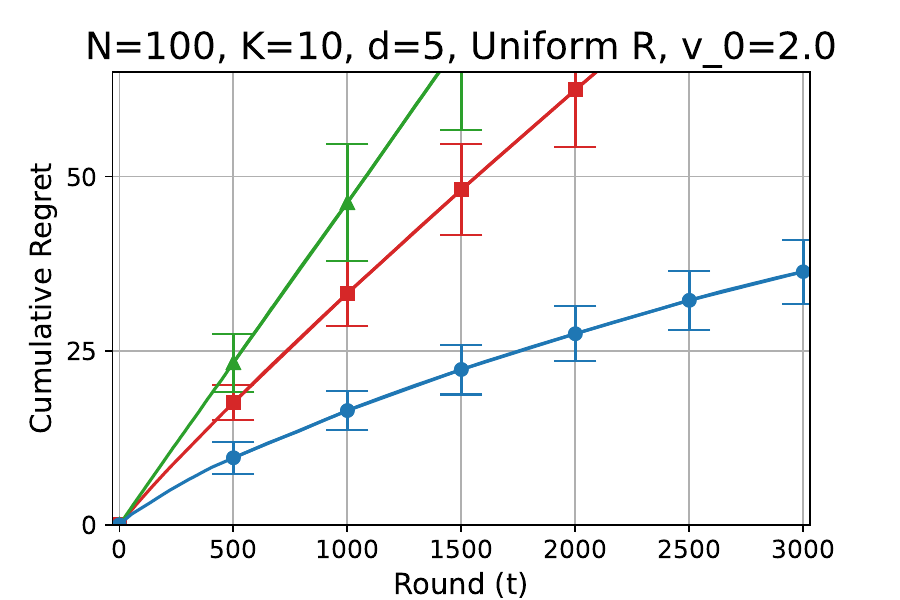}
        \label{fig:v_0=10}
    \end{subfigure}
    \begin{subfigure}[b]{0.32\textwidth}
        \includegraphics[width=\textwidth, trim=0mm 0mm 11mm 5mm, clip]{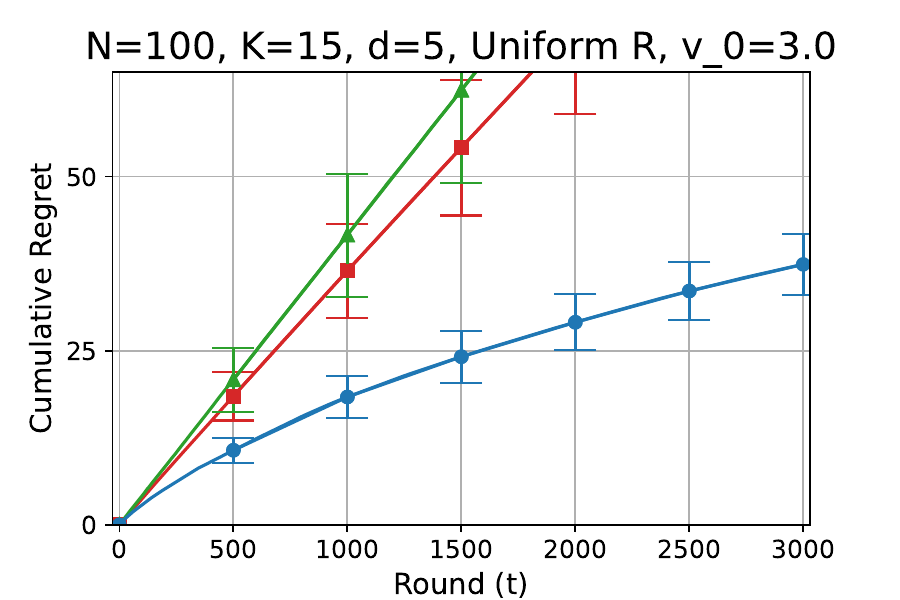}
        \label{fig:v_0=15}
    \end{subfigure}
    \caption{Cumulative regret under uniform rewards with $v_0 = \Theta(K)$.}
    \label{fig:varying_v0}
\end{figure*}
%
For each instance, we sample the true parameter $\wb^\star$ from a uniform distribution in $[-1/\sqrt{d},1/\sqrt{d}]^d$. 
For the context features $x_{ti}$, we sample each $x_{ti}$ independently and identically distributed (i.i.d.)  from a multivariate Gaussian distribution $\mathcal{N}(\mathbf{0}_d, \Ib_d)$ and clip it to range $[-1/\sqrt{d},1/\sqrt{d}]^d$.
Therefore, we ensure that $\| \wb^\star \|_2 \leq 1$ and $\| x_{ti} \|_2 \leq 1$, satisfying Assumption~\ref{assum:bounded_assumption}.
For each experimental configuration, we conducted $20$ independent runs for each instance and reported the average cumulative regret (Figure~\ref{fig:regret_main}) and runtime per round (Figure~\ref{fig:runtime}) for each algorithm.
The error bars in Figure~\ref{fig:regret_main} and~\ref{fig:varying_v0} represent the standard deviations (1-sigma error).
We have omitted the error bars in Figure~\ref{fig:runtime} because they are minimal.

In the uniform reward setting where $r_{ti}=1$, the combinatorial optimization step to select the assortment simply involves sorting items by their utility estimate. 
In contrast, in the non-uniform reward setting, rewards are sampled from a uniform distribution in each round, i.e., $r_{ti} \sim \operatorname{Unif}(0,1)$.
For combinatorial optimization in this setting, we solve an equivalent linear programming (LP) problem that is solvable in polynomial-time~\citep{rusmevichientong2010dynamic, davis2014assortment}.
The experiments are run on Xeon(R) Gold 6226R CPU @ 2.90GHz (16 cores).

Figure~\ref{fig:runtime} presents additional empirical results on  the runtime per round. 
Our algorithm \AlgName{} demonstrates a constant computation cost for each round, while the other algorithms exhibit a linear dependence on $t$.
It is also noteworthy that the runtime for uniform rewards is approximately $10$ times faster than that for non-uniform rewards. 
This difference arises because we use linear programming (LP) optimization for assortment selection in the non-uniform reward setting, which is more computationally intensive.

Furthermore, Figure~\ref{fig:varying_v0} illustrates the cumulative regrets of the proposed algorithm compared to other baseline algorithms under uniform rewards with $v_0=K/5$.
Since $v_0$ is proportional to $K$, an increase in $K$ does not improve the regret.
This observation is also consistent with our theoretical results.
\section{Technical Errors in~\citet{agrawal2023tractable}}
\label{app_sec:tech_error_agrawl}
In this section, we discuss the technical errors in~\citet{agrawal2023tractable}. There are two main significant errors:
There are mainly two significant errors:

\textbf{1. Equation (16).}
\begin{align*}
    \alpha_i(\mathbf{X}_{S_t}, \theta_t, \theta^{\star}) x_{ti} :=\mu_i(\mathbf{X}_{S_t}^\top \theta^\star) - \mu_i(\mathbf{X}_{S_t}^\top \theta_t),
    \numberthis \label{eq:agrawal_error_1}
\end{align*}
where $\mathbf{X}_{S_t}$ is a design matrix whose columns are the attribute vectors $x_{ti}$ of the items in the assortment $S_t$ and $\mu_i(\mathbf{X}_{S_t}^\top \theta) = P_t(i | S_t, \theta)$.

It appears that the authors may have intended to derive this equation using a first-order exact Taylor expansion. However, in  MNL bandits, this equation generally does not hold. Consider a counterexample where $x_{ti}=0$, $x_{tj} \neq 0$ for $j \neq i$, and $\theta^\star \neq \theta_t$. Then, the left-hand side of Equation~\eqref{eq:agrawal_error_1} is equals to $0$ (since $x_{ti}=0$), but the right-hand side of Equation~\eqref{eq:agrawal_error_1} is not $0$, because the denominators of each $\mu_i(\mathbf{X}_{S_t}^\top \theta^\star)$ and $\mu_i(\mathbf{X}_{S_t}^\top \theta_t)$ differ. This equation only holds in special cases, such as when $K=1$, which corresponds to the logistic bandit case. 
Equation (16) in~\citet{agrawal2023tractable} serves as the foundation for the entire proof in their paper.
Consequently, all subsequent results derived from it are also incorrect. 

\textbf{2. Cauchy-Schwarz inequality in the regret analysis (Page 46)}
\\
When using the Cauchy-Schwarz inequality on the regret before applying the elliptical potential lemma, they indeed incur an additional $\sqrt{K}$ factor:
\begin{align*}
    \sum_{t=1}^T \min \left(\sum_{i \in S_t}\|x_{ti}\|_{\mathbf{J}_t^{-1}},1 \right) \leq \sqrt{KT} \sqrt{\min \left(\sum_{i \in S_t} 
    \|x_{ti}\|_{\mathbf{J}_t^{-1}}^2 ,1 \right)}.
\end{align*}
Hence, their regret should actually be $\tilde{\mathcal{O}}(d\sqrt{KT} + d^2/\kappa)$, which is worse than the result in our Theorem~\ref{thm:upper_bound} (upper bound under uniform rewards) by a factor of $K$.


\end{document}